\documentclass{article}
\pdfoutput=1

\usepackage[preprint]{neurips_2021}

\usepackage{microtype}
\usepackage{graphicx}
\usepackage{wrapfig}
\usepackage{subfigure}
\usepackage{booktabs} 
\usepackage{tcolorbox}

\usepackage{hyperref}

\usepackage{notation}
\usepackage{cleveref}
\usepackage{multirow, array}
\usepackage{tikz}
\usetikzlibrary{arrows.meta}
\usepackage{bbold}
\usepackage{xcolor}
\usepackage{color-edits}
\usepackage{tcolorbox}
\usepackage{enumitem}
\usepackage{nicefrac}

\newcommand{\negspacepresubsec}{0em}
\newcommand{\negspacesubsec}{0em}

\crefname{section}{\S}{\S\S}
\crefname{subsection}{\S}{\S\S}
\crefname{figure}{Fig.}{Figs.}
\crefname{table}{Tab.}{Tabs.}

\newcommand{\affnum}[1]{{\normalsize\rm\textsuperscript{\,#1}}}
\newcommand{\affiliation}[2]{\normalsize\rm\textsuperscript{#1}#2}
\newcommand{\name}[1]{\textbf{#1}}
\def\and{\ }

\title{Backward-Compatible Prediction Updates: \\A Probabilistic Approach}

\author{%
\name{Frederik Tr{\"a}uble}\affnum{1,2} \hspace{-2pt}\thanks{Work done while FT and JvK were interning at Amazon.} \and 
\name{Julius von K{\"u}gelgen}\affnum{1,2,3} \hspace{-2pt}\footnotemark[1] \and
\name{Matthäus Kleindessner}\affnum{1} \\
\name{Francesco Locatello}\affnum{1} \and
\name{Bernhard Schölkopf}\affnum{1} \and
\name{Peter Gehler}\affnum{1}\thanks{Correspondence to: \texttt{frederik.traeuble@tuebingen.mpg.de}  and \texttt{pgehler@amazon.com}}

\\[6pt]

\affiliation{1}{Amazon Tübingen, Germany} \\
\affiliation{2}{Max Planck Institute for Intelligent Systems, Tübingen, Germany}\\
\affiliation{3}{Department of Engineering, University of Cambridge, United Kingdom}
}

\begin{document}

\maketitle

\vspace{-1.5em}

\begin{abstract}

When machine learning systems meet real world applications, accuracy is only one of several requirements.
In this paper, we assay a complementary perspective originating from the increasing availability of pre-trained and regularly improving state-of-the-art models. 
While new improved models develop at a fast pace, downstream tasks vary more slowly or stay constant. 
Assume that we have a large unlabelled data set for which we want to maintain accurate predictions.
Whenever a new and presumably better ML models becomes available, we encounter two problems: (i) given a limited budget, which data points should be re-evaluated using the new model?; and (ii) if the new predictions differ from the current ones, should we update? 
Problem (i) is about compute cost, which matters for very large data sets and models. 
Problem (ii) is about maintaining consistency of the predictions, which can be highly relevant for downstream applications; our demand is to avoid negative flips, i.e., changing correct to incorrect predictions. 
In this paper, we formalize the Prediction Update Problem and present an efficient probabilistic approach as answer to the above questions. 
In extensive experiments on standard classification benchmark data sets, we show that our method outperforms alternative strategies along key metrics for backward-compatible prediction updates.
\end{abstract}

\section{Introduction}
\label{sec:introduction}

The machine learning~(ML) community develops new models at a fast pace:
 for example, just in the past year, the state-of-the-art on ImageNet has changed at least
five times~\citep{dosovitskiy2020image,xie2020self,touvron2020fixing,foret2020sharpness,pham2020meta}. As reproducibility has increasingly been scrutinized~\citep{pineau2018iclr,sinha2020neurips,pineau2019iclr}, it is now common practice to release pre-trained models upon publication.
In this work we take the perspective of an owner of an unlabelled data set who
is interested in keeping
 the best possible predictions at all times.
When a new pre-trained model is released, we face what we refer to as the 
\emph{Prediction Update Problem}: (i) decide which points in the data set to re-evaluate with the new model, and (ii) integrate the new, possibly contradicting, predictions. For this task, we postulate the following three 
 desiderata:
 \vspace{-0.2em}
\begin{tcolorbox}
\vspace{-0.5em}
\begin{enumerate}[leftmargin=3pt]
 \setlength\itemsep{0em}
\item The prediction updates should improve overall accuracy.
\item The prediction updates should avoid introducing new errors.
\item The prediction updates should be as cheap as possible since the
target data set could be huge.
\end{enumerate}
\vspace{-0.95em}
\end{tcolorbox}
 \vspace{-0.4em}
We consider the setting in which the target data set for which we wish to maintain
predictions  is fully unlabelled (i.e., the ground-truth labels are unknown) and may come from a different distribution than the one 
 on which models have been pre-trained, but with overlap in the label space.
This is a transductive or semi-supervised problem, but, due to computational constraints, we avoid any model fitting or fine-tuning and rely solely on the predictions of 
the pre-trained models that are released over time. 
Typically, these models exhibit
  increased performance on their labelled training domain (e.g., the ImageNet validation or test set)
as evidence for being good candidates for re-evaluation.
Clearly, one goal of updating the predictions stored for the
target data set is to improve overall performance, e.g.,
top-k accuracy for classification.
At the same time, the stored
 predictions may form an intermediate step in a larger ML pipeline or are 
 accessible to users. 
 This is the reason for our second desideratum: we would like to be \emph{backward-compatible}, i.e., new predictions should not flip previously correct predictions (\emph{negative flips}). 
 Finally, we aim to reduce computational cost during inference and 
 to avoid evaluating the entire data set which may be prohibitive in practice and unnecessary if we are already somewhat certain about a prediction.

\looseness=-1In this paper, we motivate and formalize the \textit{Prediction Update Problem} and
describe its relation to various 
  relevant research areas like ensemble learning, domain adaption, active learning, and others.
We propose a probabilistic approach that maintains a posterior distribution over the unknown true labels
by combining all previous model re-evaluations.
Based on these uncertainty estimates, we devise an efficient \textit{selection strategy} which only chooses those examples with highest posterior label entropy for re-evaluation
in order 
 to reduce computational cost. Furthermore, we consider different prediction-update strategies to decide whether to change the stored predictions, taking asymmetric costs for negative and positive flips into account.
Using the task of image classification as a case study, we perform extensive experiments on common benchmarks (ImageNet, CIFAR10, and ObjectNet) and demonstrate that our approach achieves competitive accuracy and introduces much fewer negative flips across a range of
 computational budgets,
  thus showing that our three desiderata are not necessarily at odds.

\textbf{Contributions \,}
We highlight the following contributions:
\begin{itemize}
\item We introduce the \emph{Prediction Update Problem} which addresses some common, but previously unaddressed challenges faced in real world ML systems~(\cref{sec:problem_setting}).
\item We propose a probabilistic, model-agnostic approach for the \emph{Prediction Update Problem}, based on Bayesian belief estimates of the true label combined with an efficient selection and different prediction-update strategies~(\cref{sec:methods}).
\item We contextualise this understudied problem setting as well as our method with related work~(\cref{sec:limitations} \& \cref{sec:related_work}) and discuss several 
extensions and limitations~(\cref{sec:limitations}).
\item We demonstrate that our 
approach successfully outperforms alternative approaches and accomplishes all our desiderata in experiments across multiple common benchmark datasets (CIFAR-10, ImageNet, and ObjectNet) and practically relevant scenarios~(\cref{sec:experiments}).
\end{itemize}

\subsection{Backward-Compatible Prediction Systems}
\vspace{\negspacesubsec}

In real world ML applications, empirical performance is only one of several requirements.
When humans interact with automatic predictions, they will start to build mental models 
of how these models operate and whether and when their predictions can be trusted. 
This is described as Human-AI teams by~\cite{bansal2019updates} who argue to ``make the human factor a first-class consideration of AI updates''.

An example from~\cite{bansal2019updates} is autopilot functionality in cars for which drivers will build expectations in which driving situations the autopilot is safe to engage. It is important not to violate these assumptions when updating the models over the air.
AI assisted medical decision processes are another example of a high stake application where medical professionals need to understand when systems can be trusted.

Consider the example of automatically tagging images in a user's photo collection.
Those tags are used for example in photo search.
As models progress, the overall accuracy on all uploaded images may increase, but for any single user the experience can deteriorate if previously correct searches now show wrong results.
Even worse, if errors fluctuate over the user's photo collection as the result of prediction-updates, the user's trust will be eroded.
This ``cost'' is asymmetric and the negative experience may outweigh the benefit of better predictions on other images.

In contrast to carefully curated and labelled ML benchmarks, many real-world data sets are magnitudes larger (up to billions of samples) and entirely unlabelled.
Having no feedback which predictions are correct is a common scenario: consider any type of private data such as health data, photo collections, or personal information.
Because the data is private, we can neither train on it, nor collect feedback, nor observe the effect of predictions.
On the other hand, such data is valuable to an individual:  she has an interest to keep it up to date with the best possible predictions.
Since it is of little consolation to her if an update of the model improves predictions on average but on her data it gets worse, the update costs are asymmetric.
Service providers often rely on models pre-trained on a different data set, and
 the desire to be backward-compatible arises naturally in this setting~\citep{yan2020positive,shen2020towards,bansal2019updates,srivastava2020empirical}.

\looseness=-1This is an understudied problem where progress will have large impact. ML systems are becoming pervasive, and their accuracy will continue to increase. Being able to seamlessly transfer them to existing data will be crucial for real-world ML systems.

\begin{figure*}
	\centering
    	\includegraphics[width=0.9\textwidth]{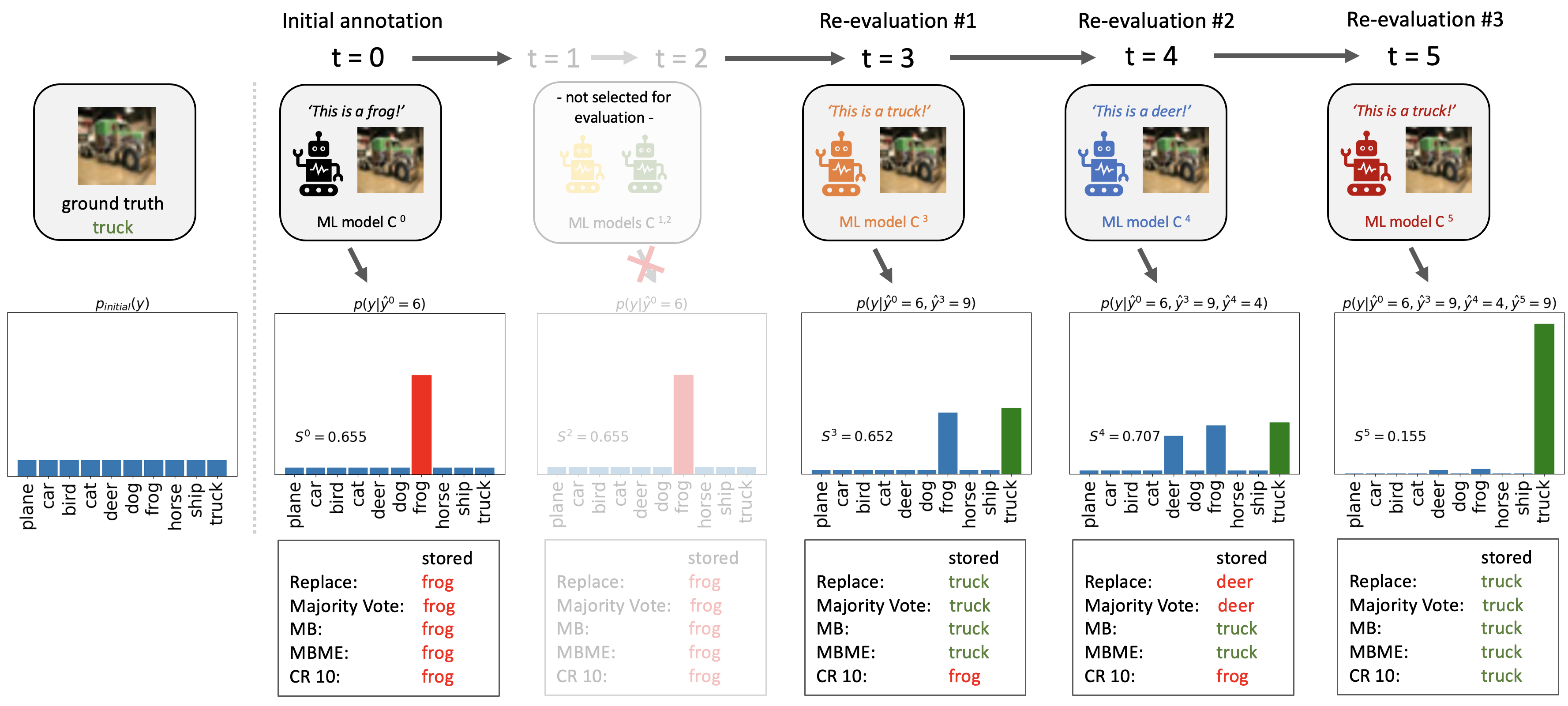}
	\vspace{-0.5em}
	\caption{\small \textbf{Overview of our proposed Bayesian approach to the Prediction Update Problem.}
	Starting from a uniform prior, we maintain a posterior distribution $p(y=k|\yh^{0:t})$ (\textit{middle}) over the unknown true label $y$ of an unlabelled sample which takes the predictions $\yh^{0:t}$ from new ML models $C^t$ (\textit{top}) arriving over time $t=0, ..., T$ into account.
	Given a limited compute budget $B^t$, we re-evaluate those samples with highest posterior label entropy $S^t$ at each time step, e.g., the example shown is first selected in time step $t=3$ after the initial annotation at $t=0$.
	 We then consider different strategies for deciding whether to update the stored prediction (\textit{bottom}) based on our changed beliefs. Note that the non-probabilistic baselines ``Replace'' (always update to last prediction) and ``Majority Vote'' (resolve ties by using the last prediction) incorrectly update the stored prediction from ``truck'' to ``deer'' in step $t=4$. Our strategies (MB, MBME, CR-10) which rely on the estimated label posterior, on the other hand, avoid such a \textit{negative flip}, which is one of our key goals.
	 }
    	\label{fig:overview_figure}
	\vspace{-1em}
\end{figure*}

\section{The Prediction Updates Problem Setting}
\label{sec:problem_setting}

\textbf{Target data set \,} 
We are given a large, unlabelled target data set $\Dcal^\trg=\{\xb_n\}_{n=1}^N$ comprising $N$ independent and identically distributed (i.i.d.) observations $\xb_n \in\Xcal\subseteq\RR^d$ 
drawn from a
target distribution $\PP_\Xb^\trg$.
The ground-truth labels $y_n\in \Ycal=\{1,...,K\}$ distributed according to $\PP_{Y|\Xb}^\trg$ are \emph{not observed}.
Note that we are particularly interested in a scenario where $N$ may be extremely large.

\textbf{Models \,}
\looseness=-1Over time $t=0, 1, ..., T$ we successively gain access to
 classifiers
$C^0, C^1, \ldots, C^T : \Xcal\rightarrow \Ycal$
which have been trained on a labelled 
data set 
 $\Dcal^\src$ 
from a potentially different source distribution $\PP_{\Xb, Y}^\src$ over $\Xcal\times\Ycal$. For simplicity, we assume that the observation space $\Xcal$ and label space $\Ycal$ are shared.
We consider both the 
standard 
scenario where the models $\{C^t\}_{t=0}^T$ are trained on a 
labelled set from the same domain ($\PP^\src=\PP^\trg$); and the transfer scenario where we deploy a model trained on a labelled ML benchmark to a different data set ($\PP^\src\neq\PP^\trg$). 
We assume that $\{C^t\}_{t=0}^T$ are improving in performance on the training data set. Therefore, denoting by $A_t$ the estimated accuracy of $C^t$ on $\PP_{\Xb, Y}^\src$, we have $A^t\leq A^{t+1}$ $\forall t$.
 As motivating example, consider an object recognition task in the wild and let $C_t$ be the winning entry of the ImageNet competition in year~$t$.
 
\textbf{Labelling \,}
To relate the source and target distributions and to justify applying $\{C^t\}_{t=0}^T$ to our target data set $\Dcal^\trg$, we make the commonly used covariate shift assumption $\PP^\trg_{Y|\Xb}=\PP^\src_{Y|\Xb}$, i.e. the conditional label distribution is shared across source and target distributions ~\citep{shimodaira2000improving, sugiyama2012machine}.

We denote the \emph{predicted} label by $C^t$ for $\xb_n$ by~$\yh_n^t=C^t(\xb_n)$
and the \emph{stored} prediction for $\xb_n$ after time step $t$ by $l_n^t$.
The target data set is then initially fully labelled by $C^0$, i.e., $l_n^0:=\yh_n^0$.

\textbf{Objective \,} 
\looseness=-1As new classifiers $\{C^t\}_{t\geq 1}$ become available, our main objective is to maintain the best estimates $\{l_n^t\}_{n=1}^N$  on our target data set at all times and improve overall accuracy, while, at the same time, maintaining backward compatibility by minimising the number of \emph{negative flips}, i.e., the number of previously correctly stored predictions that are incorrectly changed.
The key challenge is that no ground truth labels for our target data set are available, so that we have no feedback on which predictions are correct and which are wrong.
For each test sample $\xb_n$ and each time step $t\geq1$, we thus need to decide whether or not to update the previously stored prediction $l_n^{t-1}$ based on the current and previous model predictions $\yh_n^t$ and $\yh_n^{0:t-1}$, respectively.

\textbf{Limited evaluation budget \,}
\looseness=-1Re-evaluating all samples (so-called \emph{backfilling}) can be very costly and requires significant resources.  Since we consider $N$ to be very large, we also consider a limited budget of at most $B^t \leq N$ sample re-evaluations for step $t$. We thus additionally need to decide how to allocate this budget and select a subset of samples to be re-evaluated by $C^t$ at every step.

\section{Our Method}
\label{sec:methods}
Having specified the setting, we next describe our proposed method for the Prediction Update Problem.
We start by providing a Bayesian approach for maintaining and updating our beliefs about the unknown true labels as new predictions become available~(\cref{sec:Bayesian_updates}), followed by describing strategies for selecting candidate samples for re-evaluation~(\cref{sec:selection_strategy}) and for updating the stored predictions based on our changed beliefs~(\cref{sec:label_update_strategies}).
Our framework is summarised in Figure~\ref{fig:overview_figure}.

\vspace{\negspacepresubsec}
\subsection{Bayesian Approach}
\label{sec:Bayesian_updates}
\vspace{\negspacesubsec}
Since the true labels $\{y_n\}_{n=1}^N$ are unknown to us, we treat them as random quantities over which we maintain uncertainty estimates. We then perform Bayesian reasoning to update our beliefs as new evidence in the form of predictions $\yh_n^t$ from newly-available classifiers $C^t$ arrives over time $t=1, ..., T$.
In standard Bayesian notation, the true labels $y_n$ thus take the role of unknown parameters $\theta$ and the predictions $\yh_n^t$  of data $x$.
Since $\Dcal^\trg$ is sampled i.i.d., we reason about each label $y_n$ independently of the others, i.e., the following is the same for all $n$.

\textbf{Prior \,}
Lacking label information on the target data set, we choose a uniform prior over $\Ycal$ for all $y_n$, i.e.,
$p(y_n=k)=\nicefrac{1}{K}, \, \forall k\in\Ycal$.
If (estimates of) the class probabilities on $\Dcal^\trg$ are available, we may instead use these as a more informative prior.

\textbf{Likelihood \,}
Next, we need to specify a likelihood function $p(\yh_n^{0:T}|y_n=k)$  for the observed model predictions $\yh_n^{0:T}$ given a  value $k$ of the true label $y_n$.
We make the following simplifying assumption.
\begin{assumption}[Conditionally independent classifiers]
\label{ass:conditionally_independent_classifiers}
The different classifiers' predictions $\yh_n^{0:T}$ are conditionally independent
  given the true label $y_n$,
 i.e.,
  the likelihood factorises as
\begin{equation}
\label{eq:conditionally_independent_classifiers_assumption}
p(\yh_n^{0:T}|y_n=k)=\prod_{t=0}^T p(\yh_n^t |y_n=k). 
 \end{equation}%
 \end{assumption}%
In a standard Bayesian setting, this corresponds to the assumption of conditionally independent observations given the parameters; we refer to~\cref{sec:limitations} for further discussion.
The main advantage of Assumption~\ref{ass:conditionally_independent_classifiers} is that the factors $p(\yh_n^t |y_n=k)$ on the RHS of~\eqref{eq:conditionally_independent_classifiers_assumption} have a natural interpretation: these are the (normalised) confusion matrices $\pi^t$
 of the classifiers $C^t$, i.e., we denote by
\begin{equation*}
\label{eq:confusion_matrices}
\pi^t(i,k):=p(\yh^t=i|y=k),
\end{equation*}%
the probability that $C^t$ predicts class $i$ given that the true label is $k$,
which is the same for all $n$; see below and~\cref{sec:limitations} for more details on how we estimate $\pi^t$  in practice.

 \textbf{Posterior \,}
 At every time step $t\geq0$, we can then compute our posterior belief about the true label $y_n$ given model predictions $\yh_{n}^{0:t}$ according to Bayes rule,
\begin{equation}
\label{eq:update_rule}
p(y_n=k|\yh_{n}^{0:t})=
\frac{\pi^t(\yh_n^t,k) p(y_n=k|\yh_{n}^{0:t-1})}
{\sum_{i\in\Ycal}{\pi^t(\yh_n^t,i) p(y_n=i|\yh_{n}^{0:t-1})}}
\end{equation}%
where we have used Assumption~\ref{ass:conditionally_independent_classifiers} to write 
$p(\hat{y}_{n}^t|y_n=k, \yh_{n}^{0:t-1}) = p(\hat{y}_{n}^t|y_n=k) = \pi^t(\yh_n^t,k)$.
The posterior at step $t-1$ acts as  prior for step $t$,
 so we do not have to store all previous predictions.

\textbf{Estimating Confusion Matrices \,}
In practice, 
$\pi^t$ are generally not known and we instead use their (maximum likelihood) estimates $\pih^t$
 from the source distribution.
If the number of classes $K$ is large compared to the amount of labelled source data,\footnote{For example, on ImageNet we have $K=1000$ which would require estimating 1 million parameters.
 } we only estimate the diagonal elements $\pih^t_{kk}$ (i.e., the class-specific accuracies) and set the $K(K-1)$ off-diagonal elements to be constant,
\begin{equation*}
\pih^t(i,k)=\frac{1-\pih^t(k,k)}{K-1} \quad \quad \forall i\neq k, 
\end{equation*}
so that
$\sum_{i=1}^K{\pih^t(i,k) = 1}$
 $\forall k\in\Ycal$.
 We refer to~\cref{sec:limitations} for further discussion on the estimation of $\pi^t$.

\vspace{\negspacepresubsec}
\subsection{Selecting Candidates for Re-evaluation}
\label{sec:selection_strategy}
\vspace{\negspacesubsec}
Given the
 label posteriors
  computed according to~\eqref{eq:update_rule}, we compute the Shannon entropies~\citep{shannon1948mathematical}
\begin{equation*}
\label{eq:entropy}
    S_{n}^{t}=-\sum_{k\in\Ycal} p(y_n=k|\yh_{n}^{0:t}) \log{p(y_n=k|\yh_{n}^{0:t})},
\end{equation*}
which provide a simple measure of
uncertainty in the true label $y_n$ after step $t$.
We then select and re-evaluate the $B^t$ samples with highest posterior label entropy $S_n^t$ to update our beliefs.

\vspace{\negspacepresubsec}
\subsection{Prediction-Update Strategies}
\label{sec:label_update_strategies}
\vspace{\negspacesubsec}
Finally, we need a strategy for deciding whether and how to update the previously stored prediction $l_n^{t-1}$ based on our new beliefs. We consider three such prediction-update strategies.

\textbf{MaxBelief (MB) \,} 
The simplest approach is to always update
 based on the maximum a posteriori 
 belief, i.e., 
$l_n^t := \lh^t_n = \argmax_{k\in\Ycal} \,p(y_n=k|\yh_{n}^{0:t})$. 
We refer to this strategy as MaxBelief (MB).

\textbf{MaxBeliefMinEntropy (MBME) \,}
A slightly more sophisticated approach is to also take the change in posterior entropy into account and only update when it has decreased:
\begin{equation*}
\label{eq:MBME_update}
l_n^t := 
\begin{cases}
\lh^t_n 
  &\mathrm{if} \quad S_n^t<S_n^{t-1}\\
 l_n^{t-1} &\mathrm{otherwise.}
 \end{cases}
\end{equation*}
We refer to this strategy as MaxBeliefMinEntropy (MBME).

\textbf{CostRatio (CR) \,}
So far, we have not taken the assumed larger penalty for negative flips into account.
We therefore now develop a third approach based on asymmetric flip costs.
We denote the cost of a negative flip (\NF) by $c^{\NF}>0$ and that of a positive flip (\PF) by $c^{\PF}<0$. 

We need to decide whether to update the previously stored prediction $l_n^{t-1}$ based on our updated beliefs $p(y_n=k|\yh^{0:t}_n)$.
Denote the MAP label estimate after step $t$
by 
$\lh_n^t= \argmax_{k\in\Ycal} \,p(y_n=k|\yh_{n}^{0:t})$.
If $\lh_n^t=l_n^{t-1}$ there is no reason to change the stored prediction.
Suppose that $\lh_n^t\neq l_n^{t-1}$.
We then need to reason about the (estimated) positive and negative flip probabilities when changing the stored prediction from $l_n^{t-1}$ to $\lh_n^t$.
A positive flip (PF) occurs if $\lh_n^t$ is the correct label (and hence $l_n^{t-1}$ is not), and, vice versa, a negative flip occurs if $l_n^{t-1}$ is correct (and hence $\lh_n^t$ is not):
\begin{equation*}
\ph^\PF_n(l_n^{t-1}\rightarrow \lh_n^t) = p(y_n=\lh_n^t
|\yh^{0:t}_n), \quad \quad \quad\quad 
\ph^\NF_n(l_n^{t-1}\rightarrow \lh_n^t) = p(y_n=l_n^{t-1}
|\yh^{0:t}_n).
\end{equation*}
If neither $l_n^{t-1}$ nor $\lh_n^t$ are the correct label, the flip is inconsequential which we assume incurs zero cost.
The estimated cost of changing the stored prediction from $l_n^{t-1}$ to $\lh_n^t$ is thus:
\begin{equation*}
\ch(l_n^{t-1}\rightarrow \lh_n^t) = c^\NF\ph^\NF_n(l_n^{t-1}\rightarrow \lh_n^t) +c^\PF \ph^\PF_n(l_n^{t-1}\rightarrow \lh_n^t).
\end{equation*}
We only want to change the prediction if $\ch(l_n^{t-1}\rightarrow \lh_n^t)<0$, i.e.,
\begin{equation}
\label{eq:cost_ratio_posterior_ratio}
\frac{\ph^\PF_n(l_n^{t-1}\rightarrow \lh_n^t)}{\ph^\NF_n(l_n^{t-1}\rightarrow \lh_n^t)}
=\frac{p(y_n=\lh_n^t |\yh^{0:t}_n)}{p(y_n=l_n^{t-1} |\yh^{0:t}_n)} 
>-\frac{c^\NF}{c^\PF}
\end{equation}
leading to the following update rule:
\begin{equation*}
\label{eq:CR_update}
l_n^t := 
\begin{cases}
 \lh_n^t,  &\mathrm{if} \quad \lh_n^t=l_n^{t-1}, \\
 \lh_n^t,  &\mathrm{if} \quad \lh_n^t\neq l_n^{t-1} \land \ch(l_n^{t-1}\rightarrow \lh_n^t)<0,\\
 l_n^{t-1} &\mathrm{otherwise.}
 \end{cases}
\end{equation*}
Note that~\eqref{eq:cost_ratio_posterior_ratio} has an intuitive interpretation: we only want to update the currently stored prediction (thus potentially risking a negative flip) if our belief in a different label is larger than that in the current one by a factor exceeding $|c^\NF/c^\PF|$. We therefore refer to this strategy as CostRatio (CR).

\section{Discussion: Extensions and Limitations}
\label{sec:limitations}
We discuss current limitations of our method and propose extensions to address them in future work.

\textbf{Soft vs. Hard Labels \,}
\looseness=-1Our approach presented in~\cref{sec:methods} assumes deterministic classifiers which output hard labels, i.e., only the most likely class.
This allows for maximum flexibility and a wide range of classifier models that can be used in conjunction with this method.
However, our Bayesian framework can easily be adapted to also allow for probabilistic classifiers 
which output soft labels, i.e., vectors of class probabilities.
Since deep neural networks are known to have unreliable uncertainty estimates~\citep{mackay1995bayesian, szegedy2013intriguing, goodfellow2014explaining,nguyen2015deep}, we deliberately choose to work with hard labels.
If, however, well-calibrated probabilistic classifiers are available (and can be scaled to huge data sets), taking this additional information into account will likely lead to more accurate posterior estimates and thus better performance.

\textbf{Assumption of Conditionally-Independent Classifiers \,}
Since the models $\{C^t\}$ are typically trained and developed on the same data and may even build on insights from prior architectures, our assumption of conditionally independent predictions on $\Dcal^\trg$ does likely not hold exactly in practice. 
It should therefore rather be understood as an approximation that enables tractable posterior inference.
Our experiments~(\cref{sec:experiments}) suggest that it is a useful approximation that yields competitive performance. 
Properly incorporating estimated model correlations may yield further improvements.

\textbf{Confusion Matrix Estimates \,}
Unless labelled data from $\PP^\trg$ is
 available, the confusion matrices $\{\pi^t\}$ need to be estimated from $\PP^\src$.
This is only an approximation because they 
may change as a result of $\PP^\src_\Xb\neq\PP^\trg_\Xb$, and taking such shifts into account could yield more accurate posterior estimates.
For this, one may use ideas from the field of \textit{unsupervised domain adaptation}~\citep{pan2009survey, sugiyama2012machine, ganin2015unsupervised}.
One could use an importance-weighting approach~\citep{shimodaira2000improving} to give more weight to points which are representative of $\PP^\trg_\Xb$ when estimating $\pi^t$ from $\PP^\src_{\Xb,Y}$.
As an example, in further experiments in the supplement we studied estimating the
off-diagonal elements using Laplace smoothing~\citep{good1953population, robbins1956empirical}. 

\textbf{Other Selection Strategies \,}
Consider an ambiguous image 
that could be either a zucchini or a cucumber~\citep{beyer2020we}.
Such a sample would have large label entropy and could thus potentially be selected for re-evaluation again and again. 
To overcome this hurdle, one could decompose label uncertainty into epistemic (reducible) and aleatoric (irreducible) uncertainty~\citep{der2009aleatory, kendall2017uncertainties} and only re-evaluate samples with high aleatoric uncertainty, i.e., those
 with high expected information gain~\citep{lindley1956measure}.
Such considerations also play a role in the field of \textit{active learning}~\citep{settles_survey, yan2016} 

\textbf{Growing dataset size \,} Our method is not constrained to fixed dataset sizes and can  accommodate for the addition of new data. New samples can be added at any time using a uniform prior over labels. Given their high initial entropy, they would then be naturally selected for (re-)evaluation first.

\textbf{Adaptive Budgets \,}
Currently, we consider a fixed local budget of $B^t$ re-evaluations at every time step.
A possible extension would be to allow for a global budget of $B^\mathrm{total}$ evaluations spread over all time-steps, i.e., to devise a strategy for deciding whether to (a) keep re-evaluating or (b) save budget for the next better model, potentially 
using techniques 
from reinforcement learning 
\citep{sutton_barto_book}. 

\textbf{On the Cost of ``Neutral" Flips \,}
For simplicity, we have assumed that ``neutral'' flips (i.e., changing a label estimate from an incorrect to a different incorrect one) bear no cost.
However, as motivated in~\cref{sec:introduction}, it is well conceivable that even such neutral flips have a cost due to the potential to disrupt downstream robustness. If this is the case, it can easily be incorporated into our CR update strategy. 

\section{Related Work}
\label{sec:related_work}

Besides the aforementioned connections, our problem setting bears resemblance to several other areas of ML.
In the following, 
we discuss the main differences and commonalities.

\textbf{Backward compatibility\,} \looseness=-1The term was first introduced by \citet{bansal2019updates} in the context of humans making decisions based on an AI's prediction (e.g., medical expert systems or driver supervision in semi-autonomous vehicles). They contextualise that even though an AI's predictive performance might increase overall, \emph{incompatible} predictions in updated models severely hurt overall performance and trust, and propose to penalize negative flips w.r.t.\,an older model when training a newer model.
\citet{yan2020positive} show that with standard training, there can be a significant number of negative flips, even if the  two models only  differ  in their random initializations.  
They then reduce the number of negative flips 
by giving more weight to training points that are correctly classified by the reference model, which they call `positive-congruent training'.
Previous work on backward-compatible learning is concerned with training a \textit{new} model.
Here, we focus on updating the stored predictions rather than updating the stored models.
This makes our approach more generally applicable and complements the use-cases of backward-compatible learning.
Backward compatibility was being further studied empirically by \citet{srivastava2020empirical} who emphasize that this also causes problems for large multi-component AI systems. They propose two key metrics to characterize  backward compatibility: (i) Backward Trust Compatibility (BTC), first mentioned in \cite{bansal2019updates}, measuring the fraction of predictions that are still predicted correctly after a model update; and (ii) Backward Error Compatibility (BEC), which corresponds to the probability that an incorrect prediction after an update is not new.

\textbf{Ensemble Learning \,}
Ensemble methods aim to combine several ML models into a single model with higher performance than each of the individual models. Common techniques are boosting~\citep{adaboost}, bagging~\citep{bagging}, Bayesian model averaging~\citep{hoeting1999}. 
Our approach falls into the latter category. We compute the posterior probability~\eqref{eq:update_rule} in the same way as the 
well-known Naive Bayes combiner~\citep{kuncheva2014}.
The classifier corresponding to our MB strategy goes back to at least~\citet{nitzan1982} and has been thoroughly analyzed~\citep{berend2015}. There are also Bayesian techniques that avoid Assumption~\ref{ass:conditionally_independent_classifiers}, but these either make some parametric assumptions~\citep{kim2012bayesian} or assume a very special form of dependence~\citep{boland1989}.

\section{Experiments}
\label{sec:experiments}
We now evaluate our Bayesian approach to the Prediction Update Problem against different baselines using the task of image classification as a case study.

\subsection{Experimental Setup}
\label{sec:experimental_setup}

\textbf{Data Sets \,}
\looseness=-1 
We use the three widely accepted benchmark data sets ImageNet1K~\citep{deng2009imagenet} (1K classes, 50k validation set), ObjectNet~\citep{barbu2019objectnet} (313 classes, 50k validation set) and CIFAR-10~\citep{krizhevsky2009learning} (10 classes, 10k validation set).
To imitate our assumed setting of deploying pre-trained models to an unlabelled target data set, we only use the corresponding validation sets as $\Dcal^\trg$. 
The ground truth labels are only used post-hoc to compute performance metrics and are not seen during the $T$ update steps.
Of the 313 classes in ObjectNet, 113 are shared with ImageNet, corresponding to a subset of 18,547 images.
ObjectNet 
images
exhibit more realistic variations than those in ImageNet.
It only has a test set and thus constitutes a challenging transfer scenario for object recognition models.
We deploy ImageNet-pretrained models both on ImageNet and on the above subset of ObjectNet, thus simulating the cases that the source and target distributions are the same or different, respectively.
For the former, we split the ImageNet validation set in half and use one half to estimate 
$\pi^t$ and the other as $\Dcal^\trg$. For the latter, we estimate $\pi^t$ from the full ImageNet validation set and evaluate on ObjectNet.

\textbf{Models \& Architectures \,}
\looseness=-1 
To emulate the setting of sequentially improving classifiers arriving over time, we use the following 17 models and architectures with many of them setting a new ``state-of-the-art'' on ImageNet at the time they were first introduced:
AlexNet~\citep{krizhevsky2012imagenet}; VGG-11, 13, 16, and 19~\citep{simonyan2014very}; ResNet-18, 34, 50, 101, and 152~\citep{he2016deep}; SqueezeNet~\citep{iandola2016squeezenet}; GoogLeNet~\citep{szegedy2015going}; InceptionV3~\citep{szegedy2016rethinking}; MobileNetV2 \citep{sandler2018mobilenetv2} 
DenseNet-121 and 169~\citep{huang2017densely}, and ResNeXt-101 32x8d~\citep{xie2017aggregated}. 
For ease of reproducibility, we use pre-trained models from the torchvision model zoo~\citep{paszke2019pytorch} and~\citep{huyphan20214431043}.

\textbf{Performance Metrics \,}
\looseness=-1 
Recall that our goal is to: (i) improve overall accuracy, (ii) avoid negative flips, and (iii) use as few re-evaluations as possible.
To assess these different aspects, we report the following metrics:
  (i) \textit{final accuracy} of the stored predictions 
  (\textbf{Acc}) and
\textit{accuracy improvement} over the initial accuracy of $C^0$~(\textbf{$\Delta$Acc});
(ii) the \textit{cumulative number of negative flips} from time $t=0$ to $T$ ($\Sigma$~\textbf{\NF}),
 the \textit{average negative flip rate} experienced per iteration, i.e., $\frac{\Sigma~\NF}{N\cdot T}$ (\textbf{\NFR}),
 and the ratio of accumulated positive to negative flips (\textbf{\PF \,/ \NF});
 (iii) the evaluation budget available to each strategy as percentage of the data set size, i.e., a budget of 10 means that 10\% of all samples can be re-evaluated at each time step: $B^t=0.1N, \forall t$;
 finally, we measure the connective backward compatibility between (i) and (ii) via Backward Trust Compatibility (\textbf{BTC}) and Backward Error Compatibility (\textbf{BEC}) \cite{bansal2019updates}.

\begin{figure}[]
    \includegraphics[width=0.55\linewidth]{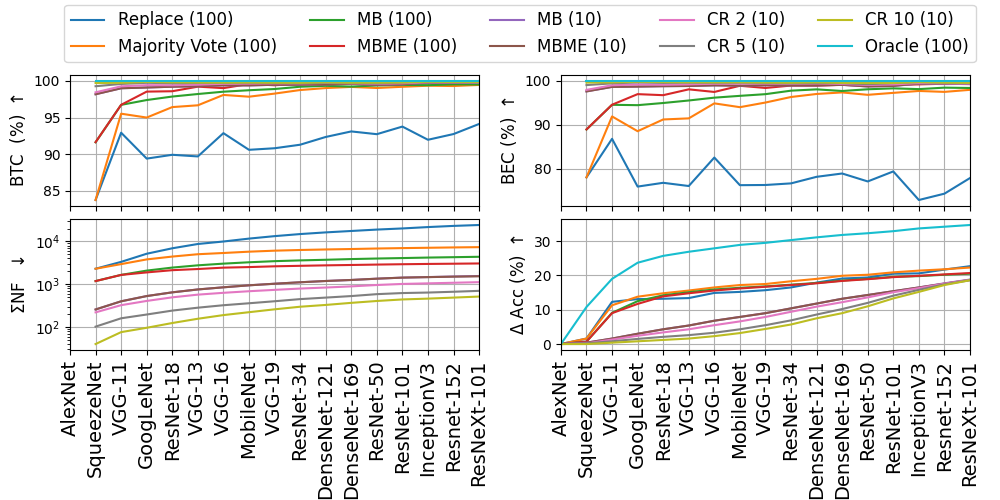}
    \includegraphics[width=0.44\linewidth]{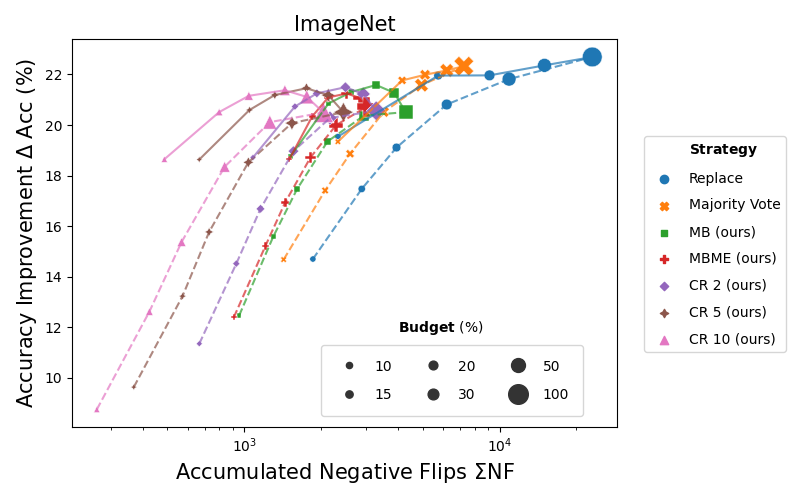}
    \vspace{2em}
    \caption{\small
     \textbf{Left:} Temporal evolution for ImageNet $\rightarrow$ ImageNet over $T=16$ prediction-update steps for a subset of strategies and budgets.  \textbf{Right:} Comparison of prediction-update strategies across different budgets after $T=16$ prediction-update steps. Dashed lines correspond to the ablation using a random selection strategy. }
\label{fig:imagenet_improving}
\end{figure}

\textbf{Baselines and Oracle \,} 
\looseness=-1 
We compare our method against two baselines:
(i) \textbf{Replace}
always updates the stored prediction with that predicted by the most recent classifier (a.k.a.~\textit{backfilling});
(ii) \textbf{Majority Vote} takes into account previous model predictions and updates the stored prediction according to the majority prediction---in case of a tie, the prediction of the most recent classifier is chosen.
For reference, we also compare our method against an \textbf{Oracle}, which performs a prediction update if and only if this would lead to a positive flip; it thus incurs zero negative flips by definition (knowing the ground truth label). We emphasize that, in practice, we do not have that information in our setting.

\textbf{Selection- and prediction-update Strategies \,}
For all methods, we select $B^t\leq N$ samples
using the posterior label entropy selection strategy from~\cref{sec:selection_strategy}, thus having baselines incorporating some elements of our method, but also compare with randomly selecting samples for re-evaluation.
We use the prediction-update strategies MB, MBME and CR from~\cref{sec:label_update_strategies} and consider cost ratios of $|c^\NF/c^\PF| \in \{2, 5, 10\}$ for the latter (e.g., CR 2). 

\subsection{Results for ImageNet $\rightarrow$ ImageNet}
\label{sec:results_imagenet}

In \cref{fig:imagenet_improving} (left), we show the temporal evolution of backwards compatibility scores, negative flips and accuracy gains for prediction-updates on the ImageNet validation set for a subset of strategies and budgets. 
A complete account of final performances with additional metrics is shown in \cref{tab:imagenet_objectnet_improving} (left).

\begin{table}[t!]
\small
\caption{\small
Results for ImageNet $\rightarrow$ ImageNet (left) and ImageNet $\rightarrow$ ObjectNet (right): all metrics refer to final performance for the improving model sequence from~\cref{fig:imagenet_improving} and~\cref{fig:objectnet_improving} respectively. The character \textbf{E} or \textbf{R} in front of the strategy indicates that selection for re-evaluation is based on the entropy criterion or sampled randomly.
}
\label{tab:imagenet_objectnet_improving}
\resizebox{\linewidth}{!}{
\begin{tabular}{l|l|rrrrrrr}
\toprule
 & \textbf{Strategy}   &   Avg. \textbf{BTC} $\uparrow$&    Avg. \textbf{BEC} $\uparrow$   &   \textbf{Acc} (\%) $\uparrow$ &   $\Delta$\textbf{Acc} (\%) $\uparrow$&   $\Sigma$ \textbf{\NF} $\downarrow$&   \textbf{\NFR}\, (\%) $\downarrow$&  \textbf{ \PF}\, / \textbf{\NF} $\uparrow$\\
\midrule
 & Oracle  &    100      & 100     &       91.2 &        34.7 &       0   &  0    &       -     \\
  \midrule
\parbox[c]{2mm}{\multirow{7}{*}{\rotatebox[origin=c]{90}{Budget = 100\%}}}  & Replace   & 91.37 & 77.71   &       \textbf{79.2} &        \textbf{22.7} &   24214   &  6.05 &       1.2 \\
& Majority Vote  & 97.18 & 93.95 &       78.9 &        22.3 &    7352   &  1.84 &         1.8 \\
& MB      & 98.32 & 96.45       &       77.1 &        20.5 &    4378   &  1.09 &         2.2\\
& MBME    & 98.78 & 97.69     &       77.3 &        20.7 &    3057   &  0.76 &         2.7 \\
& CR 2     & 98.72 & 97.19      &       77.1 &        20.6 &    3368   &  0.84 &         2.5\\
& CR 5        & 99.06 & 97.82      &       77.1 &        20.5 &    2520   &  0.63 &         3\\
& CR 10     & \textbf{99.22} &\textbf{98.15}     &       77   &        20.5 &    \textbf{2112}   &  \textbf{0.53} &        \textbf{3.4}\\
 \midrule
\parbox[t]{2mm}{\multirow{9}{*}{\rotatebox[origin=c]{90}{Budget = 30\%}}}  & R:Replace    & 97.56 & 94.53     &       77.4 &        20.8 &    6546.4 &  1.64 &         1.8\\
& R:Majority Vote   & 98.6  & 97.18 &       77.1 &        20.5 &    3616.4 &  0.9  &         2.4\\
 &E:Replace    & 96.53 & 91.01      &       \textbf{78.5} &        \textbf{22}   &      9708 &  2.43 &         1.6\\
& E:Majority Vote   & 98.03 & 95.63 &       \textbf{78.5} &        \textbf{22}   &      5232 &  1.31 &         2.1\ \\
& E:MB        & 98.71 & 97.25      &       78.1 &        21.6 &    3375   &  0.84 &         2.6 \\
& E:MBME     & 98.98 & 98.04       &       77.8 &        21.2 &    2577   &  0.64 &         3.1 \\
& E:CR 2   & 99.02 & 97.86         &       78   &        21.5 &    2578   &  0.64 &         3.1 \\
& E:CR 5      & 99.32 & 98.43       &       78   &        21.5 &    1831   &  0.46 &         3.9\\
& E:CR 10    & \textbf{99.44} & \textbf{98.67}        &       77.9 &        21.4 &    \textbf{1517}   &  \textbf{0.38} &         \textbf{4.5}\\
 \midrule
\parbox[t]{2mm}{\multirow{9}{*}{\rotatebox[origin=c]{90}{Budget = 10\%}}}   & R:Replace   & 99.22 & 98.63       &       71.3 &        14.7 &    1958.4 &  0.49 &         2.9\\
& R:Majority Vote  & 99.4  & 98.98   &       71.2 &        14.7 &    1481.4 &  0.37 &         3.5\\
& E:Replace      & 99.04 & 98.12     &        \textbf{76.1} &         \textbf{19.5} &      2468 &  0.62 &         3 \\
& E:Majority Vote   & 99.06 & 98.18  &       75.9 &        19.3 &      2417 &  0.6  &         3  \\
& E:MB       & 99.38 & 98.89      &       75.3 &        18.8 &    1557   &  0.39 &         4   \\
& E:MBME      & 99.38 & 98.92        &       75.2 &        18.7 &    1533   &  0.38 &         4 \\
& E:CR 2     & 99.55 & 99.22       &       75.3 &        18.7 &    1118   &  0.28 &         5.2 \\
& E:CR 5     & 99.72 & 99.51      &       75.2 &        18.6 &     700   &  0.18 &         7.7 \\
& E:CR 10     &  \textbf{99.79} &  \textbf{99.64}      &       75.2 &        18.6 &      \textbf{515}   &   \textbf{0.13} &         \textbf{10.1}\\
\bottomrule 
\end{tabular}
\hspace*{1.0cm}
\begin{tabular}{l|l|rrrrrrr}
\toprule
& \textbf{Strategy}      &   Avg. \textbf{BTC} $\uparrow$&    Avg. \textbf{BEC} $\uparrow$     &   \textbf{Acc} (\%) $\uparrow$&   $\Delta$\textbf{Acc} (\%) $\uparrow$&   $\Sigma$ \textbf{\NF} $\downarrow$&   \textbf{\NFR}\, (\%) $\downarrow$&  \textbf{ \PF}\, / \textbf{\NF} $\uparrow$\\
 \midrule
&  Oracle  & 100      & 100      &       50.5 &        42.6 &       0   &  0    &       -\\
  \midrule
\parbox[t]{2mm}{\multirow{7}{*}{\rotatebox[origin=c]{90}{Budget = 100\%}}}  & Replace   & 72.65 & 92.61     &      \textbf{31.9} &         \textbf{24}   &   16669   &  5.62 &         1.3\\
& Majority Vote  & 89.99 & 98.02 &       29.6 &        21.6 &    4690   &  1.58 &         1.9 \\
& MB        & 94.46 & 98.96    &       29.1 &        21.2 &    2477   &  0.83 &         2.6 \\
& MBME   & 95.86 & 99.34    &       28.6 &        20.6 &    1599   &  0.54 &         3.4  \\
 &CR 2    & 95.92 & 99.21     &       29   &        21   &    1876   &  0.63 &         3.1 \\
& CR 5      & 97.18 & 99.41    &       28.8 &         20.8&    1372   &  0.46 &         3.8  \\
& CR 10     &  \textbf{97.82} &  \textbf{99.54}     &       28.7 &         20.8 &     \textbf{1084}   &  \textbf{0.37} &   \textbf{4.6} \\
 \midrule
\parbox[t]{2mm}{\multirow{9}{*}{\rotatebox[origin=c]{90}{Budget = 30\%}}}  & R:Replace       & 92.19 & 98.26  &        \textbf{29}   &         \textbf{21}   &    4070.6 &  1.37 &         2  \\
& R:Majority Vote & 94.91 & 99.02   &       27.3 &        19.4 &    2346.6 &  0.79 &         2.5 \\
 &E:Replace     & 91.75 & 98.14    &        \textbf{29}   &         \textbf{21}   &      4316 &  1.45 &         1.9 \\
& E:Majority Vote   & 93.54 & 98.76 &       28.2 &        20.3 &      2970 &  1    &         2.3 \\
& E:MB         & 96.24 & 99.35     &       27.8 &        19.9 &    1565   &  0.53 &         3.4\\
& E:MBME    & 96.64 & 99.48       &       26.9 &        18.9 &    1280   &  0.43 &         3.7 \\
 &E:CR 2      & 97.42 & 99.55     &       27.7 &        19.7 &    1074   &  0.36 &         4.4 \\
 &E:CR 5       & 98.43 & 99.71      &       27.4 &        19.4 &     689   &  0.23 &         6.2\\
 &E:CR 10       &  \textbf{98.91} &  \textbf{99.79}    &       27.1 &        19.2 &      \textbf{504}   &   \textbf{0.17} &          \textbf{8.1} \\
 \midrule
\parbox[t]{2mm}{\multirow{9}{*}{\rotatebox[origin=c]{90}{Budget = 10\%}}}   & R:Replace     & 97.49 & 99.6     &       22.1 &        14.2 &     996.6 &  0.34 &         3.6 \\
& R:Majority Vote  & 97.86 & 99.68 &       21.6 &        13.6 &     808.8 &  0.27 &         4.1 \\
 & E:Replace   & 97.5  & 99.6     &        \textbf{23.7} &         \textbf{15.7} &       996 &  0.34 &         3.9  \\
& E:Majority Vote  & 97.5  & 99.6  &        \textbf{23.7} &         \textbf{15.7} &       996 &  0.34 &         3.9  \\
& E:MB         & 98.08 & 99.72     &       22.7 &        14.8 &     696   &  0.23 &         4.9\\
& E:MBME       & 98.08 & 99.72     &       22.7 &        14.8 &     696   &  0.23 &         4.9 \\
& E:CR 2      & 98.68 & 99.83     &       20.7 &        12.8 &     427   &  0.14 &         6.6 \\
& E:CR 5        & 99.32 & 99.92    &       18.2 &        10.3 &     197   &  0.07 &        10.7 \\
& E:CR 10     &  \textbf{99.57} &  \textbf{99.95}      &       17.2 &         9.2 &      \textbf{122}   &   \textbf{0.04} &         \textbf{15}  \\
\bottomrule
\end{tabular}}
\end{table}

For the evolution of \textbf{$\Delta$Acc} in~\cref{fig:imagenet_improving} (left), we observe that, unsurprisingly, strategies with 100\% budget experience a more rapid gain in accuracy than those with 10\%. 
Among the budget-constrained strategies, the CR strategy with large cost ratio shows the slowest increase, which makes sense as it requires a substantial change in posterior belief for updating a stored prediction and is thus more conservative. 
Interestingly, however, the \textit{final} accuracies  only differ marginally across both strategies and budgets which is also apparent from the minor differences in the \textbf{$\Delta$Acc} column of~\cref{tab:imagenet_objectnet_improving}.
For the evolution of  $\Sigma$ \textbf{\NF} in~\cref{fig:imagenet_improving} (right), we observe a clear separation of strategies with a natural ordering from least conservative (Replace) to most conservative (CR 10).
These relative differences stay mostly constant over time as NFs appear to accumulate approximately linearly (note the log-scale). 
We find roughly an order of magnitude difference in $\Sigma$ \textbf{\NF} between the best non-probabilistic baseline (Majority Vote) and the best Bayesian method (CR 10). 
Especially for small budgets of up to 30\%, our Bayesian strategies 
clearly dominate the non-probabilistic baselines both in terms of accuracy and flip metrics, as can be seen from~\cref{tab:imagenet_objectnet_improving} and~\cref{fig:imagenet_improving} (right).
Moreover, the CR strategy appears to provide control over the number of negative flips via its cost-ratio hyperparameter without adversely affecting final accuracy across a range of budgets, as already observed for a budget of 10\% in~\cref{fig:imagenet_improving}.
Interestingly, the update rules 
seem to be optimal when evaluating on less than 100\% budget. 
We attribute this to posterior approximation errors on ImageNet, which is being supported by extensive ablations in the supplement.
Regarding backward compatibility (our ultimate goal), we find that \textbf{BTC} \emph{and} \textbf{BEC} scores reliably outperform the baselines across all budgets. In particular, the CR 10 strategy seems to be especially suitable with scores close to 100\%, i.e., oracle performance.

\textbf{Summary \,}
\looseness=-1 
Our method appears to successfully fulfill the three desiderata for backward-compatible prediction-updates in an i.i.d.~setting.
In particular, our CR strategy seems like the most promising candidate to (i) maintain high accuracy gains and (ii) introduce very few negative flips, when (iii) given only a small compute budget for re-evaluations.

\subsection{Results for ImageNet $\rightarrow$ ObjectNet}
\label{sec:results_objectnet}

\begin{figure}[]
    \includegraphics[width=0.55\linewidth]{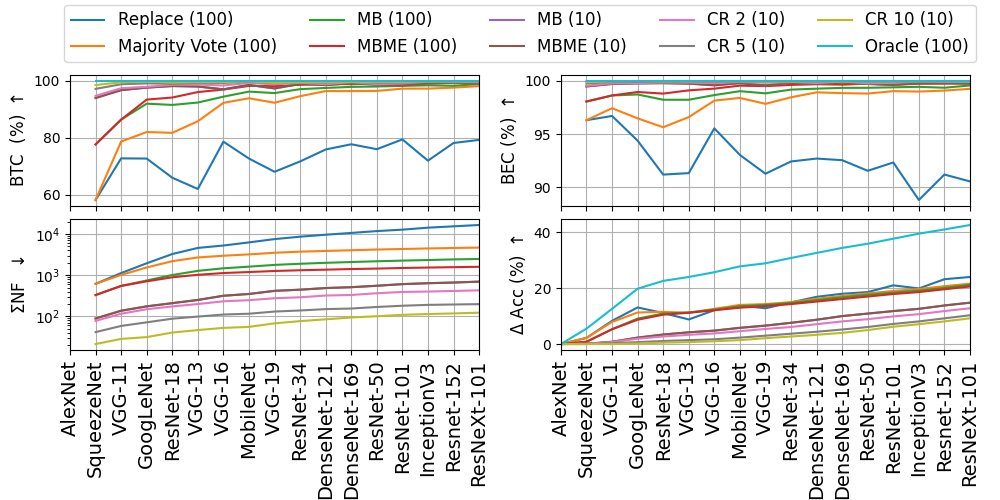}
    \includegraphics[width=0.44\linewidth]{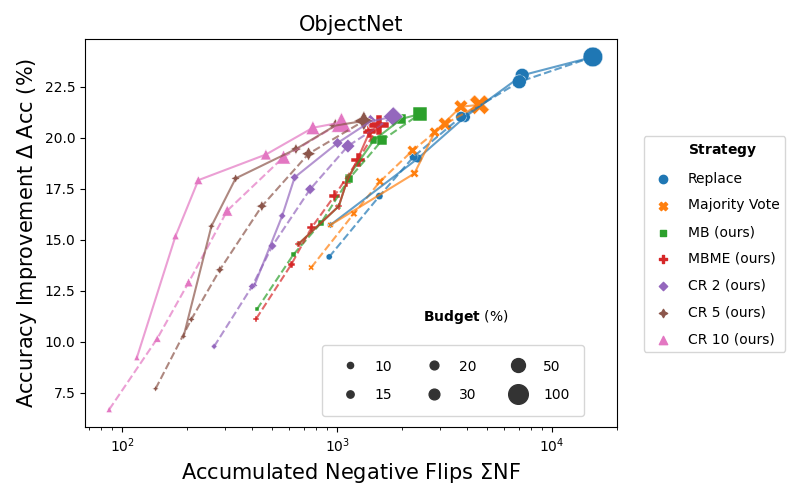}
    \vspace{-0.9em}
    \caption{\small
     \textbf{Left:}  Temporal evolution for ImageNet $\rightarrow$ ObjectNet over $T=16$ prediction-update steps for a subset of strategies and budgets. \textbf{Right:} Comparison of all strategies after $T=16$ on ObjectNet. }
\label{fig:objectnet_improving}
\end{figure}

\looseness=-1 Results for prediction-updates on ObjectNet are presented (similarly to~\cref{sec:results_imagenet}) in~\cref{fig:objectnet_improving} and \cref{tab:imagenet_objectnet_improving} (right). 
This transfer setting constitutes a much more challenging task.
Nevertheless, we observe very similar behaviour to that discussed in~\cref{sec:results_imagenet} and thus only point out the main differences.
First, we note that - despite the smaller target data set - the difference in negative flips across different strategies and budgets is even larger on ObjectNet.
For example, we observe a reduction in~$\Sigma$ \textbf{\NF} of more than two orders of magnitude between Replace (100) and CR (10), and about one order
 when comparing the two 
 for the same budget.
At the same time, differences in accuracy across strategies 
are also slightly more pronounced, especially for the smallest budget of 10\%. 
Here, the more conservative CR strategies yield lower accuracy gains while MB and MBME maintain competetive accuracy gains. Our strategies are again clearly dominating in terms of backward compatibility w.r.t.\,BTC and BEC. 
We remark that these results are agnostic to any potential differences in the label space: they are based on a posterior over all 1000 ImageNet classes whereas ObjectNet only contains a subset of 113 of these classes.
\clearpage
\subsection{Further Experiments and Ablations}
\label{subsec:further_experiments_and_ablations}

\begin{wrapfigure}{r}{6.5cm}
\vspace{-1em}
    \includegraphics[width=\linewidth]{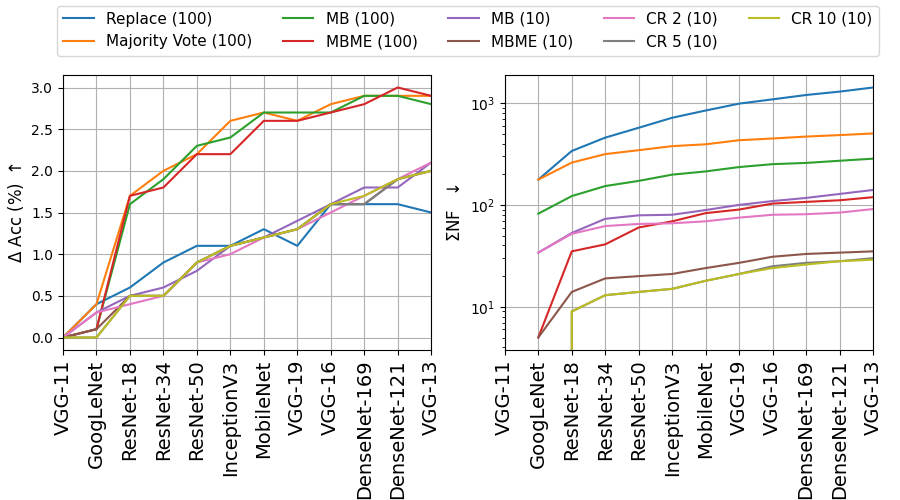}
    \vspace{-2em}
    \caption{\small
    Under smaller accuracy gains (experiments on CIFAR-10), we outperform the Replace baseline w.r.t.\,desideratum (i).
}
\vspace{-1em}
\label{fig:cifar10_improving}
\end{wrapfigure}

\textbf{Results for CIFAR-10}
In \cref{fig:cifar10_improving} we show the corresponding results on the CIFAR-10 dataset. 
Here, pre-trained models exhibit a higher level of accuracy ($\approx 93 - 95\%$)
and we thus emulate an arguably more realistic scenario with models being released more frequently, thus with smaller accuracy differences.
Our method shows very similar trends 
as we have worked out on ImageNet and ObjectNet. 
Interestingly, there is one novel characteristic: 
due to the presumably less steep increase in accuracy from one model to its successor and fewer class categories, we can form very accurate posterior beliefs which results in accuracy gains of all our methods that even outperform the accuracy-optimizing baselines concerning desideratum (i).

\textbf{Role of the Selection Strategy \,}
We also conduct a comparison between our 
entropy selection and the random baseline for all our methods across a range of budgets - see dashed lines in scatter plots. 
We find that random selection leads to substantially smaller accuracy gains, but also to fewer negative flips which is intuitive since random selection more often chooses ``easy'' samples for re-evaluation.

\textbf{Robustness to Random \& Adversary Model Sequences \,}
We have assumed that the models $C^t$ are \textit{improving} over time.
We thus also consider the scenario where $C^t$ arrive in a \textit{random} or \textit{adversarial} (i.e., strictly deteriorating) order. For the random order, we find that our methods - unlike, e.g., the replace strategy - achieve strict increases in accuracy while introducing much fewer negative flips.
Even in the adversarial case, our methods improve accuracy during the initial steps with much fewer negative flips over the entire history.
These findings suggest robustness of our approach to situations where an ordering by performance of $C^t$ may not be available.
 
\textbf{Reducing re-evaluations matters at scale \,} 
The re-evaluations of a sample using deep neural network based models clearly dominate computational cost as compared to our method.
As an example, the forward pass using the public ImageNet PyTorch models takes up to 550 (biggest VGG and ResNets) times longer than the unoptimized implementation of our method backbone. For very large data sets and with new models generally increasing in size, reducing the inference budget B is of crucial importance, emphasizing the relevance of  desideratum 3.

We refer to the supplement for more detailed results and discussion of the above experiments.

\section{Conclusion}
\label{sec:conclusion}
The Prediction Update Problem appears frequently in practice and can take different forms. It relates to many different subfields of ML that we have discussed in~\cref{sec:limitations} and~\cref{sec:related_work}, and there are interesting extensions (structured prediction, adaptive budgets) and improvements (modeling data set structure, across-dataset similarity, domain adaptation, calibration techniques) that need to be worked out. In this work, we have studied the classification case and proposed a Bayesian update rule based on simple assumptions. Empirically, we find improvements along the dimensions we set out to achieve, and we hope that progress on this problem will democratize ML usage even further as it lowers the bar for benefitting from the tremendous progress in model design seen over the last years.

\section{Acknowledgements}
\label{sec:acknowledgements}
The authors would like to thank Matthias Bethge, Thomas Brox, Chris Russell, Karsten Roth, Nasim Rahaman, Alex Smola, Yuanjun Xiong and Stefano Soatto for helpful discussions and feedback. 

\bibliographystyle{plainnat}
\bibliography{references}

\begin{thebibliography}{52}
\providecommand{\natexlab}[1]{#1}
\providecommand{\url}[1]{\texttt{#1}}
\expandafter\ifx\csname urlstyle\endcsname\relax
  \providecommand{\doi}[1]{doi: #1}\else
  \providecommand{\doi}{doi: \begingroup \urlstyle{rm}\Url}\fi

\bibitem[Bansal et~al.(2019)Bansal, Nushi, Kamar, Weld, Lasecki, and
  Horvitz]{bansal2019updates}
Gagan Bansal, Besmira Nushi, Ece Kamar, Daniel~S Weld, Walter~S Lasecki, and
  Eric Horvitz.
\newblock Updates in human-{AI} teams: Understanding and addressing the
  performance/compatibility tradeoff.
\newblock In \emph{Proceedings of the AAAI Conference on Artificial
  Intelligence}, volume~33, 2019.

\bibitem[Barbu et~al.(2019)Barbu, Mayo, Alverio, Luo, Wang, Gutfreund,
  Tenenbaum, and Katz]{barbu2019objectnet}
Andrei Barbu, David Mayo, Julian Alverio, William Luo, Christopher Wang, Dan
  Gutfreund, Josh Tenenbaum, and Boris Katz.
\newblock Objectnet: A large-scale bias-controlled dataset for pushing the
  limits of object recognition models.
\newblock \emph{Advances in neural information processing systems}, 32, 2019.

\bibitem[Berend and Kontorovich(2015)]{berend2015}
Daniel Berend and Aryeh Kontorovich.
\newblock A finite sample analysis of the naive bayes classifier.
\newblock \emph{J. Mach. Learn. Res.}, 16\penalty0 (1), 2015.

\bibitem[Beyer et~al.(2020)Beyer, H{\'e}naff, Kolesnikov, Zhai, and
  Oord]{beyer2020we}
Lucas Beyer, Olivier~J H{\'e}naff, Alexander Kolesnikov, Xiaohua Zhai, and
  A{\"a}ron van~den Oord.
\newblock Are we done with imagenet?
\newblock \emph{arXiv preprint arXiv:2006.07159}, 2020.

\bibitem[Boland et~al.(1989)Boland, Proschan, and Tong]{boland1989}
Philip~J Boland, Frank Proschan, and Yung~Liang Tong.
\newblock Modelling dependence in simple and indirect majority systems.
\newblock \emph{Journal of Applied Probability}, 1989.

\bibitem[Breiman(1996)]{bagging}
Leo Breiman.
\newblock Bagging predictors.
\newblock \emph{Machine learning}, 24\penalty0 (2), 1996.

\bibitem[Deng et~al.(2009)Deng, Dong, Socher, Li, Li, and
  Fei-Fei]{deng2009imagenet}
Jia Deng, Wei Dong, Richard Socher, Li-Jia Li, Kai Li, and Li~Fei-Fei.
\newblock Imagenet: A large-scale hierarchical image database.
\newblock In \emph{2009 IEEE conference on computer vision and pattern
  recognition}. Ieee, 2009.

\bibitem[Der~Kiureghian and Ditlevsen(2009)]{der2009aleatory}
Armen Der~Kiureghian and Ove Ditlevsen.
\newblock Aleatory or epistemic? does it matter?
\newblock \emph{Structural safety}, 31\penalty0 (2), 2009.

\bibitem[Dosovitskiy et~al.(2021)Dosovitskiy, Beyer, Kolesnikov, Weissenborn,
  Zhai, Unterthiner, Dehghani, Minderer, Heigold, Gelly,
  et~al.]{dosovitskiy2020image}
Alexey Dosovitskiy, Lucas Beyer, Alexander Kolesnikov, Dirk Weissenborn,
  Xiaohua Zhai, Thomas Unterthiner, Mostafa Dehghani, Matthias Minderer, Georg
  Heigold, Sylvain Gelly, et~al.
\newblock An image is worth 16x16 words: Transformers for image recognition at
  scale.
\newblock 2021.

\bibitem[Foret et~al.(2021)Foret, Kleiner, Mobahi, and
  Neyshabur]{foret2020sharpness}
Pierre Foret, Ariel Kleiner, Hossein Mobahi, and Behnam Neyshabur.
\newblock Sharpness-aware minimization for efficiently improving
  generalization.
\newblock 2021.

\bibitem[Freund and Schapire(1997)]{adaboost}
Yoav Freund and Robert~E Schapire.
\newblock A decision-theoretic generalization of on-line learning and an
  application to boosting.
\newblock \emph{Journal of computer and system sciences}, 55\penalty0 (1),
  1997.

\bibitem[Ganin and Lempitsky(2015)]{ganin2015unsupervised}
Yaroslav Ganin and Victor Lempitsky.
\newblock Unsupervised domain adaptation by backpropagation.
\newblock In \emph{International conference on machine learning}. PMLR, 2015.

\bibitem[Good(1953)]{good1953population}
Irving~J Good.
\newblock The population frequencies of species and the estimation of
  population parameters.
\newblock \emph{Biometrika}, 40\penalty0 (3-4), 1953.

\bibitem[Goodfellow et~al.(2015)Goodfellow, Shlens, and
  Szegedy]{goodfellow2014explaining}
Ian~J Goodfellow, Jonathon Shlens, and Christian Szegedy.
\newblock Explaining and harnessing adversarial examples.
\newblock \emph{International Conference on Learning Representations}, 2015.

\bibitem[He et~al.(2016)He, Zhang, Ren, and Sun]{he2016deep}
Kaiming He, Xiangyu Zhang, Shaoqing Ren, and Jian Sun.
\newblock Deep residual learning for image recognition.
\newblock In \emph{Proceedings of the IEEE conference on computer vision and
  pattern recognition}, 2016.

\bibitem[Hoeting et~al.(1999)Hoeting, Madigan, Raftery, and
  Volinsky]{hoeting1999}
Jennifer~A Hoeting, David Madigan, Adrian~E Raftery, and Chris~T Volinsky.
\newblock Bayesian model averaging: a tutorial.
\newblock \emph{Statistical science}, 1999.

\bibitem[Huang et~al.(2017)Huang, Liu, Van Der~Maaten, and
  Weinberger]{huang2017densely}
Gao Huang, Zhuang Liu, Laurens Van Der~Maaten, and Kilian~Q Weinberger.
\newblock Densely connected convolutional networks.
\newblock In \emph{Proceedings of the IEEE conference on computer vision and
  pattern recognition}, 2017.

\bibitem[Iandola et~al.(2016)Iandola, Han, Moskewicz, Ashraf, Dally, and
  Keutzer]{iandola2016squeezenet}
Forrest~N Iandola, Song Han, Matthew~W Moskewicz, Khalid Ashraf, William~J
  Dally, and Kurt Keutzer.
\newblock Squeezenet: Alexnet-level accuracy with 50x fewer parameters and $<$
  0.5 mb model size.
\newblock \emph{arXiv preprint arXiv:1602.07360}, 2016.

\bibitem[Kendall and Gal(2017)]{kendall2017uncertainties}
Alex Kendall and Yarin Gal.
\newblock What uncertainties do we need in {B}ayesian deep learning for
  computer vision?
\newblock In \emph{Proceedings of the 31st International Conference on Neural
  Information Processing Systems}, 2017.

\bibitem[Kim and Ghahramani(2012)]{kim2012bayesian}
Hyun-Chul Kim and Zoubin Ghahramani.
\newblock {B}ayesian classifier combination.
\newblock In \emph{Artificial Intelligence and Statistics}, 2012.

\bibitem[Krizhevsky et~al.(2009)Krizhevsky, Hinton,
  et~al.]{krizhevsky2009learning}
Alex Krizhevsky, Geoffrey Hinton, et~al.
\newblock Learning multiple layers of features from tiny images.
\newblock 2009.

\bibitem[Krizhevsky et~al.(2012)Krizhevsky, Sutskever, and
  Hinton]{krizhevsky2012imagenet}
Alex Krizhevsky, Ilya Sutskever, and Geoffrey~E Hinton.
\newblock Imagenet classification with deep convolutional neural networks.
\newblock \emph{Advances in neural information processing systems}, 2012.

\bibitem[Kuncheva(2014)]{kuncheva2014}
Ludmila~I Kuncheva.
\newblock \emph{Combining pattern classifiers: methods and algorithms}.
\newblock John Wiley \& Sons, 2014.

\bibitem[Lindley(1956)]{lindley1956measure}
Dennis~V Lindley.
\newblock On a measure of the information provided by an experiment.
\newblock \emph{The Annals of Mathematical Statistics}, 1956.

\bibitem[MacKay(1995)]{mackay1995bayesian}
David~JC MacKay.
\newblock {B}ayesian neural networks and density networks.
\newblock \emph{Nuclear Instruments and Methods in Physics Research Section A:
  Accelerators, Spectrometers, Detectors and Associated Equipment},
  354\penalty0 (1), 1995.

\bibitem[Nguyen et~al.(2015)Nguyen, Yosinski, and Clune]{nguyen2015deep}
Anh Nguyen, Jason Yosinski, and Jeff Clune.
\newblock Deep neural networks are easily fooled: High confidence predictions
  for unrecognizable images.
\newblock In \emph{Proceedings of the IEEE conference on computer vision and
  pattern recognition}, 2015.

\bibitem[Nitzan and Paroush(1982)]{nitzan1982}
Shmuel Nitzan and Jacob Paroush.
\newblock Optimal decision rules in uncertain dichotomous choice situations.
\newblock \emph{International Economic Review}, 1982.

\bibitem[Pan and Yang(2009)]{pan2009survey}
Sinno~Jialin Pan and Qiang Yang.
\newblock A survey on transfer learning.
\newblock \emph{IEEE Transactions on knowledge and data engineering},
  22\penalty0 (10), 2009.

\bibitem[Paszke et~al.(2019)Paszke, Gross, Massa, Lerer, Bradbury, Chanan,
  Killeen, Lin, Gimelshein, Antiga, et~al.]{paszke2019pytorch}
Adam Paszke, Sam Gross, Francisco Massa, Adam Lerer, James Bradbury, Gregory
  Chanan, Trevor Killeen, Zeming Lin, Natalia Gimelshein, Luca Antiga, et~al.
\newblock Pytorch: An imperative style, high-performance deep learning library.
\newblock \emph{Advances in Neural Information Processing Systems}, 32, 2019.

\bibitem[Pham et~al.(2020)Pham, Xie, Dai, and Le]{pham2020meta}
Hieu Pham, Qizhe Xie, Zihang Dai, and Quoc~V Le.
\newblock Meta pseudo labels.
\newblock \emph{arXiv preprint arXiv:2003.10580}, 2020.

\bibitem[Phan(2021)]{huyphan20214431043}
Huy Phan.
\newblock huyvnphan/pytorch\_cifar10, jan 2021.

\bibitem[Pineau et~al.(2018)Pineau, Fried, Ke, and Larochelle]{pineau2018iclr}
Joelle Pineau, Genevieve Fried, Rosemary~Nan Ke, and Hugo Larochelle.
\newblock {ICLR}2018 reproducibility challenge.
\newblock In \emph{ICLR workshop on Reproducibility in Machine Learning}, 2018.

\bibitem[Pineau et~al.(2019)Pineau, Sinha, Fried, Ke, and
  Larochelle]{pineau2019iclr}
Joelle Pineau, Koustuv Sinha, Genevieve Fried, Rosemary~Nan Ke, and Hugo
  Larochelle.
\newblock {ICLR} reproducibility challenge 2019.
\newblock \emph{ReScience C}, 5\penalty0 (2), 2019.

\bibitem[Robbins(1956)]{robbins1956empirical}
Herbert Robbins.
\newblock An empirical {B}ayes approach to statistics.
\newblock In \emph{Proc. 3rd Berkeley Symp. Math. Statist. Probab., 1956},
  volume~1, 1956.

\bibitem[Sandler et~al.(2018)Sandler, Howard, Zhu, Zhmoginov, and
  Chen]{sandler2018mobilenetv2}
Mark Sandler, Andrew Howard, Menglong Zhu, Andrey Zhmoginov, and Liang-Chieh
  Chen.
\newblock Mobilenetv2 inverted residuals and linear bottlenecks.
\newblock In \emph{Proceedings of the IEEE conference on computer vision and
  pattern recognition}, 2018.

\bibitem[Settles(2009)]{settles_survey}
Burr Settles.
\newblock Active learning literature survey.
\newblock \emph{University of Wisconsin, Madison, Department of Computer
  Sciences}, 2009.

\bibitem[Shannon(1948)]{shannon1948mathematical}
Claude~E Shannon.
\newblock A mathematical theory of communication.
\newblock \emph{The Bell system technical journal}, 27\penalty0 (3), 1948.

\bibitem[Shen et~al.(2020)Shen, Xiong, Xia, and Soatto]{shen2020towards}
Yantao Shen, Yuanjun Xiong, Wei Xia, and Stefano Soatto.
\newblock Towards backward-compatible representation learning.
\newblock In \emph{Proceedings of the IEEE/CVF Conference on Computer Vision
  and Pattern Recognition}, 2020.

\bibitem[Shimodaira(2000)]{shimodaira2000improving}
Hidetoshi Shimodaira.
\newblock Improving predictive inference under covariate shift by weighting the
  log-likelihood function.
\newblock \emph{Journal of statistical planning and inference}, 90\penalty0
  (2), 2000.

\bibitem[Simonyan and Zisserman(2015)]{simonyan2014very}
Karen Simonyan and Andrew Zisserman.
\newblock Very deep convolutional networks for large-scale image recognition.
\newblock \emph{International Conference on Learning Representations}, 2015.

\bibitem[Sinha et~al.(2020)Sinha, Pineau, Forde, Ke, and
  Larochelle]{sinha2020neurips}
Koustuv Sinha, Joelle Pineau, Jessica Forde, Rosemary~Nan Ke, and Hugo
  Larochelle.
\newblock Neurips 2019 reproducibility challenge.
\newblock \emph{ReScience C}, 6\penalty0 (2), 2020.

\bibitem[Srivastava et~al.(2020)Srivastava, Nushi, Kamar, Shah, and
  Horvitz]{srivastava2020empirical}
Megha Srivastava, Besmira Nushi, Ece Kamar, Shital Shah, and Eric Horvitz.
\newblock An empirical analysis of backward compatibility in machine learning
  systems.
\newblock In \emph{Proceedings of the 26th ACM SIGKDD International Conference
  on Knowledge Discovery \& Data Mining}, 2020.

\bibitem[Sugiyama and Kawanabe(2012)]{sugiyama2012machine}
Masashi Sugiyama and Motoaki Kawanabe.
\newblock \emph{Machine learning in non-stationary environments: Introduction
  to covariate shift adaptation}.
\newblock MIT press, 2012.

\bibitem[Sutton and Barto(2018)]{sutton_barto_book}
Richard~S Sutton and Andrew~G Barto.
\newblock \emph{Reinforcement learning: An introduction}.
\newblock MIT press, 2018.

\bibitem[Szegedy et~al.(2014)Szegedy, Zaremba, Sutskever, Bruna, Erhan,
  Goodfellow, and Fergus]{szegedy2013intriguing}
Christian Szegedy, Wojciech Zaremba, Ilya Sutskever, Joan Bruna, Dumitru Erhan,
  Ian Goodfellow, and Rob Fergus.
\newblock Intriguing properties of neural networks.
\newblock \emph{International Conference on Learning Representations}, 2014.

\bibitem[Szegedy et~al.(2015)Szegedy, Liu, Jia, Sermanet, Reed, Anguelov,
  Erhan, Vanhoucke, and Rabinovich]{szegedy2015going}
Christian Szegedy, Wei Liu, Yangqing Jia, Pierre Sermanet, Scott Reed, Dragomir
  Anguelov, Dumitru Erhan, Vincent Vanhoucke, and Andrew Rabinovich.
\newblock Going deeper with convolutions.
\newblock In \emph{Proceedings of the IEEE conference on computer vision and
  pattern recognition}, 2015.

\bibitem[Szegedy et~al.(2016)Szegedy, Vanhoucke, Ioffe, Shlens, and
  Wojna]{szegedy2016rethinking}
Christian Szegedy, Vincent Vanhoucke, Sergey Ioffe, Jon Shlens, and Zbigniew
  Wojna.
\newblock Rethinking the inception architecture for computer vision.
\newblock In \emph{Proceedings of the IEEE conference on computer vision and
  pattern recognition}, 2016.

\bibitem[Touvron et~al.(2019)Touvron, Vedaldi, Douze, and
  J{\'e}gou]{touvron2020fixing}
Hugo Touvron, Andrea Vedaldi, Matthijs Douze, and Herv{\'e} J{\'e}gou.
\newblock Fixing the train-test resolution discrepancy: Fixefficientnet.
\newblock \emph{Advances in Neural Information Processing Systems 32}, 2019.

\bibitem[Xie et~al.(2020)Xie, Luong, Hovy, and Le]{xie2020self}
Qizhe Xie, Minh-Thang Luong, Eduard Hovy, and Quoc~V Le.
\newblock Self-training with noisy student improves imagenet classification.
\newblock In \emph{Proceedings of the IEEE/CVF Conference on Computer Vision
  and Pattern Recognition}, 2020.

\bibitem[Xie et~al.(2017)Xie, Girshick, Doll{\'a}r, Tu, and
  He]{xie2017aggregated}
Saining Xie, Ross Girshick, Piotr Doll{\'a}r, Zhuowen Tu, and Kaiming He.
\newblock Aggregated residual transformations for deep neural networks.
\newblock In \emph{Proceedings of the IEEE conference on computer vision and
  pattern recognition}, 2017.

\bibitem[Yan et~al.(2021)Yan, Xiong, Kundu, Yang, Deng, Wang, Xia, and
  Soatto]{yan2020positive}
Sijie Yan, Yuanjun Xiong, Kaustav Kundu, Shuo Yang, Siqi Deng, Meng Wang, Wei
  Xia, and Stefano Soatto.
\newblock Positive-congruent training: Towards regression-free model updates.
\newblock \emph{Conference on Computer Vision and Pattern Recognition}, 2021.

\bibitem[Yan et~al.(2016)Yan, Chaudhuri, and Javidi]{yan2016}
Songbai Yan, Kamalika Chaudhuri, and Tara Javidi.
\newblock Active learning from imperfect labelers.
\newblock \emph{Advances in Neural Information Processing Systems 29}, 2016.

\end{thebibliography}

\clearpage

\appendix
\section{Supplementary Material}

\label{sec:appendix_experiments}

In this supplement, we present additional results and ablations of experiments referred to in \cref{sec:experiments} of the main paper.

\subsection{Robustness to Random and Adversarial Model Sequences}
As already eluded to in~\cref{subsec:further_experiments_and_ablations}, in our main experiments we have assumed that the models $C^t$ are \textit{improving} over time.
In general, this assumption may be justified by having observed superior performance of a new model for the source domain on which they were trained, prior to deciding to use it for re-evaluation on the target data set.
However, such information may not always be available.

\begin{figure}[hb]
\subfigure[Random order]{
    \includegraphics[width=0.48\columnwidth]{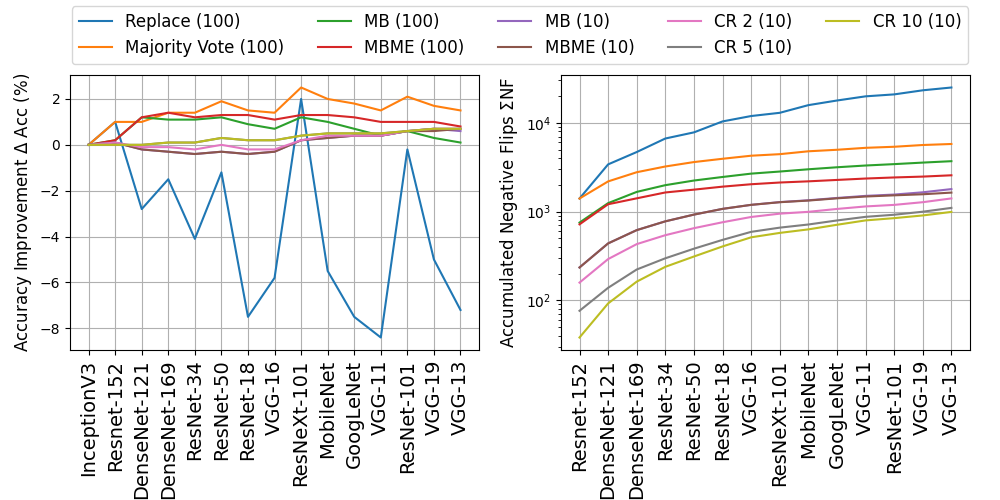}}
\hfill
\subfigure[Adversarial order]{
    \includegraphics[width=0.48\columnwidth]{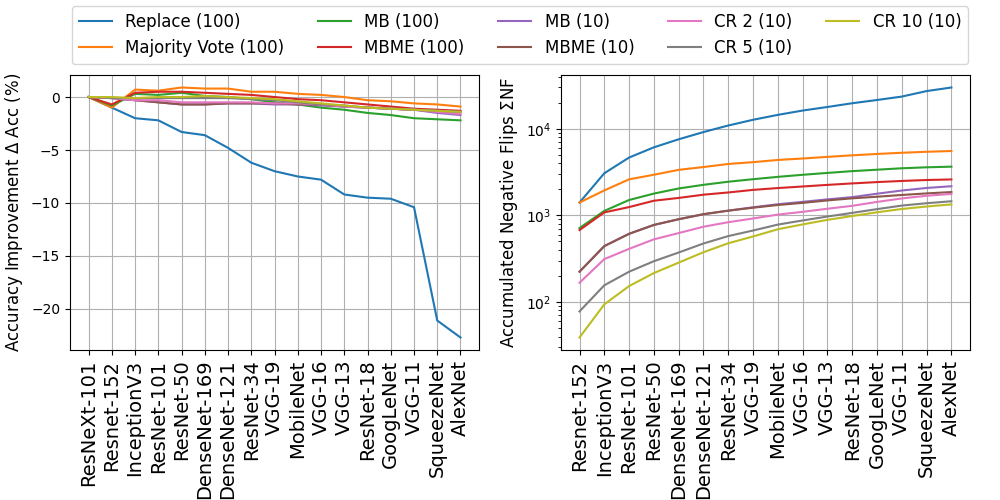}}
    \caption{Temporal evolution of accuracy improvement and accumulated negative flips on ImageNet for two scenarios where the models $C^t$ do not exhibit improving performance, but instead arrive in (a) random  or (b) adversarial order. As can be seen, our methods are robust in both these cases while introducing much fewer negative flips than the non-probabilistic baselines.}
    \label{fig:app:imagenet_robustness}
\end{figure}

To check for robustness against violations of the assumption of improving classifiers, we also consider scenarios where $C^t$ arrive in a \textit{random} or \textit{adversarial} (i.e., strictly deteriorating) order with respect to their accuracy.
To avoid unrealistically large fluctuations in the accuracy of new classifiers, we removed AlexNet and SqueezeNet as the most poorly performing models from the \textit{random} model sequence for this set of experiments.
Results on ImageNet in the form of the temporal evolution of accuracy improvement and accumulated negative flips across different strategies and budgets are shown in~\cref{fig:app:imagenet_robustness}.

For the random ordering in~\cref{fig:app:imagenet_robustness}
(a), we find that our methods---unlike, e.g., the replace strategy---achieve strict increases in \textit{final} accuracy while introducing much fewer negative flips, e.g., almost an order of magnitude fewer for our CR strategies, compared to the Majority Vote baseline.
Even in the adversarial case in~\cref{fig:app:imagenet_robustness}
 (b), our methods improve accuracy during the initial steps and introduce much fewer negative flips over the entire history.
These findings suggest robustness of our approach to situations where an ordering by performance of $C^t$ may not be available.

\subsection{Role of Confusion Matrix Estimates}
\label{subsec:appendix_confusion_matrix}
For the results of our main experiments reported in~\cref{sec:results_imagenet} and~\cref{sec:results_objectnet}, we estimated only the diagonal elements of the confusion matrices and set the off-diagonal elements to a constant, c.f.~\cref{sec:Bayesian_updates}.
To investigate the role of the estimation procedure for the confusion matrices $\pi^t$, we also report results
for the case
where we estimate the \textit{full} confusion matrix of each classifier (i.e., including the off-diagonal elements).
To avoid off-diagonal elements which are estimated to be zero due to the large number of classes and the limited size of the validation set split, we use a Laplace-smoothed version of the maximum likelihood estimate, i.e., we add a one to each count and normalise accordingly, as suggested in~\cref{sec:limitations}.

\begin{figure}[h]
\subfigure[\small ImageNet: Role of Confusion Matrix]{
    \includegraphics[width=0.3\columnwidth]{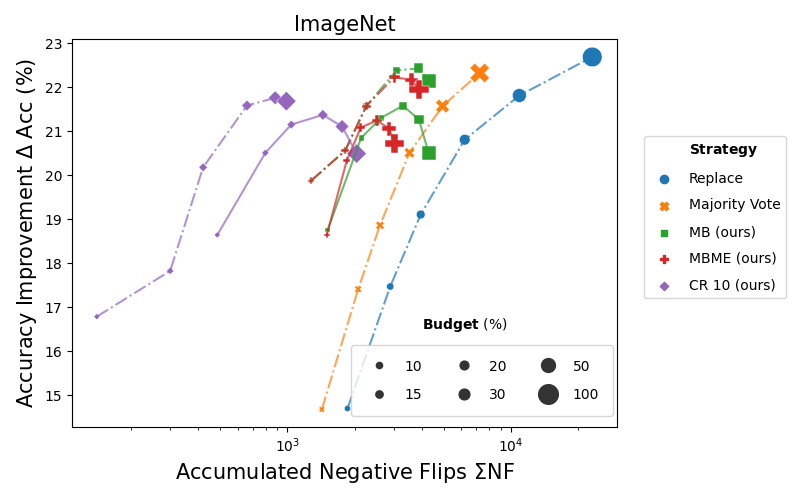}}
\hfill
\subfigure[\small ImageNet: Diagonal Confusion Matrix]{
    \includegraphics[width=0.33\columnwidth]{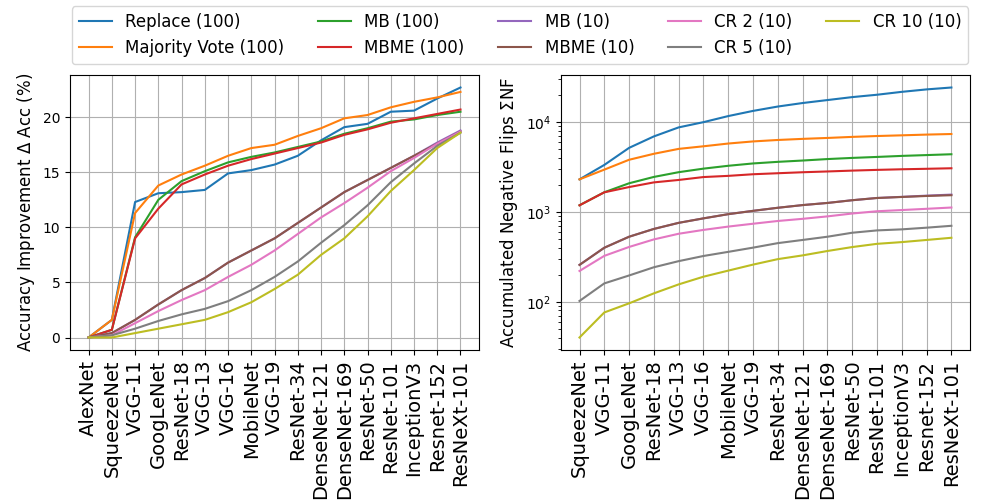}}
\hfill
\subfigure[\small ImageNet: Full Confusion Matrix]{
    \includegraphics[width=0.33\columnwidth]{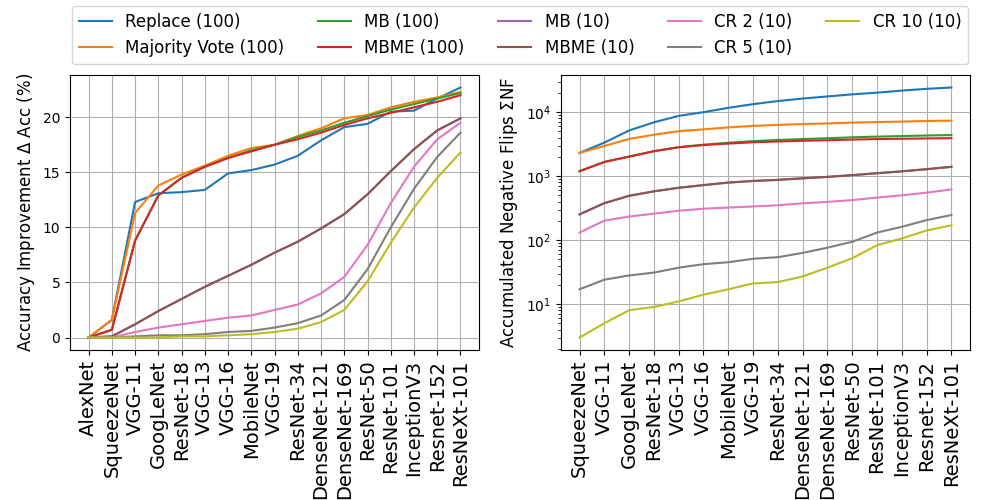}}
\subfigure[\small ObjectNet: Role of Confusion Matrix]{
    \includegraphics[width=0.3\columnwidth]{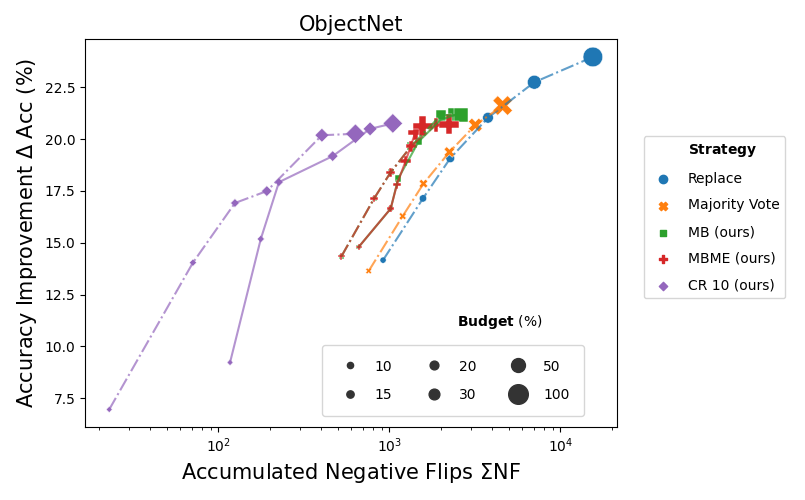}}
\hfill
\subfigure[\small ObjectNet: Diagonal Confusion Matrix]{
    \includegraphics[width=0.33\columnwidth]{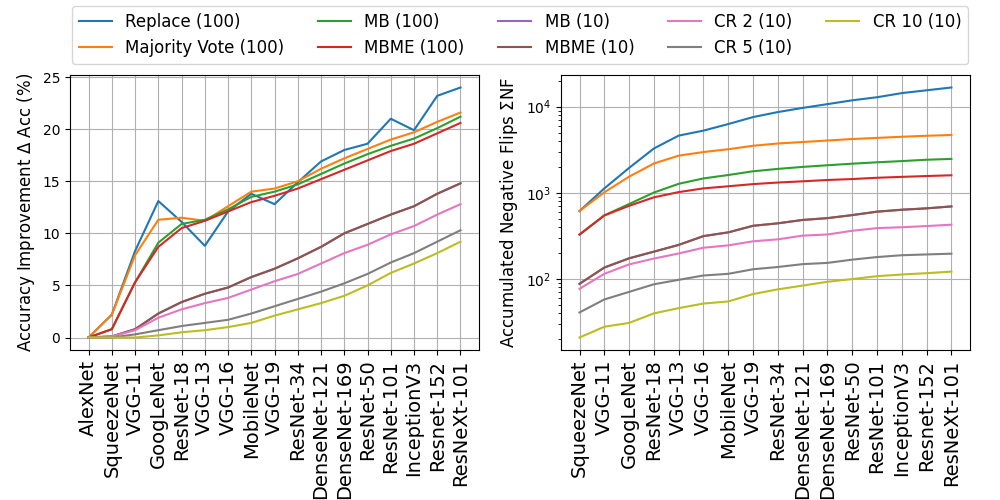}}
\hfill
\subfigure[\small ObjectNet: Full Confusion Matrix]{
    \includegraphics[width=0.33\columnwidth]{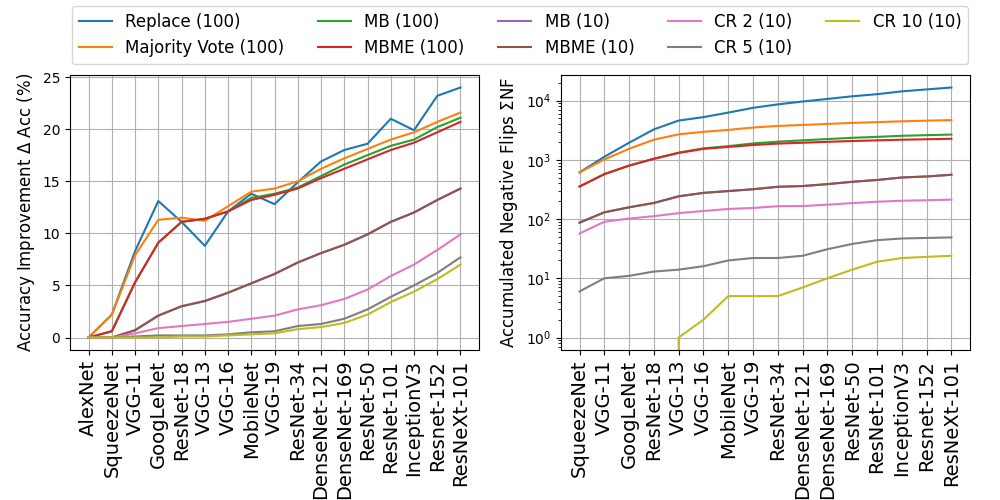}}
    \caption{Investigation into the role of different confusion matrix estimates on ImageNet (top) and ObjectNet (bottom).
    Dashed lines in the scatter plots (a) and (d) represent results for when the full confusion matrices (i.e., including the off-diagonals) are estimated from smoothed counts on a split of the validation set, and solid lines represent results under the less accurate confusion matrix estimates where only diagonal elements (i.e., the class-specific accuracies) are estimated, as in the experiments of the main paper.
    }
    \label{fig:app:diagonal_vs_full_confusion_estimation}
    \vspace{-1em}
\end{figure}

Results of this comparison between different confusion matrix estimators for ImageNet and ObjectNet are shown in~\cref{fig:app:diagonal_vs_full_confusion_estimation}.
(The corresponding results for CIFAR-10 are shown in~\cref{fig:app:diagonal_vs_full_confusion_estimation_cifar10} and are discussed in more detail in~\ref{sec:app_cifar}.)
As is apparent from the comparison on ImageNet and ObjectNet  in~\cref{fig:app:diagonal_vs_full_confusion_estimation}, we indeed observe generally improved performance from estimating full confusion matrices, as speculated in~\cref{sec:limitations}.

In terms of final accuracy, we observe a different behaviour across different budgets.
For large budgets, estimating full confusion matrices leads to substantial accuracy gains on ImageNet and to roughly equal or slightly improved accuracy on ObjectNet.
For small budgets, on the other hand, we find similar or smaller accuracy gains for full confusion matrix estimates; importantly, this effect is most pronounced for the more conservative CR prediction-update strategies.
In terms of negative flips, we observe slight to major reductions across all budgets and strategies with full confusion matrix estimates compared to only estimating the diagonals, i.e., the class-specific accuracies.

We believe that this behaviour is intuitive and can be explained as follows.
Estimating the full confusion matrices with smoothed counts generally yields smaller diagonal elements and larger off-diagonal elements.
This, in turn, means that our posterior beliefs (in particular, the ratio between two consecutive MAP values) change less drastically after new re-evaluations, so that the more conservative CR strategies update fewer labels and thus experience smaller and slower accuracy gains when only few samples can be re-evaluated.
At the same time, the resulting posterior estimates are likely more accurate which is consistent with larger gains for large budgets and the reduced number of negative flips.
In summary, better confusion matrix estimates result in a further reduction in the number of negative flips across the board. For small budgets and conservative prediction-update strategies, however, this comes with a trade-off in overall accuracy gain.

\subsection{Experiments on CIFAR-10}
\label{sec:app_cifar}
CIFAR-10 is an image classification dataset comprising images of objects from 10 distinct classes (airplanes, cars, birds, cats, deer, dogs, frogs, horses, ships, trucks).
It contains a training set of 50,000 images and a test set of size 10,000.
It is thus a rather simple classification dataset compared to ImageNet and ObjectNet but has been a key driver for advancing ML models and computer vision before the introduction of ImageNet.
Using the CIFAR-10 test set as our target dataset, we can study the Prediction Updates Problem in an i.i.d. setting with a small number of classes.

\begin{figure}[th]
\subfigure[Comparison of final performance]{
    \includegraphics[width=0.28\linewidth]{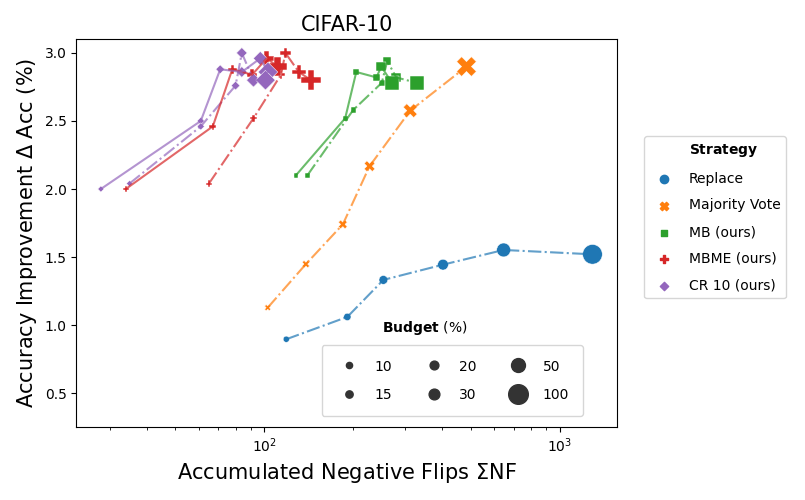}}
\hfill
\subfigure[Diagonal confusion matrix estimate]{
    \includegraphics[width=0.34\linewidth]{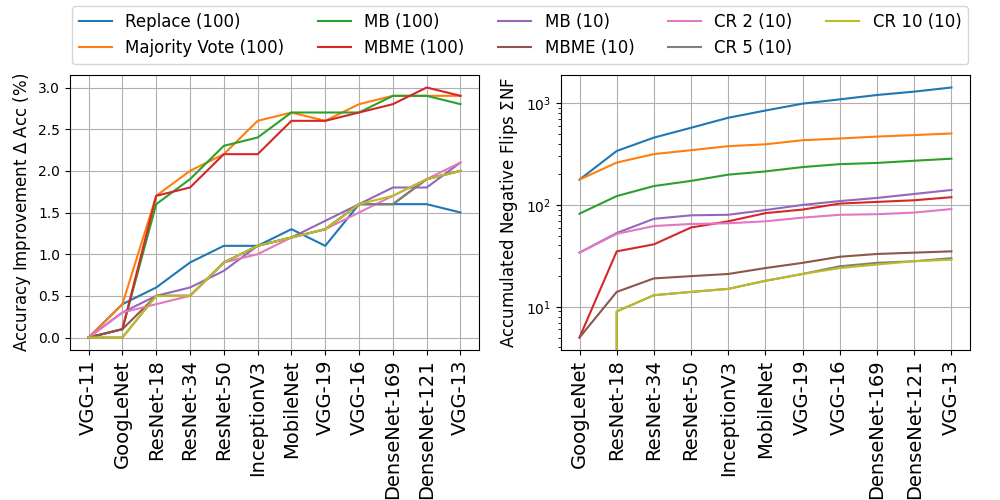}
    }
\hfill
\subfigure[Full confusion matrix estimate]{
    \includegraphics[width=0.34\linewidth]{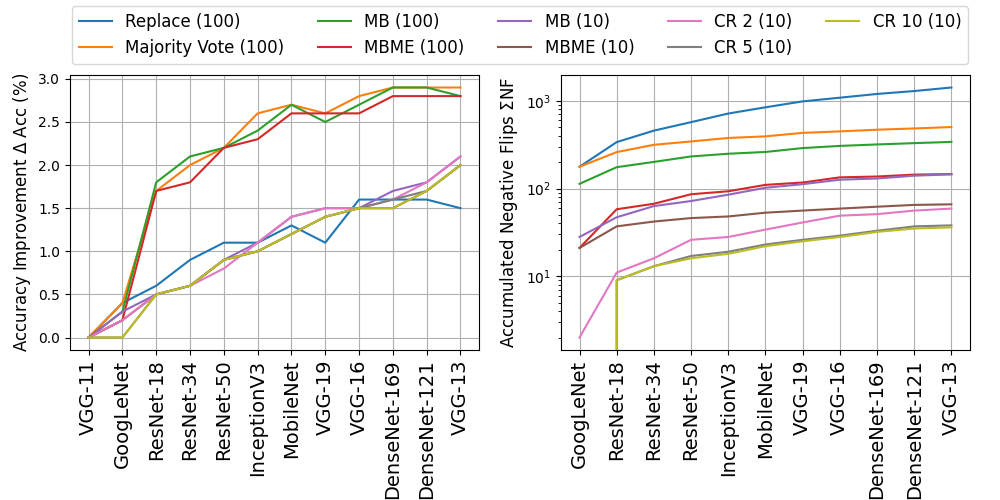}}
    \caption{Summary of prediction-updates on CIFAR-10. Similar to~\cref{fig:app:diagonal_vs_full_confusion_estimation}, dashed lines in (a) represent results with full confusion matrices estimated from smoothed counts and solid lines represent using a less accurate confusion matrix estimate where only diagonals are estimated from data.
    We observe very similar behaviour to that described for ImageNet and ObjectNet in~\cref{sec:results_imagenet} and~\cref{sec:results_objectnet}, respectively.
    The most interesting difference compared to our experiments on ImageNet and ObjectNet, is that, on CIFAR-10, our methods achieve substantially larger accuracy gains compared with the baselines, even for small budgets.
     }
    \label{fig:app:diagonal_vs_full_confusion_estimation_cifar10}
\end{figure}

Similar to our experiments on ImageNet and ObjectNet reported in~\cref{sec:experiments} where we used models which had been pre-trained on ImageNet,\footnote{\href{https://pytorch.org/vision/0.8/models.html}{ImageNet pretrained models}}
 for our experiments on CIFAR-10, we use models which have instead been pre-trained on CIFAR-10, available from an open source github repository.\footnote{\href{https://github.com/huyvnphan/PyTorch_CIFAR10}{CIFAR-10 pretrained models}}
This repository contains pre-trained models for a subset of the same architectures listed in~\cref{sec:experimental_setup}, excluding AlexNet, SqueezeNet, ResNet-101, ResNeXt;
the released VGG pre-trained models represent variants with batch normalization.
We order these CIFAR-10 pre-trained models according to their validation accuracies reported in the above repository.
Compared to ImageNet and ObjectNet, pre-trained models on CIFAR-10 exhibit a much higher level of accuracy (between 93\% and 95\%) with much smaller differences between the best and worst model.
As a consequence, our experiment on CIFAR-10 emulates a scenario in which incoming new classifiers exhibit smaller performance gains.
This might reflect a practical prediction-updates setting with a higher frequency of incoming new models.
\begin{wrapfigure}{r}{8cm}
\includegraphics[width=\linewidth]{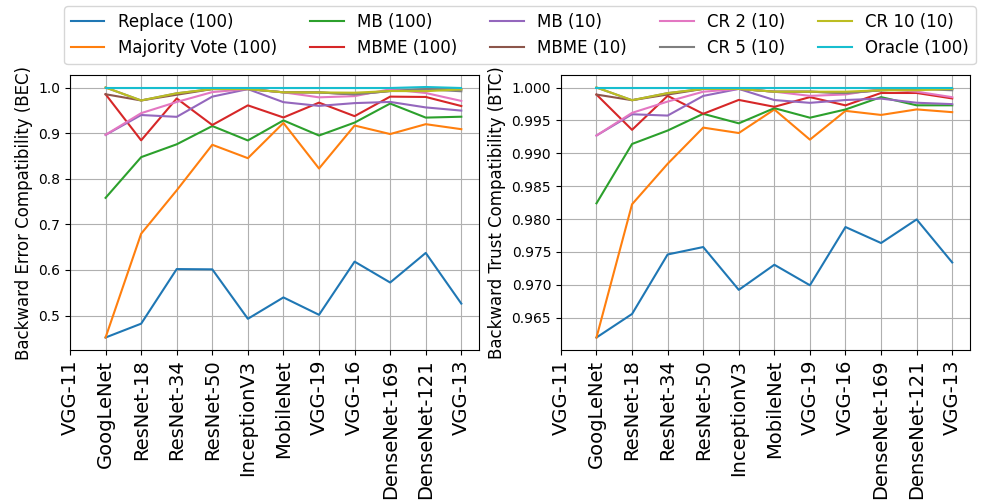}
\caption{\small BTC and BEC evolution on CIFAR10 with utilizing diagonal confusion matrix elements only.}
 \label{fig:app:diagonal_cifar10_btc_bec}
\end{wrapfigure}

The results of our experiments on CIFAR-10 are summarised in~\cref{fig:app:diagonal_vs_full_confusion_estimation_cifar10}.
Our method shows very similar trends w.r.t.\, all update strategies and metrics as we have worked out on ImageNet and ObjectNet.
Interestingly, there is one novel characteristic when it comes to the accuracy gains:
due to the presumably less steep increase in accuracy from one model to its successor and fewer class categories, we can form more accurate posterior beliefs which results in accuracy gains for our CR 10 with smallest budget that outperforms the backfilling Replace (100) baseline.
The overall accuracy gain similarly remains competitive with all Majority Vote variants while exhibiting only a fraction of the negative flips. Again, BTC and BEC strongly outperform all baselines along every update step for all our methods as can be seen in \cref{fig:app:diagonal_cifar10_btc_bec}

\subsection{On Peaking Accuracies}
In~\cref{fig:imagenet_improving} (right, solid lines), we observed a peak in $\Delta$\textbf{Acc} at 30\% budget for all our strategies when performing prediction-updates on ImageNet.
This suggests that our update rules are no longer beneficial on this dataset when re-evaluating more than the top 30\% of samples with highest posterior entropy.
From our ablation experiments on different confusion matrix estimates shown in~\cref{fig:app:diagonal_vs_full_confusion_estimation} (a), we observe that this peaking phenomenon is much less pronounced  and occurs only at larger budgets of $B\geq50\%$ when estimating the full confusion matrix based on the half of the validation set on which we do not evaluate.
We therefore conjecture that this behaviour may be related to inaccurate estimates of the posterior resulting from our approximation of the unknown confusion matrices (and possibly the assumption of conditionally independent classifiers).

To further investigate this hypothesis, we performed an ablation where we use the \textit{full} validation set both to estimate the confusion matrix (with smoothing) and for evaluation---i.e., we do not use a 50-50 split of the validation set as in~\cref{fig:app:diagonal_vs_full_confusion_estimation}  and all experiment in the main paper---thus matching almost exactly\footnote{up to the one count smoothing effect.} the confusion counts statistics of the evaluation set. As a result, we can expect our estimates of the posterior distribution to be much more accurate in this case.

\begin{figure}[h]
    \centering
    \includegraphics[width=0.4\columnwidth]{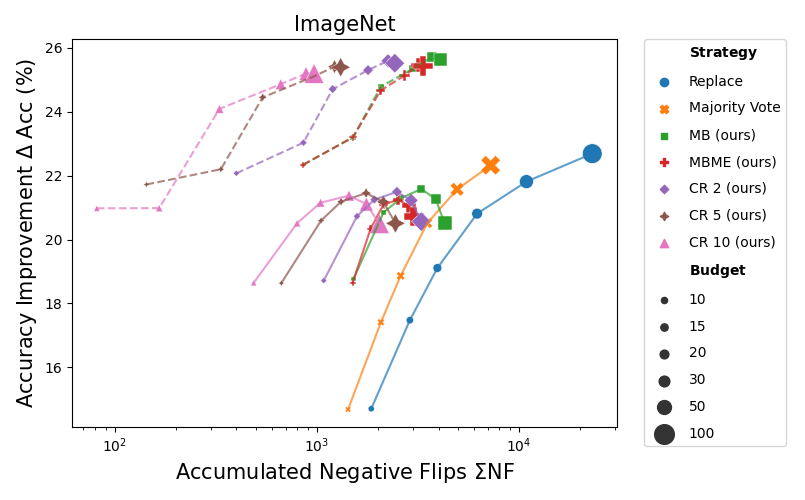}
    \caption{Further investigation into the peaking phenomenon observed for prediction-updates on ImageNet in~\cref{fig:imagenet_improving} (right).
Solid lines are as in~\cref{fig:imagenet_improving} (right), i.e., using a 50:50 split of the validation set, using one half to estimate the diagonal elements of the confusion matrices and the other half as our target evaluation set; dashed lines show the results of our ablation experiment where we use the full validation set to both estimate the exact statistics of the confusion matrix and to evaluate the different methods.
Since the evaluation set for the latter is twice as large, we scaled the reported negative flips by 0.5 for ease of comparison.}
    \label{fig:app:peaking_ablation}
\end{figure}

The results are shown in~\cref{fig:app:peaking_ablation} in dashed lines, compared to the peaking behaviour from our default method in solid lines for reference.
Apart from substantial gains in overall accuracy, we find that the peaking phenomenon disappears almost entirely,with only a very slight drop in accuracy remaining between budgets of 50\% and 100\% for some strategies.
This seems to confirm our hypothesis that the main source of the peaking behaviour are  approximation errors in our posterior estimates resulting from inaccurate confusion matrices.

\subsection{Role of the Selection Strategy}
\begin{figure}[hbtp]
\centering
\begin{minipage}{1.0\textwidth}
\includegraphics[width=0.33\textwidth]{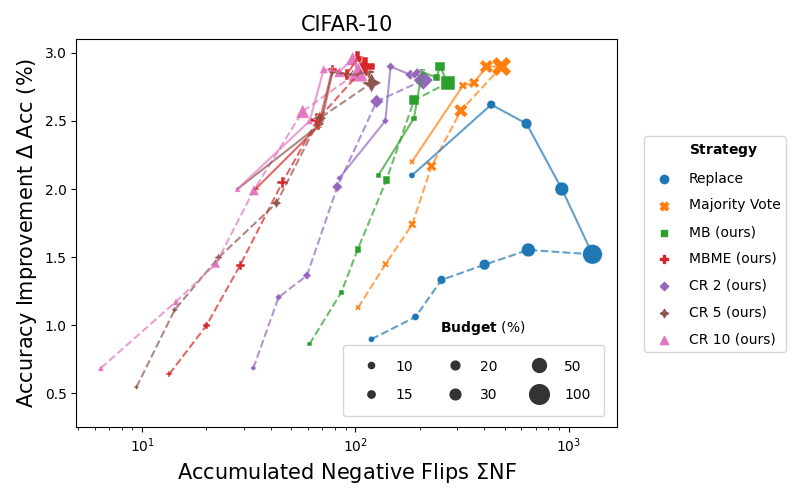}
\includegraphics[width=0.33\textwidth]{figures/plots/imagenet/diagonal/scatter_ablation_selection_ImageNet_complete.png}
\includegraphics[width=0.33\textwidth]{figures/plots/objectnet/diagonal/scatter_ablation_selection_ObjectNet_complete.png}
\end{minipage}
\vskip 0.1in
\begin{minipage}{1.0\textwidth}
\includegraphics[width=0.33\textwidth]{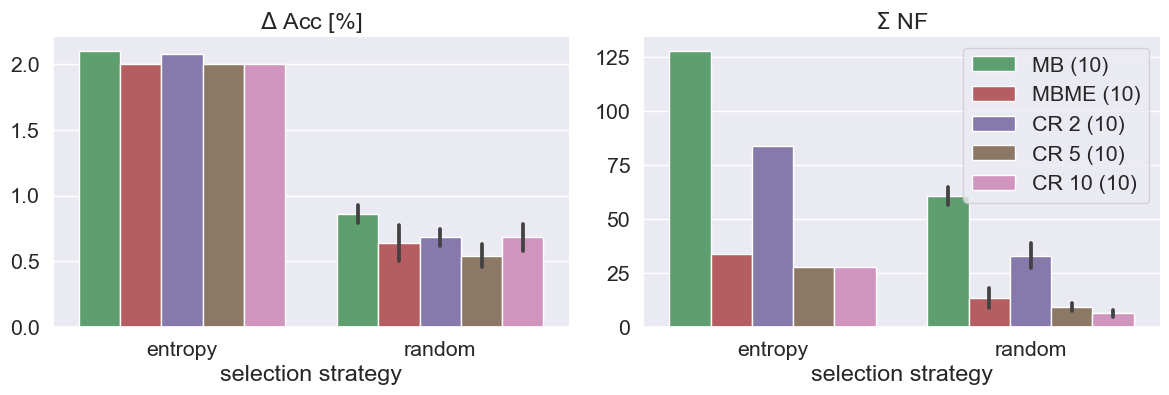}
\includegraphics[width=0.33\textwidth]{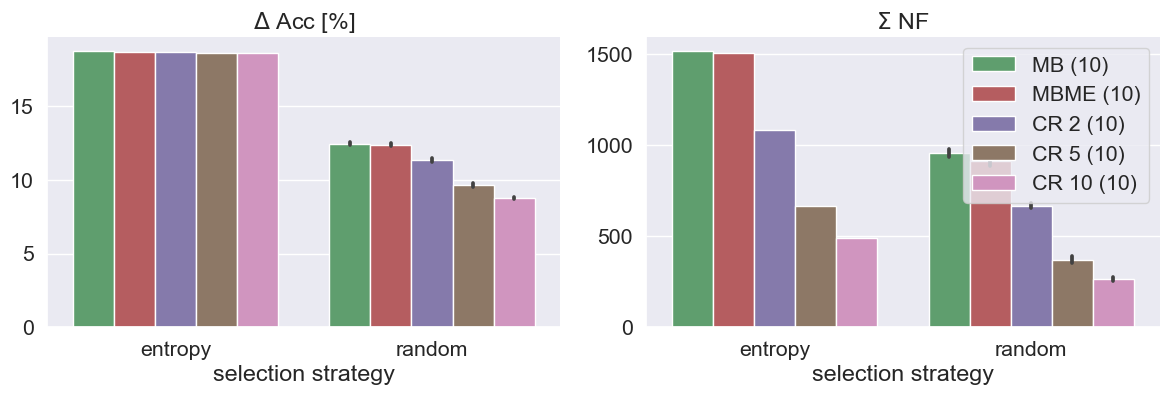}
\includegraphics[width=0.33\textwidth]{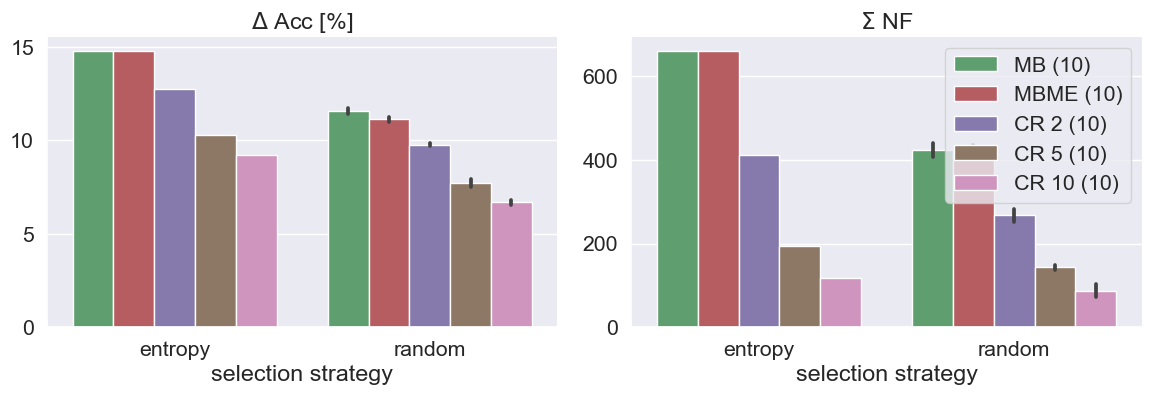}
\end{minipage}
\vskip 0.1in
\begin{minipage}{1.0\textwidth}
\includegraphics[width=0.33\textwidth]{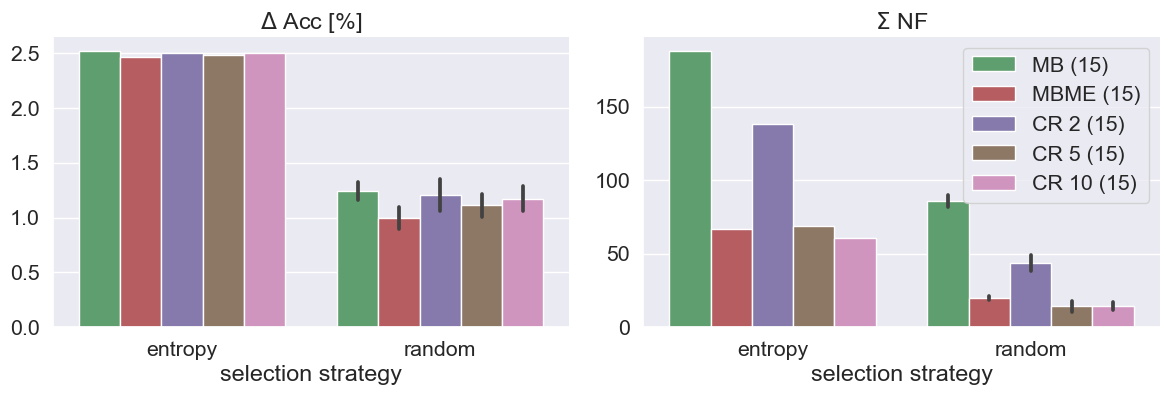}
\includegraphics[width=0.33\textwidth]{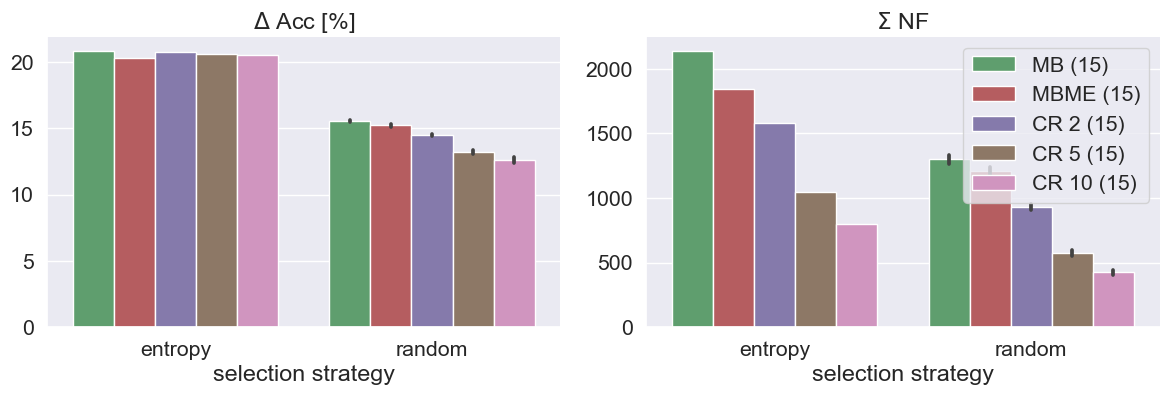}
\includegraphics[width=0.33\textwidth]{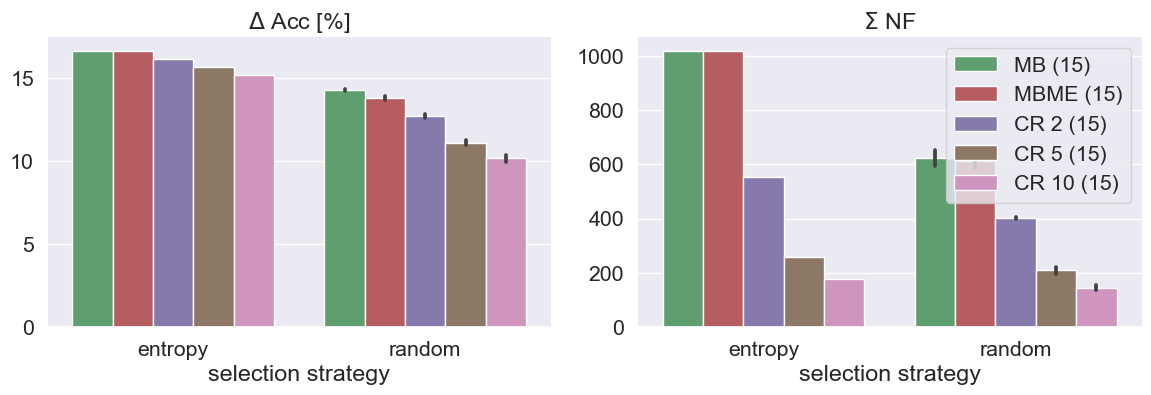}
\end{minipage}
\vskip 0.1in
\begin{minipage}{1.0\textwidth}
\includegraphics[width=0.33\textwidth]{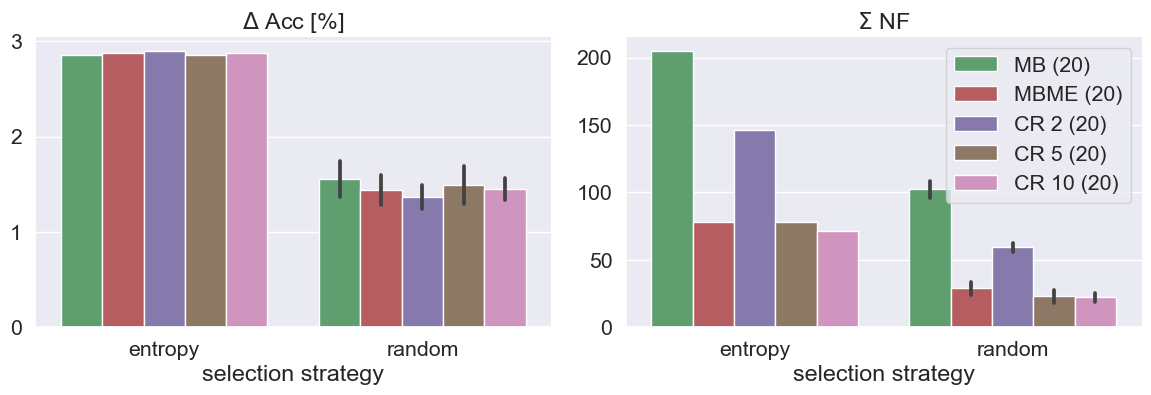}
\includegraphics[width=0.33\textwidth]{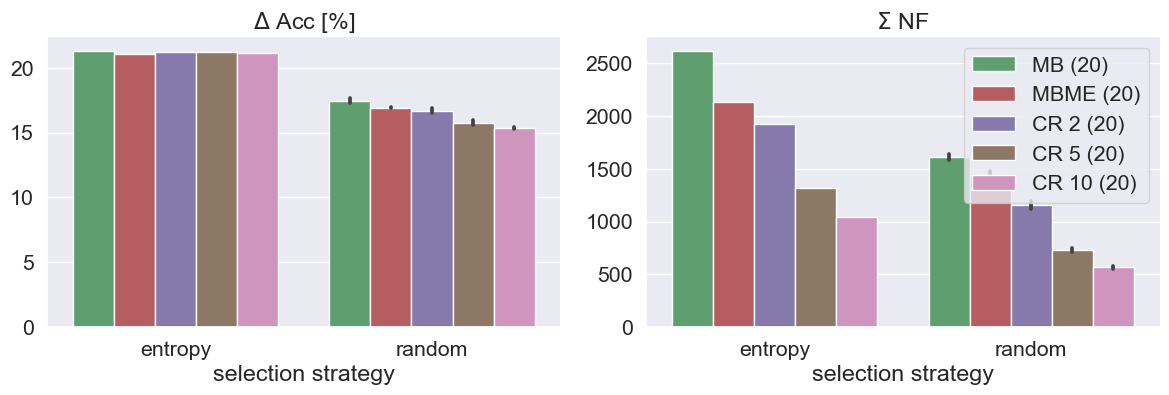}
\includegraphics[width=0.33\textwidth]{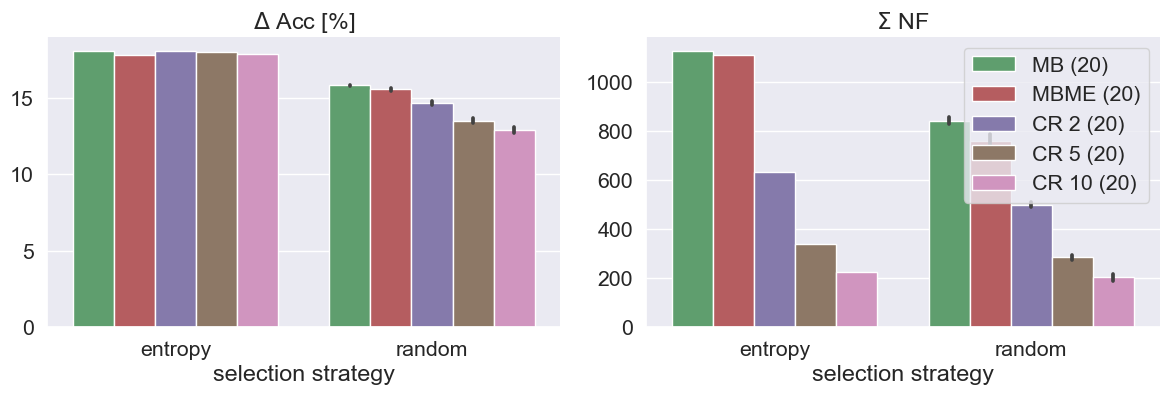}
\end{minipage}
\vskip 0.1in
\begin{minipage}{1.0\textwidth}
\includegraphics[width=0.33\textwidth]{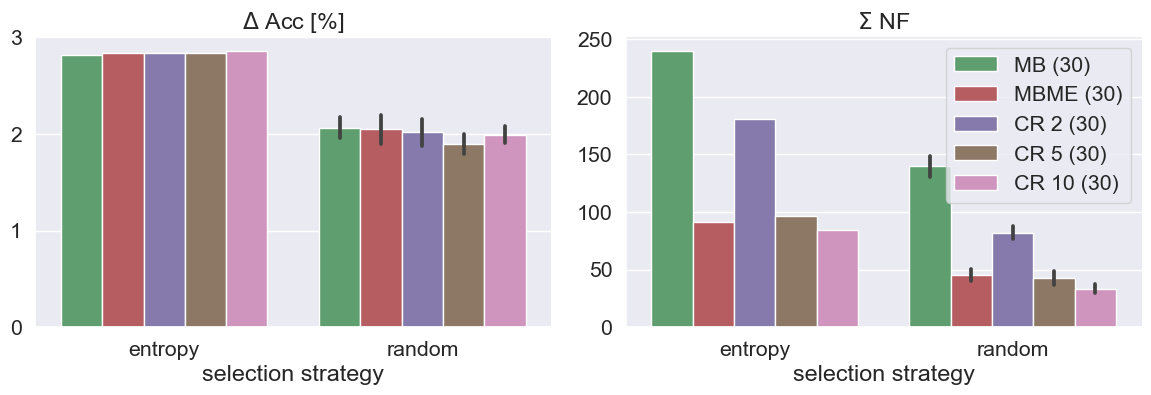}
\includegraphics[width=0.33\textwidth]{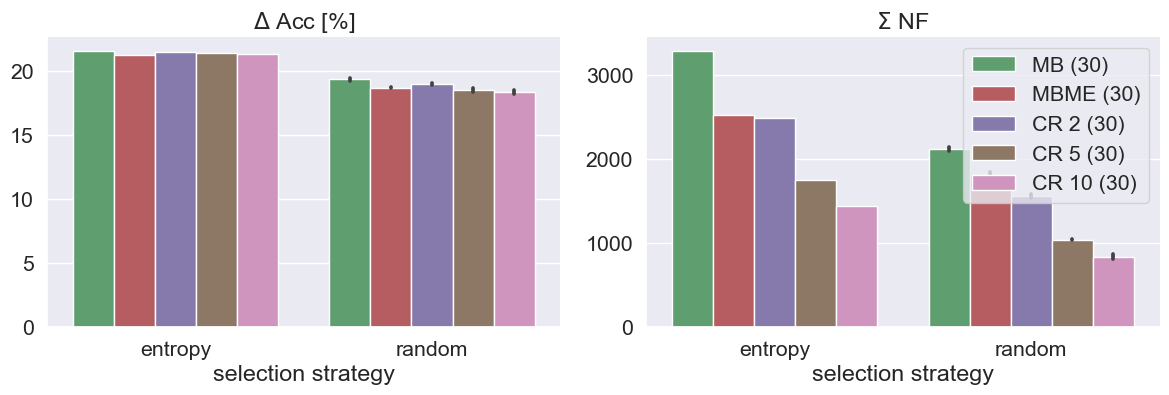}
\includegraphics[width=0.33\textwidth]{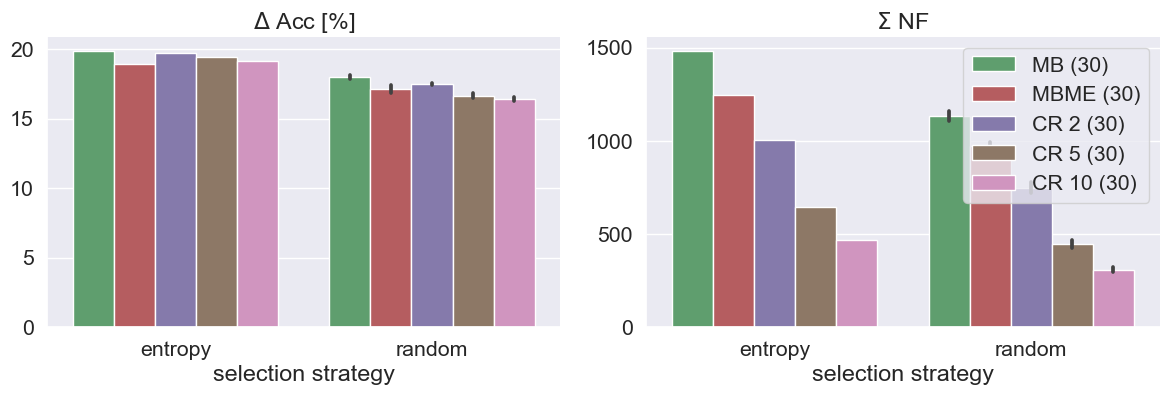}
\end{minipage}
\vskip 0.1in
\begin{minipage}{1.0\textwidth}
\includegraphics[width=0.33\textwidth]{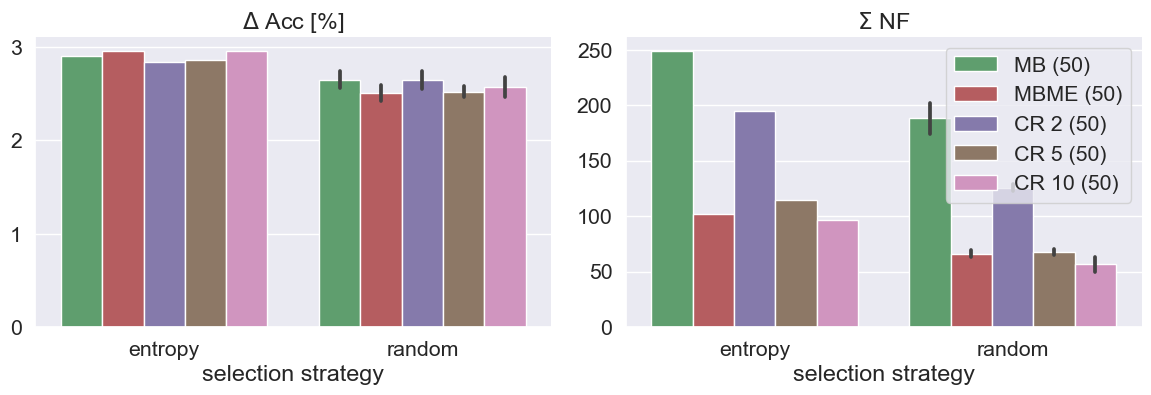}
\includegraphics[width=0.33\textwidth]{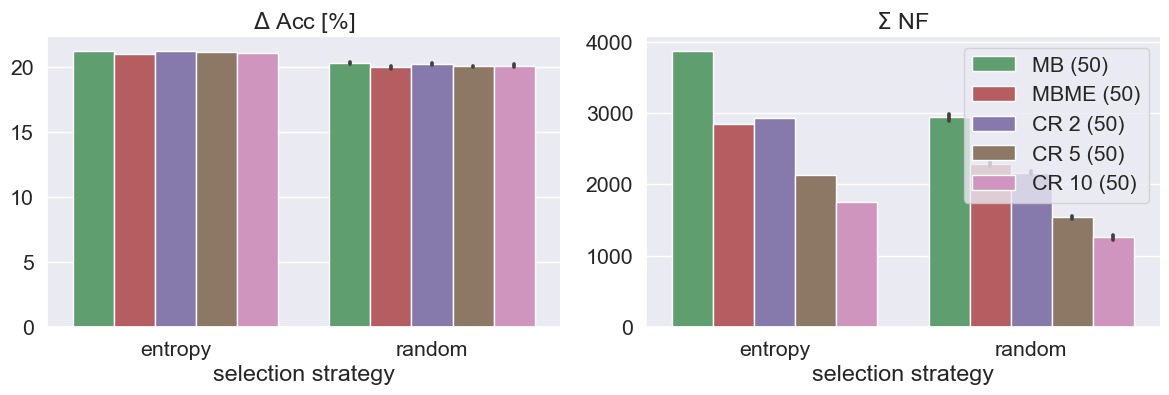}
\includegraphics[width=0.33\textwidth]{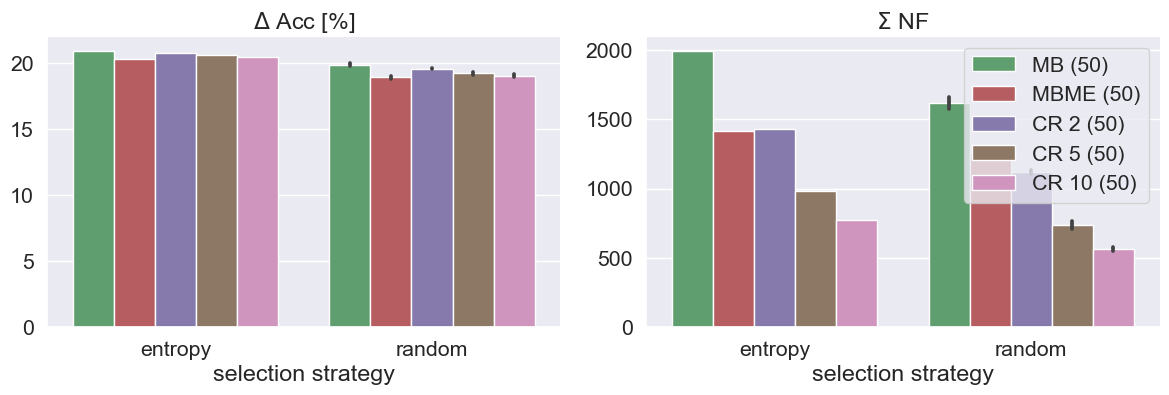}
\end{minipage}
\vspace{-1em}
    \caption{Ablation results for comparing random (dashed lines in scatter plots) and entropy-ranked (solid lines in scatter plots) selection strategies on CIFAR-10 (left column), ImageNet (middle column)  and ObjectNet (right column).
   Different rows of the bar plots correspond to, from top to bottom, $B=$10, 15, 20, 30, 50\%. Note that at 100\% budgets, the two selection strategies are identical as all samples are re-evaluated at each step.
    We find that random selection leads to substantially smaller accuracy gains, but also fewer negative flips. This is intuitive as random selection more often chooses ``easy'' samples for re-evaluation. The effect is particularly pronounced for small budgets.}
    \label{fig:app:ablation_selection_improving}
\end{figure}

We also conduct an ablation study on the role of the selection strategy for samples to be re-evaluated as was being mentioned in \cref{subsec:further_experiments_and_ablations}. In particular we conduct a comparison between our posterior entropy selection and the random selection (without replacement) baseline for all our methods across a range of budgets---see~ \cref{fig:app:ablation_selection_improving} for these ablation results on all three datasets CIFAR-10, ImageNet and ObjectNet.
Along all datasets, we find that random selection leads to substantially smaller accuracy gains, but also results in fewer negative flips. We can explain these fewer negative flips by the fact that our random selection more often chooses ``easy'' samples for re-evaluation. Under a random selection criteria, there is no preference in re-evaluating more ``controversial'' samples w.r.t\, previous predictions.
This effect becomes particularly pronounced for small budgets.
Our finding suggests that the selection strategy is indeed a relevant component, with room for improvement as discussed in~\cref{sec:limitations}. This is also highlighting the potential control capability of this component regarding our three desiderata.

\subsection{Evolution of the Stored Label Distribution}
\begin{figure}

\subfigure[After t=0 Updates]{
    \includegraphics[width=0.32\columnwidth]{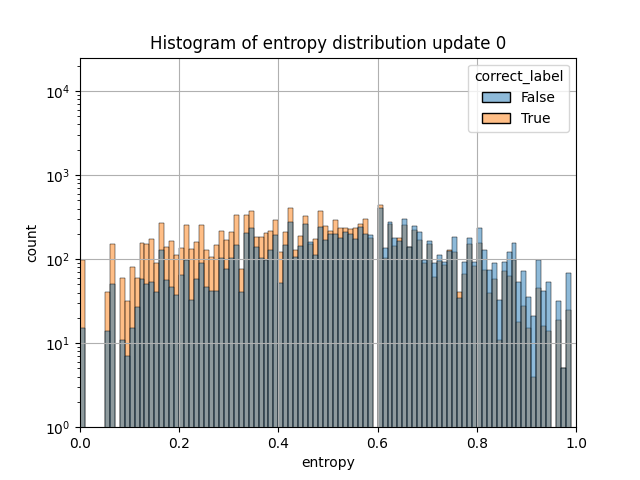}}
\hfill
\subfigure[After t=2 Updates]{
    \includegraphics[width=0.32\columnwidth]{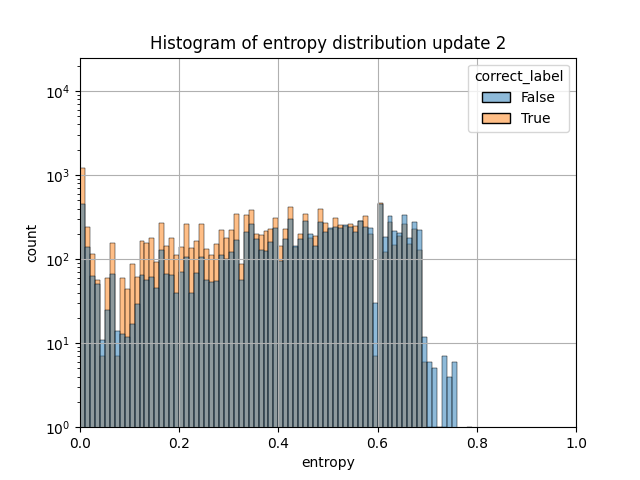}}
\hfill
\subfigure[After t=4 Updates]{
    \includegraphics[width=0.32\columnwidth]{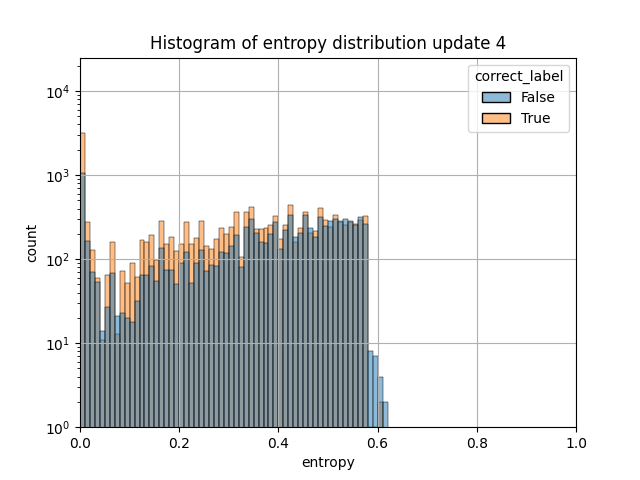}}
\vskip 0.05in
\subfigure[After t=6 Updates]{
    \includegraphics[width=0.32\columnwidth]{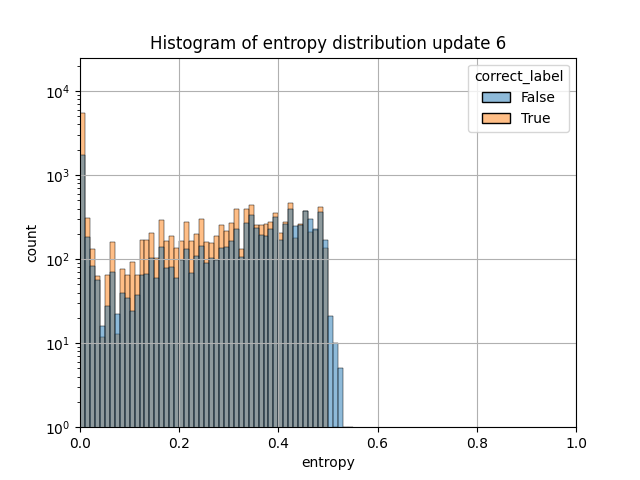}}
\hfill
\subfigure[After t=8 Updates]{
    \includegraphics[width=0.32\columnwidth]{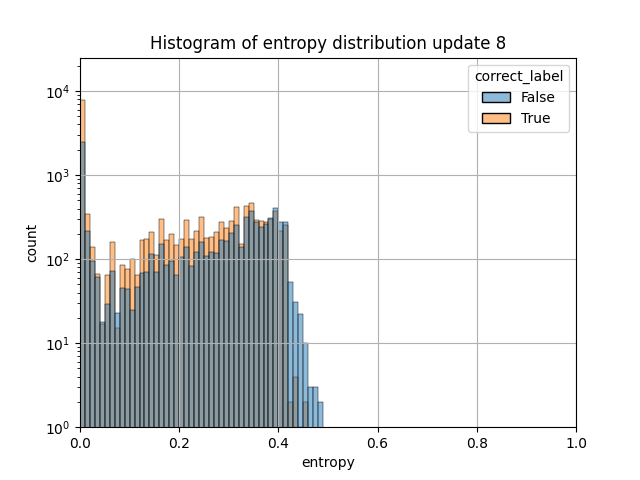}}
\hfill
\subfigure[After t=10 Updates]{
    \includegraphics[width=0.32\columnwidth]{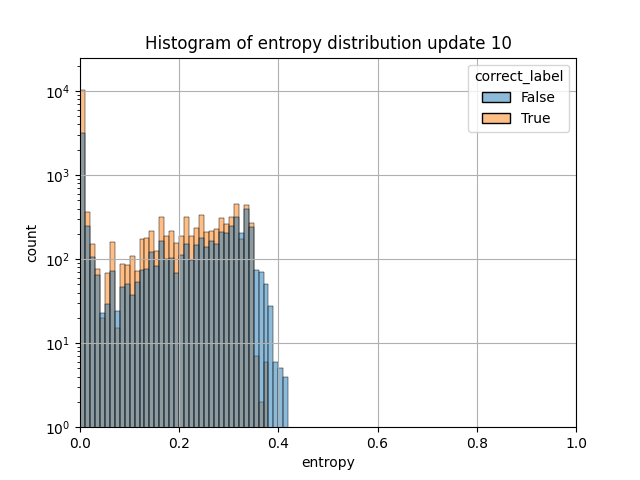}}
\vskip 0.05in
\subfigure[After t=12 Updates]{
    \includegraphics[width=0.32\columnwidth]{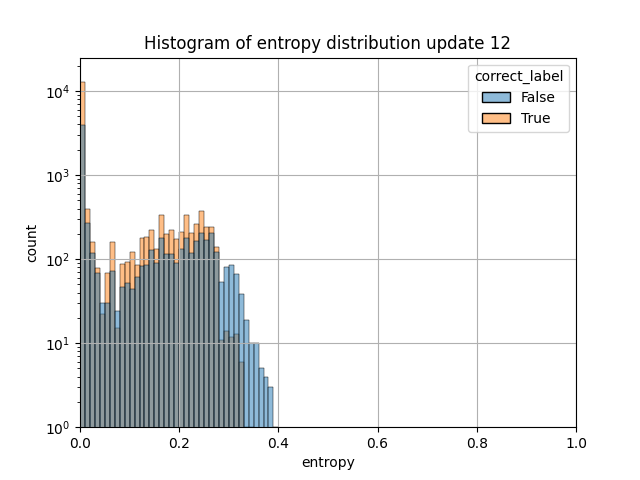}}
\hfill
\subfigure[After t=14 Updates]{
    \includegraphics[width=0.32\columnwidth]{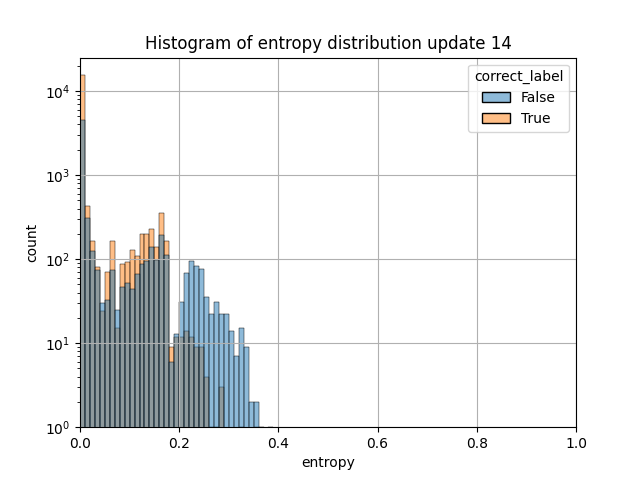}}
\hfill
\subfigure[After t=16 Updates]{
    \includegraphics[width=0.32\columnwidth]{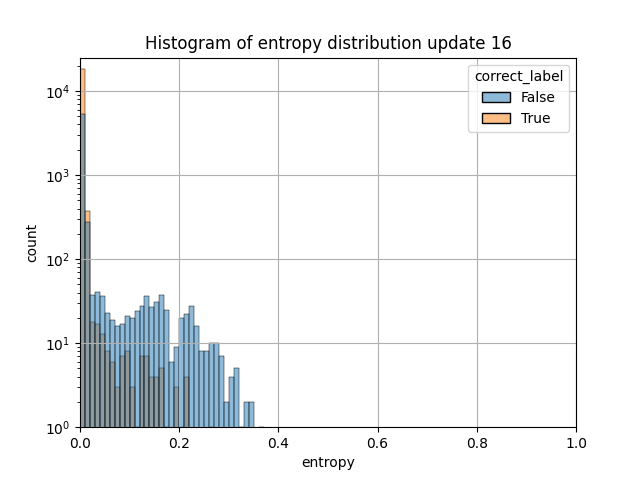}}

\caption{Count distribution evolution of correctly and incorrectly stored predictions on ImageNet using the MB update strategy under a budget of B = 10. Samples with currently stored predictions being false are depicted with blue, and correctly stored predictions with orange. We observe that under this MB strategy, samples with highest entropy tend to store false predictions, thus benefitting from re-evaluation particularly.}
    \label{fig:app:evolution_label_distribution}
\end{figure}

For additional insights which samples are getting selected for re-evaluation by our entropy selection criterion, we also show some characteristic count distribution of correctly and incorrectly stored predictions along various update steps. See \cref{fig:app:evolution_label_distribution} for this distribution after every second update step on ImageNet using the MB strategy with a compute budget of $B=10$ (\%).

\subsection{Reducing re-evaluations matters at scale}
To see that the re-evaluations of an example image using deep neural network based models clearly dominate computational cost as compared to our method backbone that involved computing the approximate posterior, label entropy and update strategy details, we measure the time for the plain model estimation vs the time for posterior update on the same Intel Xeon (Cascade Lake) CPU for the sake of comparison.
We summarize our measurement  in \cref{tab:inference_time_meassurements} for all used ImageNet models, showing the multiples in compute time required by the model inference (per image on average) compared to the timing per image of our fully unoptimized method implementation, i.e. the posterior update backbone.  For very large data sets and with new models generally increasing in size, reducing the inference budget B is of crucial importance, emphasizing the relevance of  desideratum 3.

\begin{table}
\small
\caption{Time measurements of a single forward pass versus the mean average time to compute the posterior update (0.406 milliseconds) on the same computational ressources for comparability. For reference we extrapolated how long it would take to infer the predictions from 1B images.}
\resizebox{1.0\columnwidth}{!}{
\begin{tabular}{l|l|l|l}
\toprule
Model & model re-evaluation [milliseconds] & ratio inference vs posterior update & inference for 1B images [days]\\
\midrule
alexnet &	29,6	& 73	& 343\\
squeezenet & 32,6 & 80 & 378\\
vgg11 &	125,5	&309	&1452\\
googlenet &	65,6	&162	&759\\
resnet18 &	31,5	&78	&364\\
vgg13 &164,1	&404	&1899\\
vgg16 &191,3	&471	&2215\\
mobilenet	&23,1	&57	&267\\
vgg19 &223,0	&550	&2582\\
resnet34 &50,8	&125	&587\\
densenet121 &74,4	&183	&862\\
densenet169 &93,5	&230	&1082\\
resent50 &70,6	&174	&818\\
resnet101 &114,0	&281	&1319\\
inceptionv3 &115,8	&285	&1340\\
resnet152 & 158,2	&390	&1831\\
resnext101 32x8d	&206,6	&509& 	2391\\
\bottomrule
\end{tabular}}
\label{tab:inference_time_meassurements}
\end{table}

\subsection{Result Tables of Main Experiments}
For sake of completeness and reproducibility, we also provide a complete quantitative account of all the main experiments and ablations on all three datasets discussed throughout this paper in~\cref{tab:app:imagenet_objectnet_improving}  and \cref{tab:app:cifar10_improving}.

\begin{table}
\small
\caption{Full results of all experiments on \textbf{ImageNet (left) and ObjectNet (right)} under the standard setting of an \textbf{improving sequence of models} and \textbf{estimating confusion matrices on a separate split} of the data (i.e., not on the target data set) as explained in~\cref{sec:experimental_setup}. The first number in each cell refers to estimating only the diagonal elements (i.e., the class-specific accuracies) of confusion matrices, and the \textbf{second number (in brackets) refers to estimating the full confusion matrices} with smoothing. The character \textbf{E} in front of a budget indicates that selection for re-evaluation is based on our posterior label \textbf{entropy criterion} (e.g., E30 refers to selecting the top 30\% samples with highest entropy), and the character \textbf{R} indicates selecting a \textbf{randomly} sampled subset without replacement. Note that entropy-based selection requires a posterior and is thus not applicable for the baselines Replace and Majority Vote.
}
\vskip 0.3in
\resizebox{1.0\columnwidth}{!}{
\begin{tabular}{l|rrrrrrr}
\toprule
 \textbf{Strategy}  Budget \%   &   \textbf{Acc} (\%) &   $\Delta$\textbf{Acc} (\%)&   $\Sigma$ \textbf{\NF} &   \textbf{\NFR}\, (\%) &  \textbf{ \PF}\, / \textbf{\NF} &  \textbf{Avg. BTC} &  \textbf{Avg. BEC} \\
\midrule\midrule
 100:Oracle         & 91.2 (91.2) & 34.7 (34.7) & 0 (0)             & 0.0 (0.0)   & NaN (NaN)   & 100.0 (100.0) & 100.0 (100.0) \\
 R100:Replace       & 79.2 (79.2) & 22.7 (22.7) & 24214 (24214)     & 6.05 (6.05) & 1.2 (1.2)   & 91.37 (91.37) & 77.71 (77.71) \\
 R100:Majority Vote & 78.9 (78.9) & 22.3 (22.3) & 7352 (7352)       & 1.84 (1.84) & 1.8 (1.8)   & 97.18 (97.18) & 93.95 (93.95) \\
 E100:MB            & 77.1 (78.7) & 20.5 (22.2) & 4378 (4373)       & 1.09 (1.09) & 2.2 (2.3)   & 98.32 (98.32) & 96.45 (96.4)  \\
 E100:MBME          & 77.3 (78.5) & 20.7 (22.0) & 3057 (3910)       & 0.76 (0.98) & 2.7 (2.4)   & 98.78 (98.48) & 97.69 (96.88) \\
 E100:CR 2          & 77.1 (78.5) & 20.6 (22.0) & 3368 (2342)       & 0.84 (0.59) & 2.5 (3.3)   & 98.72 (99.12) & 97.19 (98.0)  \\
 E100:CR 5          & 77.1 (78.4) & 20.5 (21.8) & 2520 (1300)       & 0.63 (0.33) & 3.0 (5.2)   & 99.06 (99.52) & 97.82 (98.86) \\
 E100:CR 10         & 77.0 (78.2) & 20.5 (21.7) & 2112 (1023)       & 0.53 (0.26) & 3.4 (6.3)   & 99.22 (99.62) & 98.15 (99.11) \\
 \midrule
 R50:Replace        & 78.4 (78.4) & 21.8 (21.8) & 11478.8 (11478.8) & 2.87 (2.87) & 1.5 (1.5)   & 95.82 (95.82) & 89.88 (89.88) \\
 R50:Majority Vote  & 78.1 (78.1) & 21.6 (21.6) & 5048.2 (5048.2)   & 1.26 (1.26) & 2.1 (2.1)   & 98.07 (98.07) & 95.92 (95.92) \\
 R50:MB             & 76.9 (77.7) & 20.4 (21.1) & 3022.8 (3052.8)   & 0.76 (0.76) & 2.7 (2.7)   & 98.83 (98.82) & 97.62 (97.57) \\
 R50:MBME           & 76.5 (77.5) & 20.0 (20.9) & 2325.4 (2917.6)   & 0.58 (0.73) & 3.1 (2.8)   & 99.08 (98.87) & 98.27 (97.72) \\
 R50:CR 2           & 76.8 (77.4) & 20.3 (20.8) & 2233.6 (1572.8)   & 0.56 (0.39) & 3.3 (4.3)   & 99.14 (99.4)  & 98.22 (98.74) \\
 R50:CR 5           & 76.6 (77.0) & 20.1 (20.4) & 1603.8 (840.6)    & 0.4 (0.21)  & 4.1 (7.1)   & 99.39 (99.68) & 98.69 (99.32) \\
 R50:CR 10          & 76.7 (76.7) & 20.1 (20.2) & 1314.6 (659.2)    & 0.33 (0.16) & 4.8 (8.6)   & 99.51 (99.75) & 98.92 (99.46) \\
 E50:Replace        & 78.9 (79.6) & 22.4 (23.1) & 15732 (15503)     & 3.93 (3.88) & 1.4 (1.4)   & 94.41 (94.53) & 85.37 (85.23) \\
 E50:Majority Vote  & 78.7 (79.1) & 22.2 (22.6) & 6318 (6420)       & 1.58 (1.6)  & 1.9 (1.9)   & 97.6 (97.58)  & 94.74 (94.55) \\
 E50:MB             & 77.8 (79.0) & 21.3 (22.4) & 3969 (3941)       & 0.99 (0.99) & 2.3 (2.4)   & 98.48 (98.5)  & 96.78 (96.71) \\
 E50:MBME           & 77.6 (78.7) & 21.1 (22.2) & 2904 (3636)       & 0.73 (0.91) & 2.8 (2.5)   & 98.85 (98.6)  & 97.79 (97.05) \\
 E50:CR 2           & 77.8 (78.8) & 21.2 (22.2) & 3014 (2128)       & 0.75 (0.53) & 2.8 (3.6)   & 98.86 (99.21) & 97.5 (98.17)  \\
 E50:CR 5           & 77.7 (78.6) & 21.2 (22.0) & 2214 (1168)       & 0.55 (0.29) & 3.4 (5.7)   & 99.18 (99.57) & 98.09 (98.96) \\
 E50:CR 10          & 77.7 (78.3) & 21.1 (21.8) & 1832 (913)        & 0.46 (0.23) & 3.9 (7.0)   & 99.33 (99.67) & 98.4 (99.19)  \\
 \midrule
 R30:Replace        & 77.4 (77.4) & 20.8 (20.8) & 6546.4 (6546.4)   & 1.64 (1.64) & 1.8 (1.8)   & 97.56 (97.56) & 94.53 (94.53) \\
 R30:Majority Vote  & 77.1 (77.1) & 20.5 (20.5) & 3616.4 (3616.4)   & 0.9 (0.9)   & 2.4 (2.4)   & 98.6 (98.6)   & 97.18 (97.18) \\
 R30:MB             & 75.9 (76.3) & 19.4 (19.7) & 2186.2 (2200.4)   & 0.55 (0.55) & 3.2 (3.2)   & 99.14 (99.14) & 98.35 (98.34) \\
 R30:MBME           & 75.3 (76.2) & 18.7 (19.6) & 1859.6 (2183.8)   & 0.46 (0.55) & 3.5 (3.2)   & 99.26 (99.14) & 98.65 (98.36) \\
 R30:CR 2           & 75.5 (75.5) & 19.0 (18.9) & 1607.0 (1092.4)   & 0.4 (0.27)  & 4.0 (5.3)   & 99.37 (99.57) & 98.8 (99.19)  \\
 R30:CR 5           & 75.1 (74.7) & 18.5 (18.2) & 1087.2 (564.8)    & 0.27 (0.14) & 5.3 (9.1)   & 99.58 (99.78) & 99.17 (99.57) \\
 R30:CR 10          & 74.9 (74.2) & 18.4 (17.7) & 875.4 (434.2)     & 0.22 (0.11) & 6.2 (11.2)  & 99.66 (99.83) & 99.33 (99.68) \\
 E30:Replace        & 78.5 (79.3) & 22.0 (22.8) & 9708 (8900)       & 2.43 (2.23) & 1.6 (1.6)   & 96.53 (96.82) & 91.01 (91.69) \\
 E30:Majority Vote  & 78.5 (79.2) & 22.0 (22.6) & 5232 (4989)       & 1.31 (1.25) & 2.1 (2.1)   & 98.03 (98.12) & 95.63 (95.79) \\
 E30:MB             & 78.1 (78.9) & 21.6 (22.4) & 3375 (3180)       & 0.84 (0.8)  & 2.6 (2.8)   & 98.71 (98.79) & 97.25 (97.37) \\
 E30:MBME           & 77.8 (78.8) & 21.2 (22.2) & 2577 (3071)       & 0.64 (0.77) & 3.1 (2.8)   & 98.98 (98.83) & 98.04 (97.49) \\
 E30:CR 2           & 78.0 (78.7) & 21.5 (22.1) & 2578 (1664)       & 0.64 (0.42) & 3.1 (4.3)   & 99.02 (99.37) & 97.86 (98.58) \\
 E30:CR 5           & 78.0 (78.3) & 21.5 (21.8) & 1831 (890)        & 0.46 (0.22) & 3.9 (7.1)   & 99.32 (99.67) & 98.43 (99.21) \\
 E30:CR 10          & 77.9 (78.1) & 21.4 (21.6) & 1517 (707)        & 0.38 (0.18) & 4.5 (8.6)   & 99.44 (99.74) & 98.67 (99.37) \\
 \midrule
 R20:Replace        & 75.7 (75.7) & 19.1 (19.1) & 4171.0 (4171.0)   & 1.04 (1.04) & 2.1 (2.1)   & 98.41 (98.41) & 96.71 (96.71) \\
 R20:Majority Vote  & 75.4 (75.4) & 18.9 (18.9) & 2690.2 (2690.2)   & 0.67 (0.67) & 2.8 (2.8)   & 98.95 (98.95) & 97.99 (97.99) \\
 R20:MB             & 74.0 (74.2) & 17.5 (17.6) & 1662.8 (1655.4)   & 0.42 (0.41) & 3.6 (3.7)   & 99.34 (99.34) & 98.81 (98.81) \\
 R20:MBME           & 73.5 (74.1) & 16.9 (17.6) & 1480.6 (1654.0)   & 0.37 (0.41) & 3.9 (3.7)   & 99.4 (99.34)  & 98.97 (98.82) \\
 R20:CR 2           & 73.2 (72.5) & 16.7 (15.9) & 1205.6 (786.2)    & 0.3 (0.2)   & 4.5 (6.1)   & 99.52 (99.68) & 99.15 (99.45) \\
 R20:CR 5           & 72.3 (71.3) & 15.8 (14.7) & 768.0 (388.8)     & 0.19 (0.1)  & 6.1 (10.5)  & 99.69 (99.84) & 99.46 (99.73) \\
 R20:CR 10          & 71.9 (70.5) & 15.4 (14.0) & 603.8 (280.2)     & 0.15 (0.07) & 7.4 (13.5)  & 99.76 (99.89) & 99.57 (99.81) \\
 E20:Replace        & 78.5 (79.0) & 22.0 (22.5) & 6191 (5430)       & 1.55 (1.36) & 1.9 (2.0)   & 97.74 (98.02) & 94.47 (95.16) \\
 E20:Majority Vote  & 78.3 (78.7) & 21.8 (22.1) & 4295 (3788)       & 1.07 (0.95) & 2.3 (2.5)   & 98.37 (98.56) & 96.46 (96.89) \\
 E20:MB             & 77.9 (78.1) & 21.3 (21.6) & 2700 (2371)       & 0.68 (0.59) & 3.0 (3.3)   & 98.96 (99.09) & 97.86 (98.11) \\
 E20:MBME           & 77.6 (78.1) & 21.1 (21.6) & 2183 (2371)       & 0.55 (0.59) & 3.4 (3.3)   & 99.13 (99.09) & 98.39 (98.11) \\
 E20:CR 2           & 77.8 (77.8) & 21.3 (21.3) & 1999 (1197)       & 0.5 (0.3)   & 3.7 (5.4)   & 99.23 (99.54) & 98.41 (99.04) \\
 E20:CR 5           & 77.7 (77.4) & 21.2 (20.8) & 1383 (599)        & 0.35 (0.15) & 4.8 (9.7)   & 99.47 (99.77) & 98.87 (99.49) \\
 E20:CR 10          & 77.7 (76.7) & 21.2 (20.2) & 1101 (461)        & 0.28 (0.12) & 5.8 (11.9)  & 99.58 (99.83) & 99.08 (99.62) \\
 \midrule
 R10:Replace        & 71.3 (71.3) & 14.7 (14.7) & 1958.4 (1958.4)   & 0.49 (0.49) & 2.9 (2.9)   & 99.22 (99.22) & 98.63 (98.63) \\
 R10:Majority Vote  & 71.2 (71.2) & 14.7 (14.7) & 1481.4 (1481.4)   & 0.37 (0.37) & 3.5 (3.5)   & 99.4 (99.4)   & 98.98 (98.98) \\
 R10:MB             & 69.0 (69.0) & 12.5 (12.4) & 991.4 (963.6)     & 0.25 (0.24) & 4.1 (4.2)   & 99.59 (99.6)  & 99.35 (99.36) \\
 R10:MBME           & 68.9 (69.0) & 12.4 (12.4) & 944.8 (976.8)     & 0.24 (0.24) & 4.3 (4.2)   & 99.61 (99.6)  & 99.38 (99.35) \\
 R10:CR 2           & 67.9 (66.0) & 11.4 (9.5)  & 703.6 (432.2)     & 0.18 (0.11) & 5.0 (6.5)   & 99.71 (99.82) & 99.54 (99.73) \\
 R10:CR 5           & 66.2 (64.2) & 9.6 (7.6)   & 393.4 (147.0)     & 0.1 (0.04)  & 7.1 (14.0)  & 99.84 (99.94) & 99.75 (99.91) \\
 R10:CR 10          & 65.3 (63.5) & 8.8 (7.0)   & 282.6 (104.8)     & 0.07 (0.03) & 8.7 (17.6)  & 99.88 (99.96) & 99.82 (99.93) \\
 E10:Replace        & 76.1 (77.3) & 19.5 (20.8) & 2468 (2320)       & 0.62 (0.58) & 3.0 (3.2)   & 99.04 (99.11) & 98.12 (98.12) \\
 E10:Majority Vote  & 75.9 (77.0) & 19.3 (20.5) & 2417 (2235)       & 0.6 (0.56)  & 3.0 (3.3)   & 99.06 (99.14) & 98.18 (98.22) \\
 E10:MB             & 75.3 (76.4) & 18.8 (19.9) & 1557 (1394)       & 0.39 (0.35) & 4.0 (4.6)   & 99.38 (99.44) & 98.89 (98.98) \\
 E10:MBME           & 75.2 (76.4) & 18.7 (19.9) & 1533 (1394)       & 0.38 (0.35) & 4.0 (4.6)   & 99.38 (99.44) & 98.92 (98.98) \\
 E10:CR 2           & 75.3 (76.1) & 18.7 (19.5) & 1118 (618)        & 0.28 (0.15) & 5.2 (8.9)   & 99.55 (99.75) & 99.22 (99.56) \\
 E10:CR 5           & 75.2 (75.2) & 18.6 (18.6) & 700 (246)         & 0.18 (0.06) & 7.7 (19.9)  & 99.72 (99.9)  & 99.51 (99.81) \\
 E10:CR 10          & 75.2 (73.3) & 18.6 (16.8) & 515 (170)         & 0.13 (0.04) & 10.1 (25.7) & 99.79 (99.93) & 99.64 (99.87) \\
\bottomrule
\end{tabular}
\hspace*{1.0cm}

\begin{tabular}{l|rrrrrrr}
\toprule
 \textbf{Strategy} (Budget \%)         &   \textbf{Acc} (\%) &   $\Delta$\textbf{Acc} (\%)&   $\Sigma$ \textbf{\NF} &   \textbf{\NFR}\, (\%) &  \textbf{ \PF}\, / \textbf{\NF}  &  \textbf{Avg. BTC} &  \textbf{Avg. BEC}\\
\midrule\midrule
100:Oracle         & 50.5 (50.5) & 42.6 (42.6) & 0 (0)           & 0.0 (0.0)   & NaN (NaN)   & 100.0 (100.0) & 99.99 (99.99) \\
 R100:Replace       & 31.9 (31.9) & 24.0 (24.0) & 16669 (16669)   & 5.62 (5.62) & 1.3 (1.3)   & 72.65 (72.65) & 92.61 (92.61) \\
 R100:Majority Vote & 29.6 (29.6) & 21.6 (21.6) & 4690 (4690)     & 1.58 (1.58) & 1.9 (1.9)   & 89.99 (89.99) & 98.02 (98.02) \\
 E100:MB            & 29.1 (29.1) & 21.2 (21.1) & 2477 (2675)     & 0.83 (0.9)  & 2.6 (2.5)   & 94.46 (94.04) & 98.96 (98.88) \\
 E100:MBME          & 28.6 (28.7) & 20.6 (20.7) & 1599 (2267)     & 0.54 (0.76) & 3.4 (2.7)   & 95.86 (94.63) & 99.34 (99.06) \\
 E100:CR 2          & 29.0 (28.8) & 21.0 (20.9) & 1876 (1611)     & 0.63 (0.54) & 3.1 (3.4)   & 95.92 (96.54) & 99.21 (99.32) \\
 E100:CR 5          & 28.8 (28.5) & 20.8 (20.6) & 1372 (853)      & 0.46 (0.29) & 3.8 (5.5)   & 97.18 (98.28) & 99.41 (99.63) \\
 E100:CR 10         & 28.7 (28.2) & 20.8 (20.3) & 1084 (660)      & 0.37 (0.22) & 4.6 (6.7)   & 97.82 (98.68) & 99.54 (99.72) \\
 \midrule
 R50:Replace        & 30.7 (30.7) & 22.7 (22.7) & 7583.8 (7583.8) & 2.56 (2.56) & 1.6 (1.6)   & 86.63 (86.63) & 96.69 (96.69) \\
 R50:Majority Vote  & 28.6 (28.6) & 20.7 (20.7) & 3281.6 (3281.6) & 1.11 (1.11) & 2.2 (2.2)   & 93.01 (93.01) & 98.62 (98.62) \\
 R50:MB             & 27.8 (27.7) & 19.9 (19.8) & 1676.6 (1780.4) & 0.56 (0.6)  & 3.2 (3.1)   & 96.13 (95.9)  & 99.3 (99.26)  \\
 R50:MBME           & 26.9 (27.4) & 18.9 (19.4) & 1280.4 (1682.0) & 0.43 (0.57) & 3.7 (3.1)   & 96.77 (96.04) & 99.48 (99.31) \\
 R50:CR 2           & 27.5 (27.3) & 19.6 (19.4) & 1165.4 (923.0)  & 0.39 (0.31) & 4.1 (4.9)   & 97.3 (97.86)  & 99.52 (99.62) \\
 R50:CR 5           & 27.2 (26.6) & 19.2 (18.7) & 780.2 (469.6)   & 0.26 (0.16) & 5.6 (8.4)   & 98.26 (98.98) & 99.67 (99.8)  \\
 R50:CR 10          & 27.0 (26.2) & 19.1 (18.3) & 599.8 (353.2)   & 0.2 (0.12)  & 6.9 (10.6)  & 98.7 (99.22)  & 99.75 (99.85) \\
 E50:Replace        & 31.0 (31.3) & 23.1 (23.4) & 7882 (7730)     & 2.66 (2.6)  & 1.5 (1.6)   & 86.26 (86.78) & 96.53 (96.59) \\
 E50:Majority Vote  & 29.5 (29.6) & 21.5 (21.7) & 3895 (3759)     & 1.31 (1.27) & 2.0 (2.1)   & 91.73 (92.2)  & 98.36 (98.41) \\
 E50:MB             & 28.9 (29.1) & 20.9 (21.2) & 2079 (2081)     & 0.7 (0.7)   & 2.9 (2.9)   & 95.23 (95.34) & 99.13 (99.13) \\
 E50:MBME           & 28.2 (28.6) & 20.3 (20.7) & 1452 (1944)     & 0.49 (0.66) & 3.6 (3.0)   & 96.18 (95.52) & 99.41 (99.19) \\
 E50:CR 2           & 28.7 (28.7) & 20.8 (20.8) & 1506 (1136)     & 0.51 (0.38) & 3.6 (4.4)   & 96.64 (97.54) & 99.37 (99.52) \\
 E50:CR 5           & 28.5 (28.4) & 20.6 (20.4) & 1049 (569)      & 0.35 (0.19) & 4.6 (7.7)   & 97.83 (98.88) & 99.55 (99.75) \\
 E50:CR 10          & 28.4 (28.1) & 20.5 (20.2) & 828 (430)       & 0.28 (0.14) & 5.6 (9.7)   & 98.35 (99.16) & 99.64 (99.81) \\
 \midrule
 R30:Replace        & 29.0 (29.0) & 21.0 (21.0) & 4070.6 (4070.6) & 1.37 (1.37) & 2.0 (2.0)   & 92.19 (92.19) & 98.26 (98.26) \\
 R30:Majority Vote  & 27.3 (27.3) & 19.4 (19.4) & 2346.6 (2346.6) & 0.79 (0.79) & 2.5 (2.5)   & 94.91 (94.91) & 99.02 (99.02) \\
 R30:MB             & 25.9 (25.9) & 18.0 (17.9) & 1185.8 (1218.6) & 0.4 (0.41)  & 3.8 (3.7)   & 97.12 (97.01) & 99.52 (99.5)  \\
 R30:MBME           & 25.1 (25.7) & 17.2 (17.8) & 1008.4 (1208.0) & 0.34 (0.41) & 4.2 (3.7)   & 97.41 (97.03) & 99.59 (99.51) \\
 R30:CR 2           & 25.4 (24.9) & 17.5 (16.9) & 782.2 (601.2)   & 0.26 (0.2)  & 5.2 (6.2)   & 98.05 (98.46) & 99.68 (99.76) \\
 R30:CR 5           & 24.6 (23.9) & 16.6 (16.0) & 476.2 (275.4)   & 0.16 (0.09) & 7.5 (11.8)  & 98.8 (99.32)  & 99.81 (99.89) \\
 R30:CR 10          & 24.4 (23.5) & 16.4 (15.5) & 331.6 (190.8)   & 0.11 (0.06) & 10.2 (16.1) & 99.17 (99.53) & 99.86 (99.92) \\
 E30:Replace        & 29.0 (29.7) & 21.0 (21.7) & 4316 (4093)     & 1.45 (1.38) & 1.9 (2.0)   & 91.75 (92.39) & 98.14 (98.23) \\
 E30:Majority Vote  & 28.2 (28.6) & 20.3 (20.6) & 2970 (2703)     & 1.0 (0.91)  & 2.3 (2.4)   & 93.54 (94.23) & 98.76 (98.87) \\
 E30:MB             & 27.8 (27.6) & 19.9 (19.7) & 1565 (1402)     & 0.53 (0.47) & 3.4 (3.6)   & 96.24 (96.66) & 99.35 (99.42) \\
 E30:MBME           & 26.9 (27.6) & 18.9 (19.7) & 1280 (1402)     & 0.43 (0.47) & 3.7 (3.6)   & 96.64 (96.66) & 99.48 (99.42) \\
 E30:CR 2           & 27.7 (26.9) & 19.7 (19.0) & 1074 (675)      & 0.36 (0.23) & 4.4 (6.2)   & 97.42 (98.32) & 99.55 (99.72) \\
 E30:CR 5           & 27.4 (26.0) & 19.4 (18.1) & 689 (312)       & 0.23 (0.11) & 6.2 (11.8)  & 98.43 (99.31) & 99.71 (99.87) \\
 E30:CR 10          & 27.1 (25.4) & 19.2 (17.5) & 504 (214)       & 0.17 (0.07) & 8.1 (16.2)  & 98.91 (99.53) & 99.79 (99.91) \\
 \midrule
 R20:Replace        & 27.0 (27.0) & 19.1 (19.1) & 2453.8 (2453.8) & 0.83 (0.83) & 2.4 (2.4)   & 94.82 (94.82) & 98.97 (98.97) \\
 R20:Majority Vote  & 25.8 (25.8) & 17.8 (17.8) & 1658.8 (1658.8) & 0.56 (0.56) & 3.0 (3.0)   & 96.17 (96.17) & 99.32 (99.32) \\
 R20:MB             & 23.8 (23.7) & 15.8 (15.7) & 887.2 (892.2)   & 0.3 (0.3)   & 4.3 (4.3)   & 97.74 (97.69) & 99.64 (99.64) \\
 R20:MBME           & 23.5 (23.9) & 15.6 (15.9) & 793.8 (866.2)   & 0.27 (0.29) & 4.6 (4.4)   & 97.93 (97.77) & 99.68 (99.65) \\
 R20:CR 2           & 22.6 (21.8) & 14.7 (13.8) & 530.0 (392.2)   & 0.18 (0.13) & 6.2 (7.6)   & 98.57 (98.89) & 99.79 (99.84) \\
 R20:CR 5           & 21.5 (20.4) & 13.5 (12.5) & 304.2 (171.2)   & 0.1 (0.06)  & 9.3 (14.5)  & 99.14 (99.52) & 99.88 (99.93) \\
 R20:CR 10          & 20.9 (19.9) & 12.9 (11.9) & 215.6 (121.4)   & 0.07 (0.04) & 12.1 (19.2) & 99.4 (99.65)  & 99.91 (99.95) \\
 E20:Replace        & 26.9 (27.5) & 19.0 (19.5) & 2588 (2480)     & 0.87 (0.84) & 2.4 (2.5)   & 94.54 (95.12) & 98.91 (98.94) \\
 E20:Majority Vote  & 26.2 (26.6) & 18.3 (18.7) & 2390 (2252)     & 0.81 (0.76) & 2.4 (2.5)   & 94.8 (95.41)  & 99.0 (99.05)  \\
 E20:MB             & 26.0 (26.3) & 18.1 (18.4) & 1169 (1075)     & 0.39 (0.36) & 3.9 (4.2)   & 97.01 (97.35) & 99.53 (99.56) \\
 E20:MBME           & 25.8 (26.3) & 17.8 (18.4) & 1131 (1075)     & 0.38 (0.36) & 3.9 (4.2)   & 97.06 (97.35) & 99.54 (99.56) \\
 E20:CR 2           & 26.0 (26.2) & 18.1 (18.2) & 670 (490)       & 0.23 (0.17) & 6.0 (7.9)   & 98.14 (98.71) & 99.73 (99.8)  \\
 E20:CR 5           & 26.0 (25.6) & 18.0 (17.7) & 369 (193)       & 0.12 (0.07) & 10.1 (18.0) & 98.97 (99.55) & 99.85 (99.92) \\
 E20:CR 10          & 25.9 (24.9) & 17.9 (16.9) & 248 (134)       & 0.08 (0.05) & 14.4 (24.4) & 99.34 (99.69) & 99.9 (99.94)  \\
 \midrule
 R10:Replace        & 22.1 (22.1) & 14.2 (14.2) & 996.6 (996.6)   & 0.34 (0.34) & 3.6 (3.6)   & 97.49 (97.49) & 99.6 (99.6)   \\
 R10:Majority Vote  & 21.6 (21.6) & 13.6 (13.6) & 808.8 (808.8)   & 0.27 (0.27) & 4.1 (4.1)   & 97.86 (97.86) & 99.68 (99.68) \\
 R10:MB             & 19.5 (19.3) & 11.6 (11.4) & 450.2 (442.4)   & 0.15 (0.15) & 5.8 (5.8)   & 98.7 (98.71)  & 99.82 (99.83) \\
 R10:MBME           & 19.1 (19.1) & 11.1 (11.2) & 446.4 (449.0)   & 0.15 (0.15) & 5.6 (5.6)   & 98.69 (98.69) & 99.83 (99.82) \\
 R10:CR 2           & 17.7 (16.1) & 9.8 (8.1)   & 287.2 (196.6)   & 0.1 (0.07)  & 7.3 (8.7)   & 99.11 (99.34) & 99.89 (99.92) \\
 R10:CR 5           & 15.6 (14.1) & 7.7 (6.2)   & 151.6 (74.6)    & 0.05 (0.03) & 10.4 (16.4) & 99.5 (99.75)  & 99.94 (99.97) \\
 R10:CR 10          & 14.6 (13.5) & 6.7 (5.5)   & 94.6 (46.0)     & 0.03 (0.02) & 14.1 (23.4) & 99.68 (99.84) & 99.96 (99.98) \\
 E10:Replace        & 23.7 (25.7) & 15.7 (17.8) & 996 (849)       & 0.34 (0.29) & 3.9 (4.9)   & 97.5 (97.9)   & 99.6 (99.65)  \\
 E10:Majority Vote  & 23.7 (25.7) & 15.7 (17.7) & 996 (849)       & 0.34 (0.29) & 3.9 (4.9)   & 97.5 (97.9)   & 99.6 (99.65)  \\
 E10:MB             & 22.7 (22.3) & 14.8 (14.3) & 696 (560)       & 0.23 (0.19) & 4.9 (5.8)   & 98.08 (98.39) & 99.72 (99.78) \\
 E10:MBME           & 22.7 (22.3) & 14.8 (14.3) & 696 (560)       & 0.23 (0.19) & 4.9 (5.8)   & 98.08 (98.39) & 99.72 (99.78) \\
 E10:CR 2           & 20.7 (17.8) & 12.8 (9.9)  & 427 (213)       & 0.14 (0.07) & 6.6 (9.6)   & 98.68 (99.21) & 99.83 (99.92) \\
 E10:CR 5           & 18.2 (15.7) & 10.3 (7.7)  & 197 (49)        & 0.07 (0.02) & 10.7 (30.3) & 99.32 (99.82) & 99.92 (99.98) \\
 E10:CR 10          & 17.2 (14.9) & 9.2 (7.0)   & 122 (24)        & 0.04 (0.01) & 15.0 (54.9) & 99.57 (99.91) & 99.95 (99.99) \\\bottomrule
\end{tabular}}
\label{tab:app:imagenet_objectnet_improving}
\end{table}

\begin{table}
\small
\caption{Full results of all experiments on \textbf{CIFAR-10} under the standard setting of an \textbf{improving sequence of models} and \textbf{estimating confusion matrices on a separate split} of the data (i.e., not on the target data set) as explained in~\cref{sec:experimental_setup}. The first number in each cell refers to estimating only the diagonal elements (i.e., the class-specific accuracies) of confusion matrices, and the \textbf{second number (in brackets) refers to estimating the full confusion matrices} with smoothing. The character \textbf{E} in front of a budget indicates that selection for re-evaluation is based on our posterior label \textbf{entropy criterion} (e.g., E30 refers to selecting the top 30\% samples with highest entropy), and the character \textbf{R} indicates selecting a \textbf{randomly} sampled subset without replacement. Note that entropy-based selection requires a posterior and is thus not applicable for the baselines Replace and Majority Vote.}
\vskip 0.1in
\centering
\resizebox{0.7\columnwidth}{!}{
\begin{tabular}{l|rrrrrrr}
\toprule
 \textbf{Strategy} (Budget \%)         &   \textbf{Acc} (\%) &   $\Delta$\textbf{Acc} (\%)&   $\Sigma$ \textbf{\NF} &   \textbf{\NFR}\, (\%) &  \textbf{ \PF}\, / \textbf{\NF} &  \textbf{Avg. BTC} &  \textbf{Avg. BEC}\\
\midrule \midrule
 100:Oracle         & 99.3 (99.3) & 6.1 (6.1)   & 0 (0)         & 0.0 (0.0)   & NaN (NaN)   & 100.0 (100.0) & 100.0 (100.0) \\
 R100:Replace       & 94.7 (94.7) & 1.5 (1.5)   & 1418 (1418)   & 2.58 (2.58) & 1.1 (1.1)   & 97.26 (97.26) & 54.79 (54.79) \\
 R100:Majority Vote & 96.1 (96.1) & 2.9 (2.9)   & 502 (502)     & 0.91 (0.91) & 1.3 (1.3)   & 99.03 (99.03) & 81.94 (81.94) \\
 E100:MB            & 95.9 (95.9) & 2.8 (2.8)   & 284 (340)     & 0.52 (0.62) & 1.5 (1.4)   & 99.45 (99.34) & 89.64 (87.71) \\
 E100:MBME          & 96.1 (96.0) & 2.9 (2.8)   & 119 (146)     & 0.22 (0.27) & 2.2 (2.0)   & 99.77 (99.72) & 95.28 (94.47) \\
 E100:CR 2          & 96.0 (96.0) & 2.8 (2.8)   & 219 (170)     & 0.4 (0.31)  & 1.6 (1.8)   & 99.58 (99.67) & 91.69 (93.28) \\
 E100:CR 5          & 95.9 (95.9) & 2.8 (2.8)   & 132 (126)     & 0.24 (0.23) & 2.1 (2.1)   & 99.75 (99.76) & 94.65 (94.98) \\
 E100:CR 10         & 96.0 (96.0) & 2.9 (2.8)   & 112 (103)     & 0.2 (0.19)  & 2.3 (2.4)   & 99.79 (99.8)  & 95.49 (95.92) \\
 \midrule
 R50:Replace        & 94.7 (94.7) & 1.6 (1.6)   & 711.4 (711.4) & 1.29 (1.29) & 1.1 (1.1)   & 98.62 (98.62) & 77.8 (77.8)   \\
 R50:Majority Vote  & 95.7 (95.7) & 2.6 (2.6)   & 327.4 (327.4) & 0.6 (0.6)   & 1.4 (1.4)   & 99.37 (99.37) & 89.06 (89.06) \\
 R50:MB             & 95.8 (95.8) & 2.7 (2.6)   & 196.4 (227.8) & 0.36 (0.41) & 1.7 (1.6)   & 99.62 (99.56) & 93.41 (92.47) \\
 R50:MBME           & 95.7 (95.6) & 2.5 (2.5)   & 69.8 (90.6)   & 0.13 (0.16) & 2.8 (2.4)   & 99.87 (99.83) & 97.58 (96.93) \\
 R50:CR 2           & 95.8 (95.8) & 2.6 (2.6)   & 133.8 (104.2) & 0.24 (0.19) & 2.0 (2.3)   & 99.74 (99.8)  & 95.32 (96.24) \\
 R50:CR 5           & 95.7 (95.7) & 2.5 (2.5)   & 73.2 (69.0)   & 0.13 (0.13) & 2.7 (2.8)   & 99.86 (99.87) & 97.4 (97.59)  \\
 R50:CR 10          & 95.7 (95.7) & 2.6 (2.5)   & 61.0 (60.6)   & 0.11 (0.11) & 3.1 (3.1)   & 99.88 (99.88) & 97.83 (97.85) \\
 E50:Replace        & 95.2 (95.2) & 2.0 (2.0)   & 1019 (1012)   & 1.85 (1.84) & 1.1 (1.1)   & 98.04 (98.05) & 64.97 (64.69) \\
 E50:Majority Vote  & 96.1 (96.0) & 2.9 (2.9)   & 428 (419)     & 0.78 (0.76) & 1.3 (1.3)   & 99.18 (99.19) & 84.48 (84.84) \\
 E50:MB             & 96.1 (96.0) & 2.9 (2.8)   & 260 (289)     & 0.47 (0.53) & 1.6 (1.5)   & 99.5 (99.44)  & 90.63 (89.8)  \\
 E50:MBME           & 96.1 (96.0) & 3.0 (2.9)   & 105 (134)     & 0.19 (0.24) & 2.4 (2.1)   & 99.8 (99.74)  & 96.04 (95.09) \\
 E50:CR 2           & 96.0 (96.0) & 2.8 (2.8)   & 206 (149)     & 0.37 (0.27) & 1.7 (1.9)   & 99.6 (99.71)  & 92.4 (94.27)  \\
 E50:CR 5           & 96.0 (95.9) & 2.9 (2.8)   & 123 (112)     & 0.22 (0.2)  & 2.2 (2.2)   & 99.76 (99.79) & 95.18 (95.67) \\
 E50:CR 10          & 96.1 (96.0) & 3.0 (2.8)   & 99 (95)       & 0.18 (0.17) & 2.5 (2.5)   & 99.81 (99.82) & 96.21 (96.35) \\
 \midrule
 R30:Replace        & 94.6 (94.6) & 1.4 (1.4)   & 439.6 (439.6) & 0.8 (0.8)   & 1.2 (1.2)   & 99.15 (99.15) & 86.74 (86.74) \\
 R30:Majority Vote  & 95.3 (95.3) & 2.2 (2.2)   & 238.4 (238.4) & 0.43 (0.43) & 1.5 (1.5)   & 99.54 (99.54) & 92.54 (92.54) \\
 R30:MB             & 95.2 (95.3) & 2.1 (2.1)   & 146.2 (163.2) & 0.27 (0.3)  & 1.7 (1.7)   & 99.72 (99.68) & 95.46 (94.91) \\
 R30:MBME           & 95.2 (95.2) & 2.0 (2.1)   & 47.0 (64.2)   & 0.09 (0.12) & 3.2 (2.6)   & 99.91 (99.88) & 98.47 (97.92) \\
 R30:CR 2           & 95.2 (95.2) & 2.0 (2.0)   & 88.8 (64.6)   & 0.16 (0.12) & 2.1 (2.6)   & 99.83 (99.88) & 97.16 (97.92) \\
 R30:CR 5           & 95.1 (95.0) & 1.9 (1.9)   & 45.6 (47.0)   & 0.08 (0.09) & 3.1 (3.0)   & 99.91 (99.91) & 98.54 (98.49) \\
 R30:CR 10          & 95.2 (95.1) & 2.0 (2.0)   & 35.4 (35.4)   & 0.06 (0.06) & 3.8 (3.8)   & 99.93 (99.93) & 98.85 (98.85) \\
 E30:Replace        & 95.6 (95.4) & 2.5 (2.2)   & 688 (689)     & 1.25 (1.25) & 1.2 (1.2)   & 98.68 (98.68) & 75.18 (75.12) \\
 E30:Majority Vote  & 95.9 (95.9) & 2.8 (2.8)   & 380 (395)     & 0.69 (0.72) & 1.4 (1.4)   & 99.27 (99.24) & 86.44 (85.83) \\
 E30:MB             & 96.0 (96.1) & 2.8 (2.9)   & 253 (277)     & 0.46 (0.5)  & 1.6 (1.5)   & 99.51 (99.47) & 91.14 (90.39) \\
 E30:MBME           & 96.0 (96.2) & 2.8 (3.0)   & 96 (123)      & 0.17 (0.22) & 2.5 (2.2)   & 99.82 (99.76) & 96.45 (95.68) \\
 E30:CR 2           & 96.0 (96.2) & 2.8 (3.0)   & 192 (147)     & 0.35 (0.27) & 1.7 (2.0)   & 99.63 (99.72) & 93.12 (94.5)  \\
 E30:CR 5           & 96.0 (96.2) & 2.8 (3.0)   & 103 (98)      & 0.19 (0.18) & 2.4 (2.6)   & 99.8 (99.81)  & 96.07 (96.36) \\
 E30:CR 10          & 96.0 (96.2) & 2.9 (3.0)   & 89 (86)       & 0.16 (0.16) & 2.6 (2.7)   & 99.83 (99.83) & 96.67 (96.85) \\
 \midrule
 R20:Replace        & 94.5 (94.5) & 1.3 (1.3)   & 274.6 (274.6) & 0.5 (0.5)   & 1.2 (1.2)   & 99.47 (99.47) & 91.94 (91.94) \\
 R20:Majority Vote  & 94.9 (94.9) & 1.7 (1.7)   & 195.2 (195.2) & 0.35 (0.35) & 1.4 (1.4)   & 99.62 (99.62) & 94.06 (94.06) \\
 R20:MB             & 94.7 (94.8) & 1.6 (1.7)   & 107.0 (120.4) & 0.19 (0.22) & 1.7 (1.7)   & 99.79 (99.77) & 96.86 (96.43) \\
 R20:MBME           & 94.6 (94.6) & 1.4 (1.4)   & 31.0 (45.4)   & 0.06 (0.08) & 3.3 (2.6)   & 99.94 (99.91) & 99.09 (98.67) \\
 R20:CR 2           & 94.5 (94.5) & 1.4 (1.4)   & 65.0 (48.6)   & 0.12 (0.09) & 2.0 (2.4)   & 99.87 (99.91) & 98.08 (98.55) \\
 R20:CR 5           & 94.7 (94.7) & 1.5 (1.5)   & 26.6 (27.4)   & 0.05 (0.05) & 3.8 (3.7)   & 99.95 (99.95) & 99.19 (99.16) \\
 R20:CR 10          & 94.6 (94.6) & 1.5 (1.4)   & 24.6 (24.4)   & 0.04 (0.04) & 4.0 (3.9)   & 99.95 (99.95) & 99.25 (99.26) \\
 E20:Replace        & 95.8 (95.7) & 2.6 (2.5)   & 462 (478)     & 0.84 (0.87) & 1.3 (1.3)   & 99.11 (99.08) & 83.35 (83.11) \\
 E20:Majority Vote  & 95.9 (96.0) & 2.8 (2.8)   & 336 (347)     & 0.61 (0.63) & 1.4 (1.4)   & 99.35 (99.33) & 88.2 (87.94)  \\
 E20:MB             & 96.0 (95.9) & 2.9 (2.8)   & 217 (263)     & 0.39 (0.48) & 1.7 (1.5)   & 99.58 (99.49) & 92.45 (91.02) \\
 E20:MBME           & 96.0 (96.0) & 2.9 (2.8)   & 82 (115)      & 0.15 (0.21) & 2.8 (2.2)   & 99.84 (99.78) & 97.02 (96.04) \\
 E20:CR 2           & 96.1 (95.9) & 2.9 (2.8)   & 156 (130)     & 0.28 (0.24) & 1.9 (2.1)   & 99.7 (99.75)  & 94.53 (95.27) \\
 E20:CR 5           & 96.0 (95.9) & 2.9 (2.8)   & 84 (90)       & 0.15 (0.16) & 2.7 (2.5)   & 99.84 (99.83) & 96.86 (96.71) \\
 E20:CR 10          & 96.0 (95.9) & 2.9 (2.8)   & 75 (80)       & 0.14 (0.15) & 2.9 (2.7)   & 99.86 (99.85) & 97.22 (97.11) \\
 \midrule
 R10:Replace        & 94.1 (94.1) & 0.9 (0.9)   & 129.2 (129.2) & 0.23 (0.23) & 1.3 (1.3)   & 99.75 (99.75) & 96.31 (96.31) \\
 R10:Majority Vote  & 94.3 (94.3) & 1.1 (1.1)   & 109.4 (109.4) & 0.2 (0.2)   & 1.5 (1.5)   & 99.79 (99.79) & 96.86 (96.86) \\
 R10:MB             & 94.0 (94.2) & 0.9 (1.0)   & 64.0 (70.8)   & 0.12 (0.13) & 1.7 (1.7)   & 99.88 (99.86) & 98.23 (98.0)  \\
 R10:MBME           & 93.8 (93.8) & 0.6 (0.6)   & 14.2 (21.4)   & 0.03 (0.04) & 3.3 (2.4)   & 99.97 (99.96) & 99.61 (99.41) \\
 R10:CR 2           & 93.8 (93.8) & 0.7 (0.7)   & 35.2 (23.8)   & 0.06 (0.04) & 2.0 (2.4)   & 99.93 (99.95) & 99.03 (99.34) \\
 R10:CR 5           & 93.7 (93.7) & 0.5 (0.6)   & 11.0 (11.0)   & 0.02 (0.02) & 3.5 (3.6)   & 99.98 (99.98) & 99.69 (99.69) \\
 R10:CR 10          & 93.8 (93.8) & 0.7 (0.7)   & 6.8 (7.0)     & 0.01 (0.01) & 6.0 (5.9)   & 99.99 (99.99) & 99.81 (99.8)  \\
 E10:Replace        & 95.3 (95.4) & 2.1 (2.2)   & 206 (201)     & 0.37 (0.37) & 1.5 (1.5)   & 99.6 (99.61)  & 93.45 (93.65) \\
 E10:Majority Vote  & 95.4 (95.4) & 2.2 (2.2)   & 200 (200)     & 0.36 (0.36) & 1.6 (1.6)   & 99.61 (99.61) & 93.67 (93.69) \\
 E10:MB             & 95.3 (95.3) & 2.1 (2.1)   & 140 (145)     & 0.25 (0.26) & 1.8 (1.7)   & 99.73 (99.72) & 95.59 (95.45) \\
 E10:MBME           & 95.2 (95.2) & 2.0 (2.0)   & 35 (66)       & 0.06 (0.12) & 3.9 (2.5)   & 99.93 (99.87) & 98.92 (98.02) \\
 E10:CR 2           & 95.2 (95.2) & 2.1 (2.1)   & 91 (59)       & 0.17 (0.11) & 2.1 (2.7)   & 99.82 (99.89) & 97.26 (98.11) \\
 E10:CR 5           & 95.2 (95.2) & 2.0 (2.0)   & 30 (38)       & 0.05 (0.07) & 4.3 (3.7)   & 99.94 (99.93) & 99.05 (98.79) \\
 E10:CR 10          & 95.2 (95.2) & 2.0 (2.0)   & 29 (36)       & 0.05 (0.07) & 4.4 (3.8)   & 99.94 (99.93) & 99.09 (98.86) \\
\bottomrule
\end{tabular}}
\label{tab:app:cifar10_improving}
\end{table}

\subsection{On Utilizing Uncalibrated Soft Labels}
\begin{figure}[th]
\subfigure[CIFAR-10]{
    \includegraphics[width=0.32\linewidth]{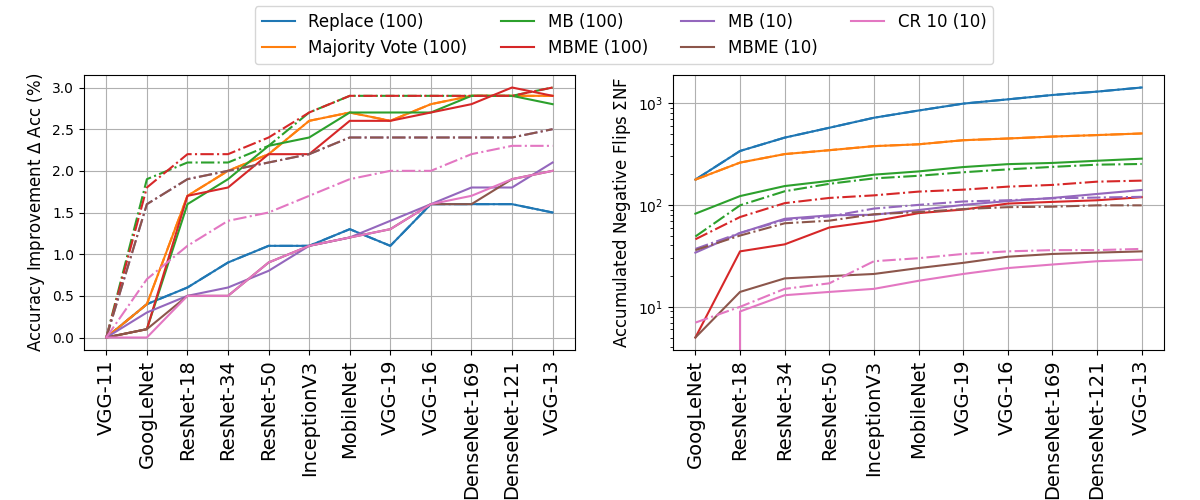}}
\hfill
\subfigure[ImageNet]{
    \includegraphics[width=0.32\linewidth]{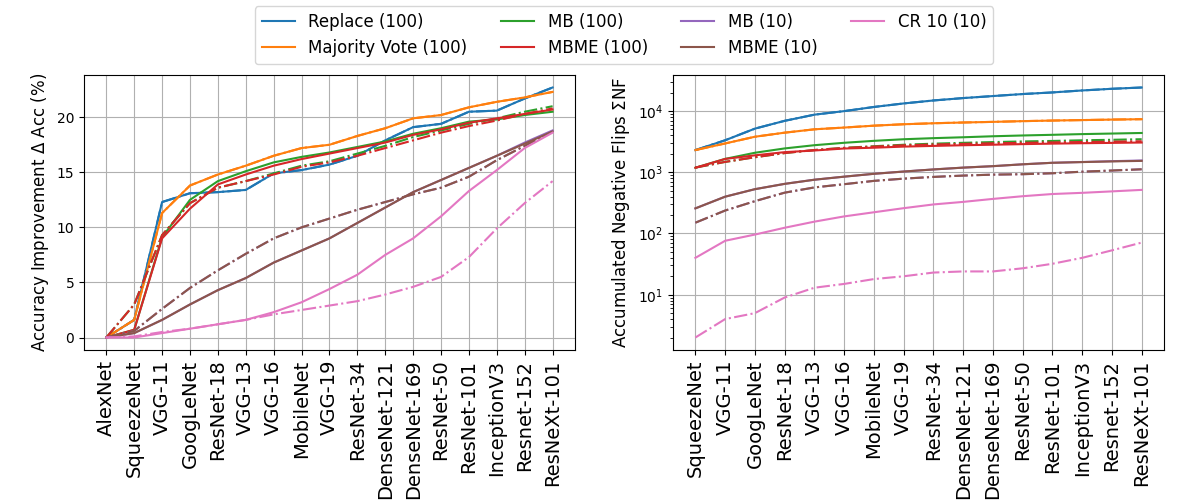}}
\hfill
\subfigure[ObjectNet]{
    \includegraphics[width=0.32\linewidth]{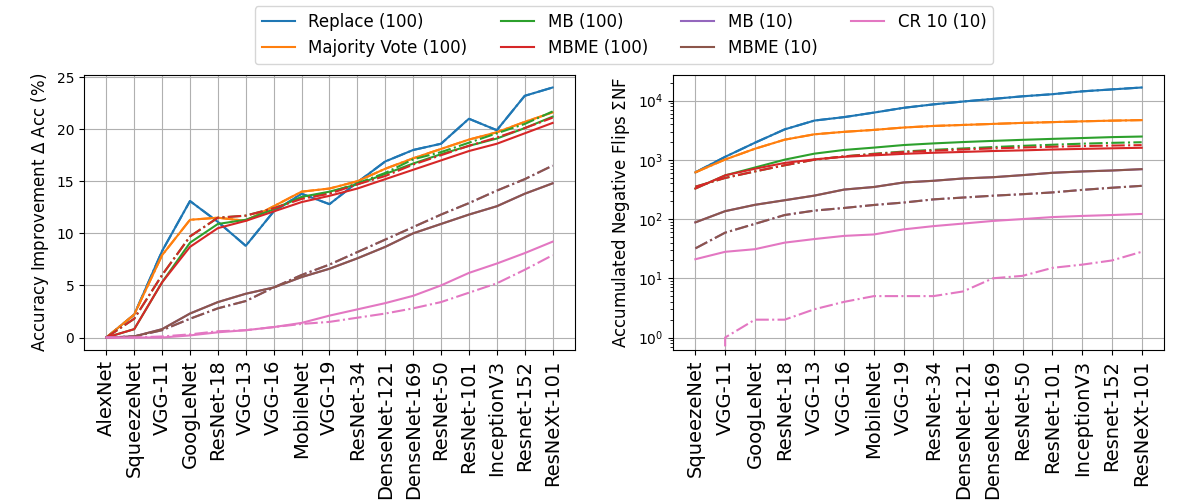}
    }
    \caption{Comparative temporal evolution plots for experiments incorporating uncalibrated softmax labels. Solid lines represent using hard labels with only diagonal estimated for confusion matrix and dashed lines reflect results using soft labels and estimating full confusion matrix with Laplace smoothing.
     }
    \label{fig:app:diagonal_vs_full_confusion_estimation_wo_vs_w_softmax}
\end{figure}

Finally, we tested our method on incroporating the uncalibrated soft labels provided as output in the pre-trained ImageNet models by repeating the ImageNet and ObjectNet experiment (in combination with full confusion matrix estimation).
As already discussed in \Cref{sec:limitations}, deep neural networks are known to have unreliable uncertainty estimates~\citep{mackay1995bayesian, szegedy2013intriguing, goodfellow2014explaining,nguyen2015deep} and we therefore do not expect these results to be representative of what our method could achieve if it has access to truly calibrated soft labels. We now utilize the full output softmax vector $\mathbf{p}^{sl}=C^t(\xb_n)$ and refine our likelihood estimate in eq.\ (2) by multiplying the confusion matrix coefficients with the soft label vector, i.e. $\pi^t(\hat{y}_n^t,k) \rightarrow \sum_i p^{sl}_i \pi^t(i,k)$. We show the temporal evolution compared to the hard-label implementation using only diagonal confusion matrix estimates in  \cref{fig:app:diagonal_vs_full_confusion_estimation_wo_vs_w_softmax} and a full account over the final performance metrics and all selection strategies and budgets in \cref{tab:app:softmax_all} and \cref{tab:app:objectnet_softmax}.  
Overall the results are slightly less consistent, most likely due to the known uncalibrated soft labels, emphasizing the strength of our method when only having deterministic labels. Specifically, on CIFAR-10, there are additional accuracy gains but often we accumulate more negative flips and BTC, BEC are typically slightly worse. On ImageNet, accuracy gains are sometimes worse without finding a clear pattern when this is the case, but negative flips are sometimes significantly lower. BTC, BEC seem to be better on the majority of update strategies and labels. On ObjectNet, we likewise find no clear pattern in the accuracy gains. However, accumulated negative flips are generally much lower and both, BTC and BEC scores are even higher (very close to 1.0) though we want to emphasize that they were already very high for the ObjectNet experiments from the main paper.

\begin{table}
\small
\caption{Full results of all experiments on \textbf{CIFAR-10 (left) and ImageNet (right)} under the standard setting of an \textbf{improving sequence of models} and \textbf{estimating confusion matrices on a separate split} of the data (i.e., not on the target data set) as explained in~\cref{sec:experimental_setup}. The first number in each cell refers to estimating only the diagonal elements (i.e., the class-specific accuracies) of confusion matrices, and the \textbf{second number (in brackets) refers to estimating the full confusion matrices} with smoothing \textbf{and incorporating soft labels}. The character \textbf{E} in front of a budget indicates that selection for re-evaluation is based on our posterior label \textbf{entropy criterion} (e.g., E30 refers to selecting the top 30\% samples with highest entropy), and the character \textbf{R} indicates selecting a \textbf{randomly} sampled subset without replacement. Note that entropy-based selection requires a posterior and is thus not applicable for the baselines Replace and Majority Vote.
}
\vskip 0.3in
\resizebox{1.0\columnwidth}{!}{
\begin{tabular}{l|rrrrrrr}
\toprule
 \textbf{Strategy}  Budget \%   &   \textbf{Acc} (\%) &   $\Delta$\textbf{Acc} (\%)&   $\Sigma$ \textbf{\NF} &   \textbf{\NFR}\, (\%) &  \textbf{ \PF}\, / \textbf{\NF} &  \textbf{Avg. BTC} &  \textbf{Avg. BEC} \\
\midrule\midrule
   100:Oracle         & 99.3 (99.3) & 6.1 (6.1)   & 0 (0)         & 0.0 (0.0)   & NaN (NaN)   & 100.0 (100.0) & 100.0 (100.0) \\
 R100:Replace       & 94.7 (94.7) & 1.5 (1.5)   & 1418 (1418)   & 2.58 (2.58) & 1.1 (1.1)   & 97.26 (97.26) & 54.79 (54.79) \\
 R100:Majority Vote & 96.1 (96.1) & 2.9 (2.9)   & 502 (502)     & 0.91 (0.91) & 1.3 (1.3)   & 99.03 (99.03) & 81.94 (81.94) \\
 E100:MB            & 95.9 (96.2) & 2.8 (3.0)   & 284 (252)     & 0.52 (0.46) & 1.5 (1.6)   & 99.45 (99.52) & 89.64 (89.69) \\
 E100:MBME          & 96.1 (96.1) & 2.9 (3.0)   & 119 (173)     & 0.22 (0.31) & 2.2 (1.9)   & 99.77 (99.67) & 95.28 (92.98) \\
 E100:CR 2          & 96.0 (96.1) & 2.8 (3.0)   & 219 (170)     & 0.4 (0.31)  & 1.6 (1.9)   & 99.58 (99.68) & 91.69 (92.98) \\
 E100:CR 5          & 95.9 (96.2) & 2.8 (3.0)   & 132 (103)     & 0.24 (0.19) & 2.1 (2.5)   & 99.75 (99.8)  & 94.65 (95.68) \\
 E100:CR 10         & 96.0 (96.2) & 2.9 (3.0)   & 112 (73)      & 0.2 (0.13)  & 2.3 (3.1)   & 99.79 (99.86) & 95.49 (96.99) \\
 \midrule
 R50:Replace        & 94.7 (94.7) & 1.6 (1.6)   & 711.4 (711.4) & 1.29 (1.29) & 1.1 (1.1)   & 98.62 (98.62) & 77.8 (77.8)   \\
 R50:Majority Vote  & 95.7 (95.7) & 2.6 (2.6)   & 327.4 (327.4) & 0.6 (0.6)   & 1.4 (1.4)   & 99.37 (99.37) & 89.06 (89.06) \\
 R50:MB             & 95.8 (96.0) & 2.7 (2.8)   & 196.4 (155.0) & 0.36 (0.28) & 1.7 (1.9)   & 99.62 (99.7)  & 93.41 (94.27) \\
 R50:MBME           & 95.7 (95.9) & 2.5 (2.7)   & 69.8 (106.0)  & 0.13 (0.19) & 2.8 (2.3)   & 99.87 (99.8)  & 97.58 (96.2)  \\
 R50:CR 2           & 95.8 (96.0) & 2.6 (2.9)   & 133.8 (99.8)  & 0.24 (0.18) & 2.0 (2.4)   & 99.74 (99.81) & 95.32 (96.25) \\
 R50:CR 5           & 95.7 (96.0) & 2.5 (2.8)   & 73.2 (59.2)   & 0.13 (0.11) & 2.7 (3.4)   & 99.86 (99.89) & 97.4 (97.77)  \\
 R50:CR 10          & 95.7 (95.9) & 2.6 (2.7)   & 61.0 (40.8)   & 0.11 (0.07) & 3.1 (4.3)   & 99.88 (99.92) & 97.83 (98.47) \\
 E50:Replace        & 95.2 (95.2) & 2.0 (2.0)   & 1019 (1012)   & 1.85 (1.84) & 1.1 (1.1)   & 98.04 (98.05) & 64.97 (64.69) \\
 E50:Majority Vote  & 96.1 (96.0) & 2.9 (2.9)   & 428 (419)     & 0.78 (0.76) & 1.3 (1.3)   & 99.18 (99.19) & 84.48 (84.84) \\
 E50:MB             & 96.1 (96.2) & 2.9 (3.0)   & 260 (257)     & 0.47 (0.47) & 1.6 (1.6)   & 99.5 (99.51)  & 90.63 (89.36) \\
 E50:MBME           & 96.1 (96.1) & 3.0 (3.0)   & 105 (170)     & 0.19 (0.31) & 2.4 (1.9)   & 99.8 (99.67)  & 96.04 (93.08) \\
 E50:CR 2           & 96.0 (96.1) & 2.8 (3.0)   & 206 (170)     & 0.37 (0.31) & 1.7 (1.9)   & 99.6 (99.68)  & 92.4 (92.92)  \\
 E50:CR 5           & 96.0 (96.2) & 2.9 (3.0)   & 123 (102)     & 0.22 (0.19) & 2.2 (2.5)   & 99.76 (99.81) & 95.18 (95.72) \\
 E50:CR 10          & 96.1 (96.2) & 3.0 (3.0)   & 99 (71)       & 0.18 (0.13) & 2.5 (3.1)   & 99.81 (99.86) & 96.21 (97.07) \\
 \midrule
 R30:Replace        & 94.6 (94.6) & 1.4 (1.4)   & 439.6 (439.6) & 0.8 (0.8)   & 1.2 (1.2)   & 99.15 (99.15) & 86.74 (86.74) \\
 R30:Majority Vote  & 95.3 (95.3) & 2.2 (2.2)   & 238.4 (238.4) & 0.43 (0.43) & 1.5 (1.5)   & 99.54 (99.54) & 92.54 (92.54) \\
 R30:MB             & 95.2 (95.6) & 2.1 (2.4)   & 146.2 (117.6) & 0.27 (0.21) & 1.7 (2.0)   & 99.72 (99.77) & 95.46 (95.92) \\
 R30:MBME           & 95.2 (95.6) & 2.0 (2.4)   & 47.0 (88.4)   & 0.09 (0.16) & 3.2 (2.4)   & 99.91 (99.83) & 98.47 (96.94) \\
 R30:CR 2           & 95.2 (95.5) & 2.0 (2.4)   & 88.8 (72.6)   & 0.16 (0.13) & 2.1 (2.6)   & 99.83 (99.86) & 97.16 (97.5)  \\
 R30:CR 5           & 95.1 (95.4) & 1.9 (2.3)   & 45.6 (40.8)   & 0.08 (0.07) & 3.1 (3.8)   & 99.91 (99.92) & 98.54 (98.61) \\
 R30:CR 10          & 95.2 (95.4) & 2.0 (2.2)   & 35.4 (28.8)   & 0.06 (0.05) & 3.8 (4.8)   & 99.93 (99.94) & 98.85 (99.03) \\
 E30:Replace        & 95.6 (95.4) & 2.5 (2.2)   & 688 (689)     & 1.25 (1.25) & 1.2 (1.2)   & 98.68 (98.68) & 75.18 (75.12) \\
 E30:Majority Vote  & 95.9 (95.9) & 2.8 (2.8)   & 380 (395)     & 0.69 (0.72) & 1.4 (1.4)   & 99.27 (99.24) & 86.44 (85.83) \\
 E30:MB             & 96.0 (96.3) & 2.8 (3.1)   & 253 (230)     & 0.46 (0.42) & 1.6 (1.7)   & 99.51 (99.56) & 91.14 (90.39) \\
 E30:MBME           & 96.0 (96.2) & 2.8 (3.0)   & 96 (163)      & 0.17 (0.3)  & 2.5 (1.9)   & 99.82 (99.69) & 96.45 (93.3)  \\
 E30:CR 2           & 96.0 (96.2) & 2.8 (3.1)   & 192 (160)     & 0.35 (0.29) & 1.7 (2.0)   & 99.63 (99.69) & 93.12 (93.28) \\
 E30:CR 5           & 96.0 (96.2) & 2.8 (3.1)   & 103 (94)      & 0.19 (0.17) & 2.4 (2.6)   & 99.8 (99.82)  & 96.07 (96.01) \\
 E30:CR 10          & 96.0 (96.3) & 2.9 (3.1)   & 89 (57)       & 0.16 (0.1)  & 2.6 (3.8)   & 99.83 (99.89) & 96.67 (97.65) \\
 \midrule
 R20:Replace        & 94.5 (94.5) & 1.3 (1.3)   & 274.6 (274.6) & 0.5 (0.5)   & 1.2 (1.2)   & 99.47 (99.47) & 91.94 (91.94) \\
 R20:Majority Vote  & 94.9 (94.9) & 1.7 (1.7)   & 195.2 (195.2) & 0.35 (0.35) & 1.4 (1.4)   & 99.62 (99.62) & 94.06 (94.06) \\
 R20:MB             & 94.7 (95.3) & 1.6 (2.2)   & 107.0 (83.2)  & 0.19 (0.15) & 1.7 (2.3)   & 99.79 (99.84) & 96.86 (97.29) \\
 R20:MBME           & 94.6 (95.3) & 1.4 (2.2)   & 31.0 (62.0)   & 0.06 (0.11) & 3.3 (2.7)   & 99.94 (99.88) & 99.09 (97.98) \\
 R20:CR 2           & 94.5 (95.2) & 1.4 (2.0)   & 65.0 (56.4)   & 0.12 (0.1)  & 2.0 (2.8)   & 99.87 (99.89) & 98.08 (98.17) \\
 R20:CR 5           & 94.7 (95.2) & 1.5 (2.0)   & 26.6 (25.6)   & 0.05 (0.05) & 3.8 (4.9)   & 99.95 (99.95) & 99.19 (99.16) \\
 R20:CR 10          & 94.6 (94.9) & 1.5 (1.8)   & 24.6 (19.6)   & 0.04 (0.04) & 4.0 (5.6)   & 99.95 (99.96) & 99.25 (99.38) \\
 E20:Replace        & 95.8 (95.7) & 2.6 (2.5)   & 462 (478)     & 0.84 (0.87) & 1.3 (1.3)   & 99.11 (99.08) & 83.35 (83.11) \\
 E20:Majority Vote  & 95.9 (96.0) & 2.8 (2.8)   & 336 (347)     & 0.61 (0.63) & 1.4 (1.4)   & 99.35 (99.33) & 88.2 (87.94)  \\
 E20:MB             & 96.0 (96.0) & 2.9 (2.8)   & 217 (207)     & 0.39 (0.38) & 1.7 (1.7)   & 99.58 (99.6)  & 92.45 (91.74) \\
 E20:MBME           & 96.0 (95.9) & 2.9 (2.8)   & 82 (148)      & 0.15 (0.27) & 2.8 (1.9)   & 99.84 (99.72) & 97.02 (94.17) \\
 E20:CR 2           & 96.1 (96.0) & 2.9 (2.8)   & 156 (133)     & 0.28 (0.24) & 1.9 (2.1)   & 99.7 (99.75)  & 94.53 (94.67) \\
 E20:CR 5           & 96.0 (96.0) & 2.9 (2.9)   & 84 (80)       & 0.15 (0.15) & 2.7 (2.8)   & 99.84 (99.85) & 96.86 (96.73) \\
 E20:CR 10          & 96.0 (96.1) & 2.9 (2.9)   & 75 (55)       & 0.14 (0.1)  & 2.9 (3.7)   & 99.86 (99.89) & 97.22 (97.89) \\
 \midrule
 R10:Replace        & 94.1 (94.1) & 0.9 (0.9)   & 129.2 (129.2) & 0.23 (0.23) & 1.3 (1.3)   & 99.75 (99.75) & 96.31 (96.31) \\
 R10:Majority Vote  & 94.3 (94.3) & 1.1 (1.1)   & 109.4 (109.4) & 0.2 (0.2)   & 1.5 (1.5)   & 99.79 (99.79) & 96.86 (96.86) \\
 R10:MB             & 94.0 (94.7) & 0.9 (1.6)   & 64.0 (51.0)   & 0.12 (0.09) & 1.7 (2.5)   & 99.88 (99.9)  & 98.23 (98.49) \\
 R10:MBME           & 93.8 (94.6) & 0.6 (1.4)   & 14.2 (36.2)   & 0.03 (0.07) & 3.3 (2.9)   & 99.97 (99.93) & 99.61 (98.93) \\
 R10:CR 2           & 93.8 (94.7) & 0.7 (1.5)   & 35.2 (32.4)   & 0.06 (0.06) & 2.0 (3.3)   & 99.93 (99.94) & 99.03 (99.03) \\
 R10:CR 5           & 93.7 (94.3) & 0.5 (1.1)   & 11.0 (14.0)   & 0.02 (0.03) & 3.5 (5.1)   & 99.98 (99.97) & 99.69 (99.59) \\
 R10:CR 10          & 93.8 (94.3) & 0.7 (1.1)   & 6.8 (7.4)     & 0.01 (0.01) & 6.0 (8.5)   & 99.99 (99.99) & 99.81 (99.78) \\
 E10:Replace        & 95.3 (95.4) & 2.1 (2.2)   & 206 (201)     & 0.37 (0.37) & 1.5 (1.5)   & 99.6 (99.61)  & 93.45 (93.65) \\
 E10:Majority Vote  & 95.4 (95.4) & 2.2 (2.2)   & 200 (200)     & 0.36 (0.36) & 1.6 (1.6)   & 99.61 (99.61) & 93.67 (93.69) \\
 E10:MB             & 95.3 (95.7) & 2.1 (2.5)   & 140 (120)     & 0.25 (0.22) & 1.8 (2.0)   & 99.73 (99.77) & 95.59 (95.48) \\
 E10:MBME           & 95.2 (95.6) & 2.0 (2.5)   & 35 (99)       & 0.06 (0.18) & 3.9 (2.3)   & 99.93 (99.81) & 98.92 (96.32) \\
 E10:CR 2           & 95.2 (95.7) & 2.1 (2.5)   & 91 (78)       & 0.17 (0.14) & 2.1 (2.6)   & 99.82 (99.85) & 97.26 (97.08) \\
 E10:CR 5           & 95.2 (95.6) & 2.0 (2.4)   & 30 (52)       & 0.05 (0.09) & 4.3 (3.3)   & 99.94 (99.9)  & 99.05 (98.04) \\
 E10:CR 10          & 95.2 (95.5) & 2.0 (2.3)   & 29 (37)       & 0.05 (0.07) & 4.4 (4.1)   & 99.94 (99.93) & 99.09 (98.73) \\
\bottomrule
\end{tabular}
\hspace*{1.0cm}

\begin{tabular}{l|rrrrrrr}
\toprule
 \textbf{Strategy} (Budget \%)         &   \textbf{Acc} (\%) &   $\Delta$\textbf{Acc} (\%)&   $\Sigma$ \textbf{\NF} &   \textbf{\NFR}\, (\%) &  \textbf{ \PF}\, / \textbf{\NF}  &  \textbf{Avg. BTC} &  \textbf{Avg. BEC}\\
\midrule\midrule
 100:Oracle         & 91.2 (91.2) & 34.7 (34.7) & 0 (0)             & 0.0 (0.0)   & NaN (NaN)   & 100.0 (100.0) & 100.0 (100.0) \\
 R100:Replace       & 79.2 (79.2) & 22.7 (22.7) & 24214 (24214)     & 6.05 (6.05) & 1.2 (1.2)   & 91.37 (91.37) & 77.71 (77.71) \\
 R100:Majority Vote & 78.9 (78.9) & 22.3 (22.3) & 7352 (7352)       & 1.84 (1.84) & 1.8 (1.8)   & 97.18 (97.18) & 93.95 (93.95) \\
 E100:MB            & 77.1 (77.6) & 20.5 (21.0) & 4378 (3463)       & 1.09 (0.87) & 2.2 (2.5)   & 98.32 (98.65) & 96.45 (97.25) \\
 E100:MBME          & 77.3 (77.4) & 20.7 (20.8) & 3057 (3207)       & 0.76 (0.8)  & 2.7 (2.6)   & 98.78 (98.74) & 97.69 (97.51) \\
 E100:CR 2          & 77.1 (77.3) & 20.6 (20.8) & 3368 (1648)       & 0.84 (0.41) & 2.5 (4.2)   & 98.72 (99.36) & 97.19 (98.7)  \\
 E100:CR 5          & 77.1 (77.0) & 20.5 (20.5) & 2520 (990)        & 0.63 (0.25) & 3.0 (6.2)   & 99.06 (99.63) & 97.82 (99.19) \\
 E100:CR 10         & 77.0 (76.8) & 20.5 (20.3) & 2112 (788)        & 0.53 (0.2)  & 3.4 (7.4)   & 99.22 (99.71) & 98.15 (99.34) \\
 \midrule
 R50:Replace        & 78.4 (78.4) & 21.8 (21.8) & 11478.8 (11463.8) & 2.87 (2.87) & 1.5 (1.5)   & 95.82 (95.83) & 89.88 (89.89) \\
 R50:Majority Vote  & 78.1 (78.1) & 21.6 (21.6) & 5048.2 (5048.8)   & 1.26 (1.26) & 2.1 (2.1)   & 98.07 (98.07) & 95.92 (95.92) \\
 R50:MB             & 76.9 (76.9) & 20.4 (20.4) & 3022.8 (2509.8)   & 0.76 (0.63) & 2.7 (3.0)   & 98.83 (99.02) & 97.62 (98.05) \\
 R50:MBME           & 76.5 (76.7) & 20.0 (20.1) & 2325.4 (2437.2)   & 0.58 (0.61) & 3.1 (3.1)   & 99.08 (99.05) & 98.27 (98.12) \\
 R50:CR 2           & 76.8 (76.5) & 20.3 (19.9) & 2233.6 (1220.8)   & 0.56 (0.31) & 3.3 (5.1)   & 99.14 (99.53) & 98.22 (99.06) \\
 R50:CR 5           & 76.6 (76.1) & 20.1 (19.5) & 1603.8 (700.4)    & 0.4 (0.18)  & 4.1 (8.0)   & 99.39 (99.73) & 98.69 (99.44) \\
 R50:CR 10          & 76.7 (75.8) & 20.1 (19.2) & 1314.6 (530.0)    & 0.33 (0.13) & 4.8 (10.1)  & 99.51 (99.8)  & 98.92 (99.57) \\
 E50:Replace        & 78.9 (79.8) & 22.4 (23.2) & 15732 (16138)     & 3.93 (4.03) & 1.4 (1.4)   & 94.41 (94.3)  & 85.37 (84.73) \\
 E50:Majority Vote  & 78.7 (79.2) & 22.2 (22.6) & 6318 (6282)       & 1.58 (1.57) & 1.9 (1.9)   & 97.6 (97.64)  & 94.74 (94.67) \\
 E50:MB             & 77.8 (77.9) & 21.3 (21.4) & 3969 (3212)       & 0.99 (0.8)  & 2.3 (2.7)   & 98.48 (98.77) & 96.78 (97.4)  \\
 E50:MBME           & 77.6 (77.6) & 21.1 (21.1) & 2904 (2972)       & 0.73 (0.74) & 2.8 (2.8)   & 98.85 (98.85) & 97.79 (97.66) \\
 E50:CR 2           & 77.8 (77.7) & 21.2 (21.1) & 3014 (1547)       & 0.75 (0.39) & 2.8 (4.4)   & 98.86 (99.41) & 97.5 (98.75)  \\
 E50:CR 5           & 77.7 (77.3) & 21.2 (20.8) & 2214 (900)        & 0.55 (0.22) & 3.4 (6.8)   & 99.18 (99.67) & 98.09 (99.24) \\
 E50:CR 10          & 77.7 (77.1) & 21.1 (20.6) & 1832 (709)        & 0.46 (0.18) & 3.9 (8.3)   & 99.33 (99.74) & 98.4 (99.38)  \\
 \midrule
 R30:Replace        & 77.4 (77.3) & 20.8 (20.8) & 6546.4 (6537.8)   & 1.64 (1.63) & 1.8 (1.8)   & 97.56 (97.56) & 94.53 (94.55) \\
 R30:Majority Vote  & 77.1 (77.1) & 20.5 (20.5) & 3616.4 (3613.6)   & 0.9 (0.9)   & 2.4 (2.4)   & 98.6 (98.6)   & 97.18 (97.18) \\
 R30:MB             & 75.9 (75.9) & 19.4 (19.4) & 2186.2 (1860.6)   & 0.55 (0.47) & 3.2 (3.6)   & 99.14 (99.27) & 98.35 (98.6)  \\
 R30:MBME           & 75.3 (75.8) & 18.7 (19.3) & 1859.6 (1855.6)   & 0.46 (0.46) & 3.5 (3.6)   & 99.26 (99.27) & 98.65 (98.61) \\
 R30:CR 2           & 75.5 (75.3) & 19.0 (18.7) & 1607.0 (918.2)    & 0.4 (0.23)  & 4.0 (6.1)   & 99.37 (99.64) & 98.8 (99.31)  \\
 R30:CR 5           & 75.1 (74.5) & 18.5 (17.9) & 1087.2 (488.4)    & 0.27 (0.12) & 5.3 (10.2)  & 99.58 (99.81) & 99.17 (99.63) \\
 R30:CR 10          & 74.9 (73.9) & 18.4 (17.3) & 875.4 (341.8)     & 0.22 (0.09) & 6.2 (13.7)  & 99.66 (99.87) & 99.33 (99.74) \\
 E30:Replace        & 78.5 (79.1) & 22.0 (22.6) & 9708 (8764)       & 2.43 (2.19) & 1.6 (1.6)   & 96.53 (96.87) & 91.01 (91.87) \\
 E30:Majority Vote  & 78.5 (78.9) & 22.0 (22.3) & 5232 (4469)       & 1.31 (1.12) & 2.1 (2.2)   & 98.03 (98.32) & 95.63 (96.24) \\
 E30:MB             & 78.1 (78.1) & 21.6 (21.5) & 3375 (2450)       & 0.84 (0.61) & 2.6 (3.2)   & 98.71 (99.06) & 97.25 (98.02) \\
 E30:MBME           & 77.8 (77.8) & 21.2 (21.3) & 2577 (2407)       & 0.64 (0.6)  & 3.1 (3.2)   & 98.98 (99.07) & 98.04 (98.07) \\
 E30:CR 2           & 78.0 (77.7) & 21.5 (21.1) & 2578 (1221)       & 0.64 (0.31) & 3.1 (5.3)   & 99.02 (99.53) & 97.86 (99.02) \\
 E30:CR 5           & 78.0 (77.4) & 21.5 (20.8) & 1831 (670)        & 0.46 (0.17) & 3.9 (8.8)   & 99.32 (99.75) & 98.43 (99.43) \\
 E30:CR 10          & 77.9 (77.1) & 21.4 (20.6) & 1517 (491)        & 0.38 (0.12) & 4.5 (11.5)  & 99.44 (99.82) & 98.67 (99.57) \\
 \midrule
 R20:Replace        & 75.7 (75.7) & 19.1 (19.1) & 4171.0 (4183.8)   & 1.04 (1.05) & 2.1 (2.1)   & 98.41 (98.41) & 96.71 (96.7)  \\
 R20:Majority Vote  & 75.4 (75.4) & 18.9 (18.8) & 2690.2 (2696.6)   & 0.67 (0.67) & 2.8 (2.7)   & 98.95 (98.94) & 97.99 (97.98) \\
 R20:MB             & 74.0 (74.6) & 17.5 (18.0) & 1662.8 (1421.2)   & 0.42 (0.36) & 3.6 (4.2)   & 99.34 (99.44) & 98.81 (98.97) \\
 R20:MBME           & 73.5 (74.3) & 16.9 (17.8) & 1480.6 (1421.8)   & 0.37 (0.36) & 3.9 (4.1)   & 99.4 (99.43)  & 98.97 (98.97) \\
 R20:CR 2           & 73.2 (73.6) & 16.7 (17.0) & 1205.6 (718.2)    & 0.3 (0.18)  & 4.5 (6.9)   & 99.52 (99.71) & 99.15 (99.48) \\
 R20:CR 5           & 72.3 (72.2) & 15.8 (15.7) & 768.0 (344.0)     & 0.19 (0.09) & 6.1 (12.4)  & 99.69 (99.86) & 99.46 (99.75) \\
 R20:CR 10          & 71.9 (71.0) & 15.4 (14.4) & 603.8 (218.4)     & 0.15 (0.05) & 7.4 (17.5)  & 99.76 (99.91) & 99.57 (99.84) \\
 E20:Replace        & 78.5 (78.4) & 22.0 (21.8) & 6191 (5239)       & 1.55 (1.31) & 1.9 (2.0)   & 97.74 (98.1)  & 94.47 (95.3)  \\
 E20:Majority Vote  & 78.3 (78.0) & 21.8 (21.4) & 4295 (3258)       & 1.07 (0.81) & 2.3 (2.6)   & 98.37 (98.77) & 96.46 (97.32) \\
 E20:MB             & 77.9 (77.2) & 21.3 (20.6) & 2700 (1825)       & 0.68 (0.46) & 3.0 (3.8)   & 98.96 (99.29) & 97.86 (98.58) \\
 E20:MBME           & 77.6 (77.2) & 21.1 (20.6) & 2183 (1825)       & 0.55 (0.46) & 3.4 (3.8)   & 99.13 (99.29) & 98.39 (98.58) \\
 E20:CR 2           & 77.8 (76.6) & 21.3 (20.1) & 1999 (879)        & 0.5 (0.22)  & 3.7 (6.7)   & 99.23 (99.66) & 98.41 (99.32) \\
 E20:CR 5           & 77.7 (75.9) & 21.2 (19.4) & 1383 (448)        & 0.35 (0.11) & 4.8 (11.8)  & 99.47 (99.83) & 98.87 (99.63) \\
 E20:CR 10          & 77.7 (75.1) & 21.2 (18.6) & 1101 (266)        & 0.28 (0.07) & 5.8 (18.4)  & 99.58 (99.9)  & 99.08 (99.78) \\
 \midrule
 R10:Replace        & 71.3 (71.3) & 14.7 (14.8) & 1958.4 (1952.6)   & 0.49 (0.49) & 2.9 (2.9)   & 99.22 (99.22) & 98.63 (98.63) \\
 R10:Majority Vote  & 71.2 (71.2) & 14.7 (14.6) & 1481.4 (1489.6)   & 0.37 (0.37) & 3.5 (3.5)   & 99.4 (99.4)   & 98.98 (98.98) \\
 R10:MB             & 69.0 (70.0) & 12.5 (13.4) & 991.4 (865.4)     & 0.25 (0.22) & 4.1 (4.9)   & 99.59 (99.65) & 99.35 (99.42) \\
 R10:MBME           & 68.9 (70.1) & 12.4 (13.6) & 944.8 (878.4)     & 0.24 (0.22) & 4.3 (4.9)   & 99.61 (99.64) & 99.38 (99.41) \\
 R10:CR 2           & 67.9 (69.2) & 11.4 (12.6) & 703.6 (424.2)     & 0.18 (0.11) & 5.0 (8.4)   & 99.71 (99.83) & 99.54 (99.72) \\
 R10:CR 5           & 66.2 (66.9) & 9.6 (10.4)  & 393.4 (161.8)     & 0.1 (0.04)  & 7.1 (17.1)  & 99.84 (99.93) & 99.75 (99.89) \\
 R10:CR 10          & 65.3 (64.9) & 8.8 (8.3)   & 282.6 (89.0)      & 0.07 (0.02) & 8.7 (24.4)  & 99.88 (99.96) & 99.82 (99.94) \\
 E10:Replace        & 76.1 (77.0) & 19.5 (20.4) & 2468 (2216)       & 0.62 (0.55) & 3.0 (3.3)   & 99.04 (99.17) & 98.12 (98.18) \\
 E10:Majority Vote  & 75.9 (76.1) & 19.3 (19.5) & 2417 (1954)       & 0.6 (0.49)  & 3.0 (3.5)   & 99.06 (99.25) & 98.18 (98.45) \\
 E10:MB             & 75.3 (75.2) & 18.8 (18.7) & 1557 (1118)       & 0.39 (0.28) & 4.0 (5.2)   & 99.38 (99.56) & 98.89 (99.18) \\
 E10:MBME           & 75.2 (75.2) & 18.7 (18.7) & 1533 (1118)       & 0.38 (0.28) & 4.0 (5.2)   & 99.38 (99.56) & 98.92 (99.18) \\
 E10:CR 2           & 75.3 (74.3) & 18.7 (17.8) & 1118 (503)        & 0.28 (0.13) & 5.2 (9.8)   & 99.55 (99.8)  & 99.22 (99.63) \\
 E10:CR 5           & 75.2 (72.7) & 18.6 (16.2) & 700 (165)         & 0.18 (0.04) & 7.7 (25.5)  & 99.72 (99.93) & 99.51 (99.88) \\
 E10:CR 10          & 75.2 (70.7) & 18.6 (14.2) & 515 (71)          & 0.13 (0.02) & 10.1 (50.8) & 99.79 (99.97) & 99.64 (99.95) \\
\bottomrule
\end{tabular}}
\label{tab:app:softmax_all}
\end{table}

\begin{table}
\small
\caption{Full results of all experiments on \textbf{ObjectNet} under the standard setting of an \textbf{improving sequence of models} and \textbf{estimating confusion matrices on a separate split} of the data (i.e., not on the target data set) as explained in~\cref{sec:experimental_setup}. The first number in each cell refers to estimating only the diagonal elements (i.e., the class-specific accuracies) of confusion matrices, and the \textbf{second number (in brackets) refers to estimating the full confusion matrices} with smoothing \textbf{and incorporating soft labels}. The character \textbf{E} in front of a budget indicates that selection for re-evaluation is based on our posterior label \textbf{entropy criterion} (e.g., E30 refers to selecting the top 30\% samples with highest entropy), and the character \textbf{R} indicates selecting a \textbf{randomly} sampled subset without replacement. Note that entropy-based selection requires a posterior and is thus not applicable for the baselines Replace and Majority Vote.}
\vskip 0.1in
\centering
\resizebox{0.7\columnwidth}{!}{
\begin{tabular}{l|rrrrrrr}
\toprule
 \textbf{Strategy} (Budget \%)         &   \textbf{Acc} (\%) &   $\Delta$\textbf{Acc} (\%)&   $\Sigma$ \textbf{\NF} &   \textbf{\NFR}\, (\%) &  \textbf{ \PF}\, / \textbf{\NF} &  \textbf{Avg. BTC} &  \textbf{Avg. BEC}\\
\midrule \midrule
 100:Oracle         & 50.5 (50.5) & 42.6 (42.6) & 0 (0)           & 0.0 (0.0)   & NaN (NaN)   & 100.0 (100.0) & 99.99 (99.99) \\
 R100:Replace       & 31.9 (31.9) & 24.0 (24.0) & 16669 (16669)   & 5.62 (5.62) & 1.3 (1.3)   & 72.65 (72.65) & 92.61 (92.61) \\
 R100:Majority Vote & 29.6 (29.6) & 21.6 (21.6) & 4690 (4690)     & 1.58 (1.58) & 1.9 (1.9)   & 89.99 (89.99) & 98.02 (98.02) \\
 E100:MB            & 29.1 (29.6) & 21.2 (21.7) & 2477 (1984)     & 0.83 (0.67) & 2.6 (3.0)   & 94.46 (95.55) & 98.96 (99.16) \\
 E100:MBME          & 28.6 (29.1) & 20.6 (21.1) & 1599 (1774)     & 0.54 (0.6)  & 3.4 (3.2)   & 95.86 (95.83) & 99.34 (99.26) \\
 E100:CR 2          & 29.0 (29.2) & 21.0 (21.2) & 1876 (931)      & 0.63 (0.31) & 3.1 (5.2)   & 95.92 (97.9)  & 99.21 (99.61) \\
 E100:CR 5          & 28.8 (28.7) & 20.8 (20.7) & 1372 (543)      & 0.46 (0.18) & 3.8 (8.1)   & 97.18 (98.86) & 99.41 (99.77) \\
 E100:CR 10         & 28.7 (28.4) & 20.8 (20.5) & 1084 (422)      & 0.37 (0.14) & 4.6 (10.0)  & 97.82 (99.16) & 99.54 (99.82) \\
 \midrule
 R50:Replace        & 30.7 (30.7) & 22.7 (22.7) & 7583.8 (7588.8) & 2.56 (2.56) & 1.6 (1.6)   & 86.63 (86.61) & 96.69 (96.69) \\
 R50:Majority Vote  & 28.6 (28.7) & 20.7 (20.7) & 3281.6 (3291.2) & 1.11 (1.11) & 2.2 (2.2)   & 93.01 (93.01) & 98.62 (98.62) \\
 R50:MB             & 27.8 (28.7) & 19.9 (20.7) & 1676.6 (1410.0) & 0.56 (0.48) & 3.2 (3.7)   & 96.13 (96.83) & 99.3 (99.41)  \\
 R50:MBME           & 26.9 (28.3) & 18.9 (20.3) & 1280.4 (1360.8) & 0.43 (0.46) & 3.7 (3.8)   & 96.77 (96.89) & 99.48 (99.43) \\
 R50:CR 2           & 27.5 (28.1) & 19.6 (20.2) & 1165.4 (663.6)  & 0.39 (0.22) & 4.1 (6.6)   & 97.3 (98.51)  & 99.52 (99.72) \\
 R50:CR 5           & 27.2 (27.3) & 19.2 (19.4) & 780.2 (361.4)   & 0.26 (0.12) & 5.6 (10.9)  & 98.26 (99.23) & 99.67 (99.85) \\
 R50:CR 10          & 27.0 (26.7) & 19.1 (18.8) & 599.8 (253.0)   & 0.2 (0.09)  & 6.9 (14.8)  & 98.7 (99.47)  & 99.75 (99.89) \\
 E50:Replace        & 31.0 (31.4) & 23.1 (23.4) & 7882 (7329)     & 2.66 (2.47) & 1.5 (1.6)   & 86.26 (87.93) & 96.53 (96.75) \\
 E50:Majority Vote  & 29.5 (29.4) & 21.5 (21.5) & 3895 (3397)     & 1.31 (1.14) & 2.0 (2.2)   & 91.73 (93.32) & 98.36 (98.55) \\
 E50:MB             & 28.9 (30.0) & 20.9 (22.1) & 2079 (1610)     & 0.7 (0.54)  & 2.9 (3.5)   & 95.23 (96.66) & 99.13 (99.31) \\
 E50:MBME           & 28.2 (29.5) & 20.3 (21.5) & 1452 (1534)     & 0.49 (0.52) & 3.6 (3.6)   & 96.18 (96.75) & 99.41 (99.35) \\
 E50:CR 2           & 28.7 (29.6) & 20.8 (21.7) & 1506 (763)      & 0.51 (0.26) & 3.6 (6.3)   & 96.64 (98.41) & 99.37 (99.67) \\
 E50:CR 5           & 28.5 (29.0) & 20.6 (21.1) & 1049 (414)      & 0.35 (0.14) & 4.6 (10.5)  & 97.83 (99.21) & 99.55 (99.82) \\
 E50:CR 10          & 28.4 (28.7) & 20.5 (20.7) & 828 (309)       & 0.28 (0.1)  & 5.6 (13.4)  & 98.35 (99.43) & 99.64 (99.86) \\
 \midrule
 R30:Replace        & 29.0 (29.0) & 21.0 (21.1) & 4070.6 (4074.0) & 1.37 (1.37) & 2.0 (2.0)   & 92.19 (92.21) & 98.26 (98.26) \\
 R30:Majority Vote  & 27.3 (27.3) & 19.4 (19.4) & 2346.6 (2348.8) & 0.79 (0.79) & 2.5 (2.5)   & 94.91 (94.88) & 99.02 (99.02) \\
 R30:MB             & 25.9 (27.3) & 18.0 (19.4) & 1185.8 (1024.8) & 0.4 (0.35)  & 3.8 (4.5)   & 97.12 (97.58) & 99.52 (99.58) \\
 R30:MBME           & 25.1 (27.1) & 17.2 (19.2) & 1008.4 (988.6)  & 0.34 (0.33) & 4.2 (4.6)   & 97.41 (97.67) & 99.59 (99.59) \\
 R30:CR 2           & 25.4 (26.5) & 17.5 (18.5) & 782.2 (482.8)   & 0.26 (0.16) & 5.2 (8.1)   & 98.05 (98.87) & 99.68 (99.8)  \\
 R30:CR 5           & 24.6 (25.2) & 16.6 (17.2) & 476.2 (236.2)   & 0.16 (0.08) & 7.5 (14.6)  & 98.8 (99.45)  & 99.81 (99.9)  \\
 R30:CR 10          & 24.4 (24.5) & 16.4 (16.5) & 331.6 (154.2)   & 0.11 (0.05) & 10.2 (20.9) & 99.17 (99.65) & 99.86 (99.93) \\
 E30:Replace        & 29.0 (30.0) & 21.0 (22.1) & 4316 (3761)     & 1.45 (1.27) & 1.9 (2.1)   & 91.75 (93.51) & 98.14 (98.35) \\
 E30:Majority Vote  & 28.2 (28.2) & 20.3 (20.2) & 2970 (2359)     & 1.0 (0.79)  & 2.3 (2.6)   & 93.54 (95.37) & 98.76 (99.0)  \\
 E30:MB             & 27.8 (28.9) & 19.9 (20.9) & 1565 (1015)     & 0.53 (0.34) & 3.4 (4.8)   & 96.24 (97.78) & 99.35 (99.57) \\
 E30:MBME           & 26.9 (28.9) & 18.9 (20.9) & 1280 (1015)     & 0.43 (0.34) & 3.7 (4.8)   & 96.64 (97.78) & 99.48 (99.57) \\
 E30:CR 2           & 27.7 (28.1) & 19.7 (20.1) & 1074 (496)      & 0.36 (0.17) & 4.4 (8.5)   & 97.42 (98.9)  & 99.55 (99.79) \\
 E30:CR 5           & 27.4 (26.9) & 19.4 (19.0) & 689 (230)       & 0.23 (0.08) & 6.2 (16.3)  & 98.43 (99.52) & 99.71 (99.9)  \\
 E30:CR 10          & 27.1 (25.9) & 19.2 (18.0) & 504 (140)       & 0.17 (0.05) & 8.1 (24.8)  & 98.91 (99.71) & 99.79 (99.94) \\
 \midrule
 R20:Replace        & 27.0 (27.1) & 19.1 (19.1) & 2453.8 (2436.4) & 0.83 (0.82) & 2.4 (2.5)   & 94.82 (94.84) & 98.97 (98.98) \\
 R20:Majority Vote  & 25.8 (25.8) & 17.8 (17.9) & 1658.8 (1663.4) & 0.56 (0.56) & 3.0 (3.0)   & 96.17 (96.16) & 99.32 (99.32) \\
 R20:MB             & 23.8 (25.3) & 15.8 (17.3) & 887.2 (744.0)   & 0.3 (0.25)  & 4.3 (5.3)   & 97.74 (98.15) & 99.64 (99.7)  \\
 R20:MBME           & 23.5 (25.5) & 15.6 (17.5) & 793.8 (736.0)   & 0.27 (0.25) & 4.6 (5.4)   & 97.93 (98.18) & 99.68 (99.7)  \\
 R20:CR 2           & 22.6 (24.3) & 14.7 (16.4) & 530.0 (339.0)   & 0.18 (0.11) & 6.2 (10.0)  & 98.57 (99.14) & 99.79 (99.86) \\
 R20:CR 5           & 21.5 (22.7) & 13.5 (14.8) & 304.2 (160.8)   & 0.1 (0.05)  & 9.3 (18.1)  & 99.14 (99.6)  & 99.88 (99.93) \\
 R20:CR 10          & 20.9 (21.5) & 12.9 (13.6) & 215.6 (98.8)    & 0.07 (0.03) & 12.1 (26.6) & 99.4 (99.74)  & 99.91 (99.96) \\
 E20:Replace        & 26.9 (29.3) & 19.0 (21.3) & 2588 (2149)     & 0.87 (0.72) & 2.4 (2.8)   & 94.54 (96.08) & 98.91 (99.07) \\
 E20:Majority Vote  & 26.2 (27.5) & 18.3 (19.6) & 2390 (1742)     & 0.81 (0.59) & 2.4 (3.1)   & 94.8 (96.61)  & 99.0 (99.26)  \\
 E20:MB             & 26.0 (27.7) & 18.1 (19.7) & 1169 (763)      & 0.39 (0.26) & 3.9 (5.8)   & 97.01 (98.26) & 99.53 (99.68) \\
 E20:MBME           & 25.8 (27.7) & 17.8 (19.7) & 1131 (763)      & 0.38 (0.26) & 3.9 (5.8)   & 97.06 (98.26) & 99.54 (99.68) \\
 E20:CR 2           & 26.0 (26.6) & 18.1 (18.6) & 670 (353)       & 0.23 (0.12) & 6.0 (10.8)  & 98.14 (99.16) & 99.73 (99.85) \\
 E20:CR 5           & 26.0 (25.1) & 18.0 (17.2) & 369 (132)       & 0.12 (0.04) & 10.1 (25.2) & 98.97 (99.7)  & 99.85 (99.94) \\
 E20:CR 10          & 25.9 (24.2) & 17.9 (16.2) & 248 (73)        & 0.08 (0.02) & 14.4 (42.2) & 99.34 (99.83) & 99.9 (99.97)  \\
 \midrule
 R10:Replace        & 22.1 (22.1) & 14.2 (14.2) & 996.6 (997.2)   & 0.34 (0.34) & 3.6 (3.6)   & 97.49 (97.5)  & 99.6 (99.6)   \\
 R10:Majority Vote  & 21.6 (21.5) & 13.6 (13.5) & 808.8 (820.4)   & 0.27 (0.28) & 4.1 (4.1)   & 97.86 (97.83) & 99.68 (99.67) \\
 R10:MB             & 19.5 (21.0) & 11.6 (13.1) & 450.2 (371.2)   & 0.15 (0.13) & 5.8 (7.6)   & 98.7 (98.95)  & 99.82 (99.85) \\
 R10:MBME           & 19.1 (20.8) & 11.1 (12.8) & 446.4 (381.6)   & 0.15 (0.13) & 5.6 (7.2)   & 98.69 (98.92) & 99.83 (99.85) \\
 R10:CR 2           & 17.7 (19.8) & 9.8 (11.9)  & 287.2 (177.2)   & 0.1 (0.06)  & 7.3 (13.4)  & 99.11 (99.48) & 99.89 (99.93) \\
 R10:CR 5           & 15.6 (17.6) & 7.7 (9.7)   & 151.6 (75.4)    & 0.05 (0.03) & 10.4 (24.9) & 99.5 (99.77)  & 99.94 (99.97) \\
 R10:CR 10          & 14.6 (15.8) & 6.7 (7.9)   & 94.6 (36.0)     & 0.03 (0.01) & 14.1 (41.8) & 99.68 (99.89) & 99.96 (99.99) \\
 E10:Replace        & 23.7 (25.8) & 15.7 (17.9) & 996 (683)       & 0.34 (0.23) & 3.9 (5.9)   & 97.5 (98.46)  & 99.6 (99.72)  \\
 E10:Majority Vote  & 23.7 (25.4) & 15.7 (17.5) & 996 (663)       & 0.34 (0.22) & 3.9 (5.9)   & 97.5 (98.48)  & 99.6 (99.72)  \\
 E10:MB             & 22.7 (24.4) & 14.8 (16.5) & 696 (364)       & 0.23 (0.12) & 4.9 (9.4)   & 98.08 (99.03) & 99.72 (99.85) \\
 E10:MBME           & 22.7 (24.4) & 14.8 (16.5) & 696 (364)       & 0.23 (0.12) & 4.9 (9.4)   & 98.08 (99.03) & 99.72 (99.85) \\
 E10:CR 2           & 20.7 (22.7) & 12.8 (14.8) & 427 (176)       & 0.14 (0.06) & 6.6 (16.6)  & 98.68 (99.52) & 99.83 (99.93) \\
 E10:CR 5           & 18.2 (19.4) & 10.3 (11.4) & 197 (50)        & 0.07 (0.02) & 10.7 (43.4) & 99.32 (99.86) & 99.92 (99.98) \\
 E10:CR 10          & 17.2 (15.9) & 9.2 (7.9)   & 122 (28)        & 0.04 (0.01) & 15.0 (53.5) & 99.57 (99.92) & 99.95 (99.99) \\
\bottomrule
\end{tabular}}
\label{tab:app:objectnet_softmax}
\end{table}

\end{document}